\documentclass{article}

    \usepackage[final]{neurips_2025}

\usepackage[utf8]{inputenc}   % allow utf-8 input
\usepackage[T1]{fontenc}      % use 8-bit T1 fonts
\usepackage{url}              % simple URL typesetting
\usepackage{booktabs}         % professional-quality tables
\usepackage{amsfonts}         % blackboard math symbols
\usepackage{nicefrac}         % compact symbols for 1/2, etc.
\usepackage{microtype}        % microtypography
\usepackage{enumitem}

\usepackage{xcolor} % ensures color support
\definecolor{myblue}{RGB}{6, 69, 173}
\usepackage[
    colorlinks=true,
    linkcolor=myblue,
    urlcolor=myblue
]{hyperref}

\usepackage{makecell}
\usepackage{adjustbox}
\usepackage{multirow}
\usepackage{wrapfig}
\usepackage{subcaption}

\usepackage{fontawesome}
\usepackage[utf8]{inputenc}
\usepackage{academicons}

\usepackage[most]{tcolorbox}
\usepackage{etoc}
\usepackage{tocloft}
\usepackage{titletoc}

\newcommand{\dataset}{Mars-Bench}

\title{
\vspace{-1.0em}
\begin{center}
  \begin{minipage}[c]{0.12\textwidth}
    \centering
    \includegraphics[height=3em]{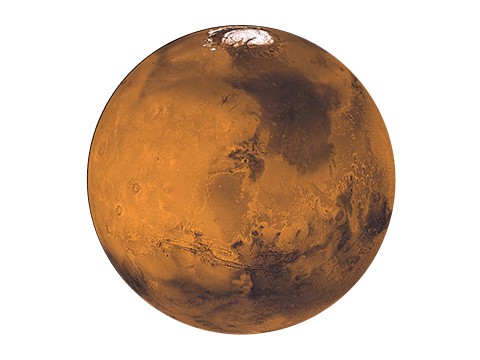}
  \end{minipage}%
  \hspace{1em}%
  \begin{minipage}[c]{0.7\textwidth}
    \raggedright
    \textbf{\Large Mars-Bench: A Benchmark for Evaluating Foundation Models for Mars Science Tasks}
  \end{minipage}
\end{center}
\vspace{-1em}
}

\author{Mirali Purohit$^{1, 3}$\textsuperscript{ \faEnvelope} \quad Bimal Gajera$^{1*}$ \quad Vatsal Malaviya$^{1*}$ \quad Irish Mehta$^{1*}$ \quad \textbf{Kunal Kasodekar}$^1$ \\ \textbf{Jacob Adler}$^2$ \quad \textbf{Steven Lu}$^3$ \quad \textbf{Umaa Rebbapragada}$^3$ \quad \textbf{Hannah Kerner}$^1$ \\\\ 
$^1$School of Computing and Augmented Intelligence, Arizona State University, Tempe, AZ, USA \\ 
$^2$School of Earth and Space Exploration, Arizona State University, Tempe, AZ, USA \\ 
$^3$Jet Propulsion Laboratory, California Institute of Technology, Pasadena, CA, USA
}

\begin{document}

\maketitle

\begin{abstract}

Foundation models have enabled rapid progress across many specialized domains by leveraging large-scale pre-training on unlabeled data, demonstrating strong generalization to a variety of downstream tasks. While such models have gained significant attention in fields like Earth Observation, their application to Mars science remains limited. A key enabler of progress in other domains has been the availability of standardized benchmarks that support systematic evaluation. In contrast, Mars science lacks such benchmarks and standardized evaluation frameworks, which have limited progress toward developing foundation models for Martian tasks. To address this gap, we introduce \dataset, the first benchmark designed to systematically evaluate models across a broad range of Mars-related tasks using both orbital and surface imagery. \dataset{} comprises 20 datasets spanning classification, segmentation, and object detection, focused on key geologic features such as craters, cones, boulders, and frost. We provide standardized, ready-to-use datasets and baseline evaluations using models pre-trained on natural images, Earth satellite data, and state-of-the-art vision-language models. Results from all analyses suggest that Mars-specific foundation models may offer advantages over general-domain counterparts, motivating further exploration of domain-adapted pre-training. \dataset{} aims to establish a standardized foundation for developing and comparing machine learning models for Mars science. Our data, models, and code are available at: \href{https://mars-bench.github.io/}{https://mars-bench.github.io/}.

\def\thefootnote{\faEnvelope}\footnotetext{Corresponding Author: mpurohi3@asu.edu}\def\thefootnote{\english{footnote}}
\def\thefootnote{*}\footnotetext{Equal Contribution}\def\thefootnote{\english{footnote}}

\end{abstract}

\section{Introduction}
\label{sec:introduction}

Over the past few years, foundation models have revolutionized specialized domains such as medical imaging \cite{moor2023med, parmar022boxbart}, Earth Observation (EO) \cite{klemmer2023satclip, reed2023scale, astruc2024anysat}, law \cite{colombo2024saullm, cui2023chatlaw}, and astronomy \cite{lanusse2023astroclip, nguyen2023astrollama, slijepcevic2024radio}. These models, pre-trained on large and diverse datasets, offer strong generalization capabilities and enable efficient fine-tuning on downstream tasks with minimal data. The EO community has embraced foundation models in the last 3-4 years, with an explosion of methods, datasets, and benchmarks aimed at improving performance across a wide range of geospatial tasks.

The key driver of progress in these domains has been the development of high-quality, standardized benchmarks. For example, BigBio \cite{fries2022bigbio} and MIMIC-IV \cite{johnson2023mimic} have accelerated model advancements by providing consistent evaluation protocols for medical applications. Benchmarks like Geo-Bench \cite{lacoste2023geo} and PANGAEA \cite{marsocci2024pangaea} have accelerated progress in EO applications by providing a suite of standardized classification and segmentation tasks for evaluating geospatial foundation models. Geo-Bench enables model developers to assess generalization across diverse data sources and use cases, creating a pathway for systematic progress. 

However, no such benchmark exists for Martian applications. Machine learning research for Mars science applications thus lags behind other science domains \cite{azari2020integrating}. Although recent studies have presented machine learning solutions for a range of Martian applications, including crater detection \cite{malvi2023automated, zheng2025automatic, delatte2019segmentation}, landmark classification \cite{wilhelm2020domars16k, wagstaff2021mars}, and cone segmentation \cite{purohit2024conequest, yang2024mapping}, these solutions and datasets lack standardization and interoperability. This results in task-specific models or datasets that cannot be easily evaluated as downstream tasks for foundation models or other machine learning advances. This results in limited evaluation of proposed Mars foundation model approaches on 1-2 downstream tasks, limiting the ability to assess model generalization or robustness \cite{vincent2024clover, ward2022improving, wang2023semi, goh2023self, purohit2024investigating}.

% Most existing datasets were created by domain experts, i.e., planetary scientists or geologists, and often lack the standardization and documentation necessary for seamless ML integration.
% A critical bottleneck is the lack of structured, machine-learning-ready datasets and benchmarks.

\begin{figure}
    \centering

    \begin{subfigure}{0.66\columnwidth}
        \centering
        \includegraphics[width=\linewidth]{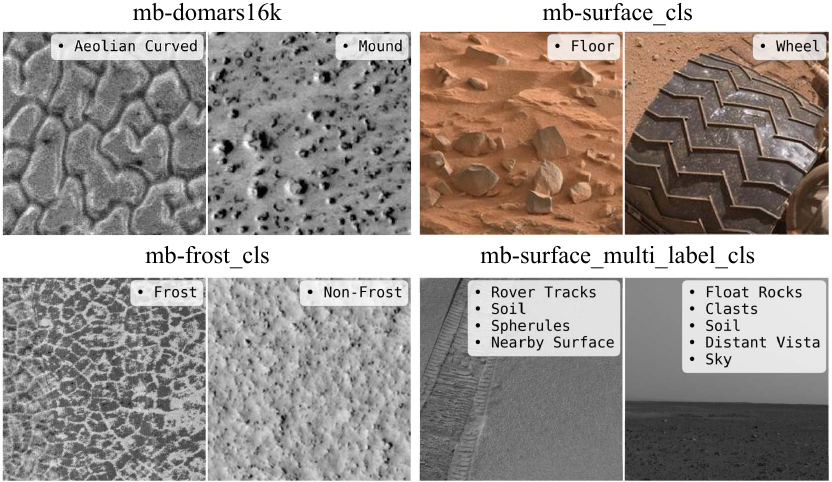}
        \caption{Classification}
    \end{subfigure}\hfill
    \begin{subfigure}{0.33\columnwidth}
        \centering
        \includegraphics[width=\linewidth]{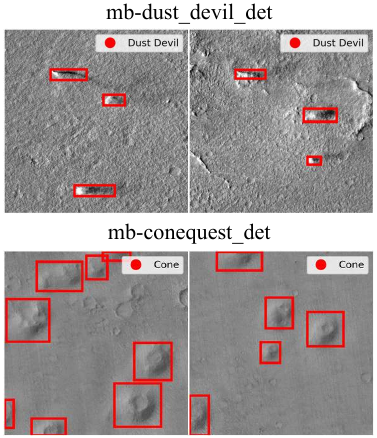}
        \caption{Object Detection}
    \end{subfigure}

    \vspace{0.05em}

    \begin{subfigure}{0.98\columnwidth}
        \centering
        \includegraphics[width=\linewidth]{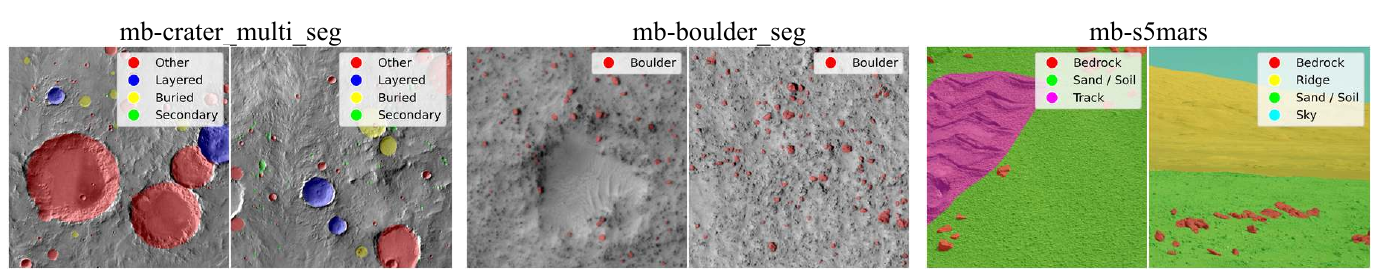}
        \caption{Segmentation}
    \end{subfigure}

    \caption{Representative samples from selected \dataset{} datasets, from all three task categories.}
    \label{fig:marsbench}
\end{figure}

This gap is particularly surprising given the richness of available Mars data. Orbiters such as the Mars Reconnaissance Orbiter (MRO) \cite{zurek2007overview} and Mars Odyssey have captured millions of images over the last 20-25 years, while surface rovers like Curiosity and Perseverance have amassed petabytes of high-resolution images. These datasets offer immense potential to study critical questions of planetary science, such as the past presence of water on Mars and the planet's habitability. Yet, the full value of these datasets remains untapped by the ML community due to their lack of standardization, incomplete documentation, and inconsistent formatting for ML workflows.

We introduce \dataset, the first comprehensive benchmark designed to systematically evaluate machine learning models across a diverse set of Mars-related tasks using both orbital and surface imagery. To create this benchmark, we curated and revamped existing datasets, performing quality checks and corrections where necessary and standardizing them in a unified, ML-ready format. The goal of \dataset{} is to provide a common framework to assess and compare the performance of foundation models on Martian data, facilitating reproducibility and accelerating scientific discovery in planetary science. Our key contributions are as follows:

\begin{itemize}[leftmargin=*]

    \item \textbf{Diverse task coverage:} \dataset{} includes 20 datasets, summarized in Table \ref{tab:marsbench}, spanning three task types: classification, segmentation, and object detection. We also provide a few-shot and partitioned versions of each dataset for evaluation under varying training sample sizes.

    \item \textbf{Scientific relevance:} \dataset{} covers a wide range of geologic features commonly studied in Mars science, including craters, cones, boulders, landslides, dust devils, atmospheric dust, etc. These tasks reflect real scientific use cases relevant to planetary scientists and geologists, who co-developed the \dataset. Samples from few \dataset{} datasets are shown in Figure \ref{fig:marsbench}.

    \item \textbf{Comprehensive evaluation:} Since no standardized pre-trained model exists for Mars data, we benchmarked performance using ImageNet-pretrained models under different training settings. We analyzed model behavior with different training set sizes. We also evaluated \dataset{} using pre-trained EO models as well as proprietary vision-language models, including Gemini and GPT.
    
    % As no standardized pre-trained model exists for Mars data, we benchmark performance using ImageNet-pretrained models under various training type setting. With this, we show analysis of model behavior w.r.t. training size. We also did evaluate \dataset{} on pre-trained EO models and propriarty vision-language models.

    % both feature extraction and end-to-end fine-tuning settings. We also compare these results to models trained from scratch, providing a detailed analysis in the Appendix.

    \item \textbf{Code, reproducibility, and baseline models:} We release full code support for all experiments in this paper, along with tools for dataset handling and results visualization. To facilitate community adoption and reproducibility, we also provide well-documented guidelines and publicly release all baseline models evaluated on \dataset{}. These models can serve as strong starting points for future applications; for example, generating initial global maps of specific geologic features (e.g., cones), which experts can later refine with minimal annotation effort.

    % \item \textbf{Code and reproducibility resources:} We release full code support for all experiments presented in the paper, along with data and results visualization. We provide well-documented guidelines and utilities to ensure reproducibility and facilitate community adoption.

    % \item \textbf{Baseline model release:} Baseline models evaluated on \dataset{} are released publicly. These baseline models can serve as starting points for future applications, e.g., enabling researchers to generate initial global maps of specific features (such as cones), which can then be refined with minimal annotation effort by an expert.

\end{itemize}

% Via Mars-Bench, our goal is to provide a common framework to assess and compare the performance of foundation models on Martian data, facilitating reproducibility and accelerating scientific discovery in planetary science.
 
% To address these challenges, we introduce Mars-Bench, the first benchmark designed to systematically evaluate models across a diverse set of Mars-related tasks using both orbital and surface imagery. Our goal is to provide a common framework to assess and compare the performance of foundation models on Martian data, facilitating reproducibility and accelerating scientific discovery in planetary science.

% We address these limitations and propose the first dataset benchmark called \dataset{}, which consists of various tasks for Mars science applications. \dataset{} has 2 types of datasets where the way the camera/sensor collects the data changes, i.e., orbiter/satellite and rover.

\section{Related Work}
\label{sec:related_work}

Over the past decade, evaluation benchmarks have played a fundamental role in identifying the limitations of existing foundation models, steering their progress in natural language processing (NLP) and computer vision (CV). For instance, general-purpose natural language understanding (NLU) benchmarks \cite{wang2019superglue, wang2024mmlupro, srivastava2023beyond} have facilitated the development of large language models (LLMs) such as GPT \cite{brown2020language}, LLaMA \cite{touvron2023llama}, and Gemini \cite{team2023gemini}. Even in specialized domains, including medical \cite{parmar022boxbart, fries2022bigbio, johnson2023mimic}, legal \cite{fei2023lawbench, guha2023legalbench}, scientific discovery \cite{majumder2024discoverybench, chen2024scienceagentbench}, security \cite{bhusal2024secure}, and finance \cite{islam2023financebench}, various benchmarks have driven progress in building domain-specific foundation models. Thus, development of quality evaluation benchmarks is necessary for building better foundation models.

In the remote sensing domain, Geo-Bench \cite{lacoste2023geo} has defined standardized evaluation protocols for a broad set of EO tasks and has quickly become a de facto benchmark. Since its release, Geo-Bench has been used to evaluate most foundation models proposed for EO over the past two years, enabling consistent comparisons across models. Other notable efforts include SustainBench \cite{yeh2021sustainbench}, which targets seven sustainable development goals, AiTLAS \cite{dimitrovski2023current}, which aggregates 22 EO datasets focused solely on classification tasks, and PANGAEA \cite{marsocci2024pangaea}, which includes 11 evaluation datasets covering diverse satellite sensors.

Despite substantial progress in other domains toward foundation models and dataset benchmarks, no benchmark currently exists for Mars science applications. The absence of a standardized evaluation framework has hindered the development of foundation models (and machine learning solutions more generally) for Mars-related tasks. While specialized datasets exist across different applications, most require significant effort to restructure into an ML-ready format or make interoperable with other datasets. Furthermore, some datasets are not usable without expert guidance from planetary scientists, further slowing progress. To address this gap, we introduce \textbf{\dataset}, the first benchmark to facilitate the development and evaluation of foundation models for Mars science tasks.

\section{\dataset}
\label{sec:mars_bench}

\dataset{} was created by curating, organizing, restructuring, and correcting existing Mars science datasets following the design principles explained in Section \ref{subsec:design_process}. While creating each dataset, our goal was to ensure accessibility and usability and provide task diversity as described in Section \ref{subsec:tasks_and_datasets}.

% provide ready-to-use, unified format datasets that can be easily interpretable by the user
% tasks, datasets, camera sensors, geologic features

\begin{table*}[t]
\centering

\resizebox{0.99\linewidth}{!}{

    \begin{tabular}{rccccccccccc}

        \multicolumn{11}{l}{\large\textbf{Classification}} \\
        \midrule
        Name & \makecell{Observation\\Source} & \makecell{Geologic\\Feature} & \makecell{Image\\Size} & \# Classes & Train & Val & Test & \# Bands & \makecell{Sensor/\\Instrument} & \makecell{Published\\Year} & Cite \\
        \midrule
        mb-atmospheric\_dust\_cls\_edr & MRO (O) & Atmospheric dust & $100 \times 100$ & 2 & 9817 & 4969 & 5214 & 1 & HiRISE & 2019 & \cite{doran20193495068} \\
        mb-atmospheric\_dust\_cls\_rdr & MRO (O) & Atmospheric dust & $100 \times 100$ & 2 & 9817 & 4969 & 5214 & 1 & HiRISE & 2019 & \cite{doran20193495068} \\
        mb-change\_cls\_ctx & MRO (O) & Surface change & $150 \times 150$ & 2 & 36 & 10 & 10 & 1 & CTX & 2019 & \cite{kerner2019toward} \\
        mb-change\_cls\_hirise & MRO (O) & Surface change & $100 \times 100$ & 2 & 3103 & 670 & 670 & 1 & HiRISE & 2019 & \cite{kerner2019toward} \\
        mb-domars16k & MRO (O) & Landmark & $200 \times 200$ & 15 & 11305 & 3231 & 1614 & 1 & CTX & 2020 & \cite{wilhelm2020domars16k} \\
        mb-frost\_cls & MRO (O) & Frost & $299 \times 299$ & 2 & 30124 & 11415 & 12249 & 1 & HiRISE & 2024 & \cite{doran2024evaluating} \\
        mb-landmark\_cls & MRO (O) & Landmark & $227 \times 227$ & 8 & 6997 & 2025 & 1793 & 1 & HiRISE & 2021 & \cite{wagstaff2021mars} \\
        mb-surface\_cls & Curiosity (R) & Surface & $256 \times 256$ & 36 & 6580 & 1293 & 1594 & 3 & Mastcam, MAHLI & 2018, 2021 & \cite{wagstaff2021mars, wagstaff2018deep}\\
        mb-surface\_multi\_label\_cls & \makecell{Opportunity, Spirit (R)} & Surface & $1024 \times 1024$ & 25 & 1762 & 443 & 739 & 1 & Pancam & 2020 & \cite{cole2020identifying} \\
        \midrule

        \\

        \multicolumn{11}{l}{\large\textbf{Segmentation}} \\
        \midrule
        Name & \makecell{Observation\\Source} & \makecell{Geologic\\Feature} & \makecell{Image\\Size} & \# Classes & Train & Val & Test & \# Bands & \makecell{Sensor/\\Instrument} & \makecell{Published\\Year} & Cite \\
        \midrule
        mb-boulder\_seg & MRO (O) & Boulder & $500 \times 500$ & 2 & 39 & 6 & 4 & 1 & HiRISE & 2023 & \cite{prieur2023automatic} \\
        mb-conequest\_seg & MRO (O) & Cone & $512 \times$ 512 & 2 & 2236 & 319 & 643 & 1 & CTX & 2024 & \cite{purohit2024conequest} \\
        mb-crater\_binary\_seg & Mars Odyssey (O) & Crater & $512 \times 512$ & 2 & 3600 & 900 & 900 & 1 & THEMIS & 2012 & \cite{robbins2012new} \\
        mb-crater\_multi\_seg & Mars Odyssey (O) & Crater & $512 \times 512$ & 5 & 3600 & 900 & 900 & 1 & THEMIS & 2021 & \cite{lagain2021mars} \\
        mb-mars\_seg\_mer & \makecell{Opportunity, Spirit (R)} & Terrain & $1024 \times 1024$ & 7 & 744 & 106 & 214 & 1 & Navcam, Pancam & 2022 & \cite{li2022stepwise} \\
        mb-mars\_seg\_msl & Curiosity (R) & Terrain & $500 \times 560$ & 7 & 2893 & 413 & 828 & 3 & Mastcam & 2022 & \cite{li2022stepwise} \\
        mb-mmls & MRO (O) & Landslide & $128 \times 128$ & 2 & 275 & 31 & 256 & 7 & CTX & 2024 & \cite{paheding2024marsls} \\
        mb-s5mars & Curiosity (R) & Terrain & $1200 \times 1200$ & 10 & 4997 & 200 & 800 & 3 & Mastcam & 2022 & \cite{zhang2022s} \\
        \midrule

        \\

        \multicolumn{11}{l}{\large\textbf{Object Detection}} \\
        \midrule
        Name & \makecell{Observation\\Source} & \makecell{Geologic\\Feature} & \makecell{Image\\Size} & \# Classes & Train & Val & Test & \# Bands & \makecell{Sensor/\\Instrument} & \makecell{Published\\Year} & Cite \\
        \midrule
        mb-boulder\_det & MRO (O) & Boulder & $500 \times 500$ & 1 & 39 & 6 & 4 & 1 & HiRISE & 2023 & \cite{prieur2023automatic} \\
        mb-conequest\_det & MRO (O) & Cone & $512 \times 512$ & 1 & 1158 & 167 & 333 & 1 & CTX & 2024 & \cite{purohit2024conequest} \\
        mb-dust\_devil\_det & MRO (O) & Dust devil & $\sim 750 \times 750$ & 1 & 1404 & 201 & 402 & 1 & CTX & 2024 & \cite{guo2024martian} \\
        \bottomrule

    \end{tabular}
}
\caption{Overview of \dataset{} datasets across all three task categories. To distinguish the benchmarked versions from their original sources, all dataset names are prefixed with "mb-", which indicates \dataset{}. Observation sources are labeled as O (Orbiter) and R (Rover). Refer to Appendix \ref{subsec:details_of_marsbench} for a detailed description and illustrative examples from each dataset.}
\label{tab:marsbench}
\end{table*}

% Overview of \dataset{} datasets across all three categories. To distinguish the \dataset{} version from the original dataset, and as some datasets are modified, we prefixed the datasets with "mb-", which stands for \dataset{}. Here, O: Orbiter and R: Rover.

% \textbf{Mars-Bench}: Characteristics of datasets in the benchmark. Since datasets are modified from their original published versions, we prepend their names with “mb-” to distinguish the Mars-Bench version from the original dataset.

\subsection{Design Principles}
\label{subsec:design_process}

\textbf{Ease of Use} A key goal was to create an accessible and user-friendly ready-to-use benchmark, supported by standardized data-loading code. We focused on unifying the data format across all tasks to reduce the engineering effort for researchers and practitioners using the dataset. We provide all possible formats in each task if there are multiple common formats. For example, different object detection models may require COCO, Pascal VOC, or YOLO format, so we provide annotations in all three formats to ensure it is easily usable in all cases and reduce time for conversion from one format to another.

% \textbf{Ease of Use} A central goal of \dataset{} is to provide an accessible, user-friendly, and ready-to-use benchmark, supported by standardized data-loading utilities. To minimize engineering overhead, we unified the data format across tasks, ensuring consistency in structure and labeling. In addition, we provide task-specific annotations in multiple widely used formats where applicable—for example, for object detection tasks, annotations are made available in COCO, Pascal VOC, and YOLO formats. This eliminates the need for users to manually convert annotation formats, streamlining experimentation and enabling seamless integration with existing model pipelines.

% \paragraph{\textbf{Correction in Data by Expert}} As in the specialized domain, it is very crucial to have quality analyzed by an expert, we also did that wherever needed. Specifically, we did quality analysis by expert in all segmentation datasets, and some classification datasets were modified and corrected by contacting the author of the original dataset. Details of which datasets have undergone correction is provided in the Appendix.

\textbf{Expert-Validated Corrections} Given the domain-specific nature of Mars science, ensuring high data quality is critical. We conducted expert-driven quality analysis and corrections wherever necessary. All segmentation datasets underwent validation by domain experts, and several classification datasets were reviewed and revised through direct correspondence with the original dataset authors. Details on which datasets were corrected or modified are provided in Appendix \ref{subsec:correction}.

% \paragraph{Dataset Transformation} As mentioned in Geo-Bench, where they transform data by reducing the number of samples and removing data imbalance, we did not transform our datasets based on these 2 aspects. The reason is that the training size of all of our datasets is less than 10k, except for one. The reason why Mars datasets are smaller is that they always require an experts (geologists or planetary scientists) to annotate data for months, while annotations for many Earth-related tasks can be done via crowd-sourcing. Also, a few datasets are imbalanced, but we did not balance the data forcefully because that would end up having only a few samples; for this part, we expect the user to adapt the data augmentation techniques.

\textbf{Dataset Splits} All datasets in \dataset{} include standardized train, validation, and test splits to facilitate consistent and reproducible evaluation. For datasets that did not originally include predefined splits, we generated them following standard practices. When original splits were available, we preserved them to maintain alignment with prior work. These splits ensure that future methods can be compared fairly and under consistent evaluation settings.

\textbf{Cross-Domain Dataset Partitioning} In some cases, we partition datasets based on attributes such as sensor type, data modality, task category, or mission origin. This design choice allows users to analyze model performance across domain shifts, e.g., evaluating cross-sensor or cross-mission generalization by isolating specific factors. Rather than aggregating data into a single dataset, separating them enables experiments in which scientists are often interested, such as how a model trained on one sensor performs on data from another. A more detailed discussion of these partitioning strategies is provided in Appendix \ref{subsec:naming_convention}.

\textbf{Permissive License} All datasets included in \dataset{} have permissive licenses allowing their re-use in the benchmark. We release the \dataset{} version of all datasets with a Creative Commons Attribution 4.0 (CC BY 4.0) license, permitting open access and use.

% In some datasets, we separate the dataset either by sensor, data type, task, or mission. The reason behind this is that instead of having data altogether, keeping it separate helps users to do analysis of their proposed method on the cross-sensor or cross-mission, etc. because scientists are often interested in how the model performs on different domains when it is trained only on one model (in our case, sensor, mission and data type). Detailed discussion provided in Appendix.

\subsection{Tasks and Datasets}
\label{subsec:tasks_and_datasets}

\dataset{} offers a diverse collection of 20 datasets spanning three task categories: classification, segmentation, and object detection. Within these categories, the benchmark supports several subtasks, i.e., classification includes binary, multi-class, and multi-label settings, while segmentation includes both binary and multi-class settings. These tasks are constructed from two primary sources of observation: orbiters (satellites) and surface rovers. In total, the benchmark integrates data from 2 Mars orbiters, 3 rovers, and 6 distinct imaging sensors.

The benchmark covers a wide range of scientifically relevant geologic features that are of high interest to the planetary science community and have been extensively studied in prior literature. \dataset{} was co-developed with expert planetary scientists to ensure its relevance to Mars science. The datasets include geologic features such as boulders, cones, craters, landslides, dust devils, frost, and atmospheric dust. Additionally, multi-class datasets have diverse classes, such as terrain-related classes (e.g., soil, sand, rock, bedrock), landmark-specific features (e.g., Swiss cheese terrain, spiders, dark dunes), and surface-related elements (e.g., ground, ridges, rover tracks), as well as rover components (e.g., inlet, dust removal tool, scoop). This diversity highlights the breadth of \dataset{} in terms of task design, sensor modalities, and variety in geologic features. See Appendix \ref{subsec:details_of_marsbench} for a detailed description and illustrative examples from each dataset.

Unlike EO datasets in which many classes, such as airports or farmland, can be annotated at scale via crowd-sourcing, Mars science datasets often require annotation by domain experts in planetary science or geology. This process is highly specialized and time-consuming, sometimes taking months to years for high-quality labeling. As a result, as shown in Table~\ref{tab:marsbench}, several datasets in \dataset{} are relatively small in size. By including these small-data tasks, \dataset{} provides a valuable testbed for research on label-limited scenarios.

\subsection{Using the Dataset}
\label{subsec:using_the_dataset}

\textbf{Availability} All datasets included in \dataset{} will be publicly released through both Hugging Face Datasets\footnote{\href{https://huggingface.co/collections/Mirali33/mars-bench}{huggingface.co/collections/Mirali33/mars-bench}} and Zenodo\footnote{\href{https://zenodo.org/communities/mars-bench/records}{zenodo.org/communities/mars-bench/records}}. Each dataset follows a standardized schema and is accompanied by metadata, documentation, and loading scripts to enable easy integration into ML pipelines.

\textbf{Target Audience} \dataset{} offers a diverse set of benchmarks designed to evaluate and compare the performance of foundation models for Mars-related tasks. It serves researchers developing models for planetary applications as well as those interested in the geologic features and data types represented in \dataset{}. \dataset{} is also designed to support the broader computer vision and machine learning communities. Researchers studying distribution shift, generalization, or domain adaptation can benefit from its coverage of underrepresented, real-world geospatial scenarios; similar in spirit to WILDS \cite{koh2021wilds}. By offering datasets with unique imaging conditions and semantics, \dataset{} enables research beyond planetary science.

% While \dataset{} is specifically motivated by Mars science applications,

% it is also designed to serve the broader computer vision and machine learning communities. Researchers interested in evaluating model performance across diverse domains, including those studying distribution shift, generalization, and domain adaptation, can benefit from \dataset{}'s coverage of underrepresented, real-world geospatial scenarios, such as WILDS \cite{koh2021wilds}. By offering datasets with unique imaging conditions and semantic content, we aim to support a wide range of research directions beyond planetary science.

\textbf{Baseline Models} In addition to datasets and code, we release baseline models for each dataset included in \dataset{}. We will release the models that currently achieve the best performance on their respective datasets. By making these models publicly available, we aim to lower the barrier for applied research. For example, researchers seeking to generate global maps of features such as cones or craters can use our pre-trained models to produce initial predictions, which can then be refined by domain experts with minimal annotation effort.

% These models have been trained and validated using standardized protocols and serve as reference points for future work.

\textbf{Software Tools}
To promote reproducibility and facilitate future research, we release an open-source toolkit that encapsulates the complete \dataset{} experimental pipeline \footnote{\href{https://github.com/kerner-lab/Mars-Bench}{github.com/kerner-lab/Mars-Bench}}. The repository includes configuration files and executable scripts that reproduce every experiment reported in this study, while permitting users to vary model architectures, hyperparameters, and data partitions with minimal effort. In addition, the toolkit provides utilities for loading datasets and visualizing both objective metrics and qualitative results at the task level as well as in aggregate.

% Additionally, the toolkit provides utilities for loading datasets and visualizing both objective metrics and qualitative results at the task level as well as in aggregate.

% thereby allowing researchers to concentrate on scientific analysis rather than boiler-plate infrastructure.

% To facilitate the usage of the benchmark, we provide a collection of tools for various parts of the experimental pipeline as part of the open-source codebase. We provide a detailed pipeline that includes all the experiments performed in this paper, i.e., providing options for hyperparameters, models. Our pipeline also includes tools for loading datasets, visualizing task-wise and aggregated results, and visualizing results.

\section{Experiments}
\label{sec:experiments}

\textbf{Model Selection} For each task category, we select well-established and widely adopted model architectures representative of current best practices. For classification tasks, we evaluate ResNet101 \cite{he2016deep}, SqueezeNet1.1 \cite{iandola2016squeezenet}, InceptionV3 \cite{szegedy2016rethinking}, Swin Transformer (SwinV2-B) \cite{liu2021swin}, and Vision Transformer (ViT-L/16) \cite{dosovitskiy2020image} architectures. For segmentation, we use U-Net \cite{ronneberger2015u}, DeepLabV3+ \cite{chen2017rethinking}, SegFormer \cite{xie2021segformer}, and Dense Prediction Transformer (DPT) \cite{ranftl2021vision}architectures. For object detection, we evaluate YOLO11 \cite{redmon2016you}, SSD \cite{liu2016ssd}, RetinaNet \cite{lin2017focal}, and Faster R-CNN \cite{ren2015faster}.

% Unlike GEO-Bench, which focuses on varying model depth within a single architecture family, we instead prioritize diversity across architecture types. This choice allows us to observe how different design paradigms, such as convolutional versus transformer-based models, perform on Mars-specific datasets. This decision is further motivated by findings from prior work \cite{kaplan2020scaling}, which suggest that larger versions of a model may improve performance only if sufficient training data is available.

\textbf{Training Settings}
We analyze model performance under three different training strategies: (1) training from scratch with randomly initialized weights, (2) using a pre-trained model as a frozen feature extractor, and (3) full fine-tuning of pre-trained models with all weights trainable. As noted in Section \ref{sec:introduction}, no existing foundation model has been trained specifically for Mars tasks. Therefore, we use models pre-trained on large-scale datasets such as ImageNet (for classification and segmentation) or COCO (for detection) as initialization for transfer learning or feature extraction.

\textbf{Hyperparameter Tuning}
Since the performance of deep learning models is often sensitive to hyperparameter choices, we conducted a grid search over several hyperparameter configurations for each model, task, and training type combination. The best-performing setting was selected based on early stopping criteria applied to validation metrics. All hyperparameter ranges and selected values for each configuration are detailed in Appendix \ref{subsec:pipeline_and_hyperparameters} to ensure reproducibility.

\subsection{Reporting Results}
\label{subsec:reporting_results}

We adopt an identical methodology to \cite{agarwal2021deep} and \cite{lacoste2023geo} to present our results derived from thousands of experiments. Our objective is to report both task-specific outcomes and aggregated results across all tasks with reliable confidence intervals as recommended by \cite{agarwal2021deep}. Specifically, for each combination of model, dataset, and training strategy, we first conduct hyperparameter tuning to identify the optimal settings. Subsequently, we retrain each combination using the selected hyperparameters on seven distinct random seeds, since prior work indicates that results based on only 3–5 random seeds may not be sufficiently robust \cite{agarwal2021deep}. We follow the exact evaluation and reporting methodology as in \cite{agarwal2021deep} and \cite{lacoste2023geo}, including IQM computation, bootstrapped confidence intervals, and normalization; detailed reporting setup and metrics are provided in Appendix \ref{subsec:reporting_results_appendix}.

\section{Results and Analysis}
\label{sec:analysis}

In this section, we present baseline results for all three tasks. We structure our analysis around key research questions, which are addressed in the subsections below.

\subsection{Which model architecture performs best on Mars science tasks, when pre-trained on natural images?}
\label{subsec:analysis_1}

\begin{figure*}[h!]
    \centering
    \includegraphics[width=0.99\textwidth]{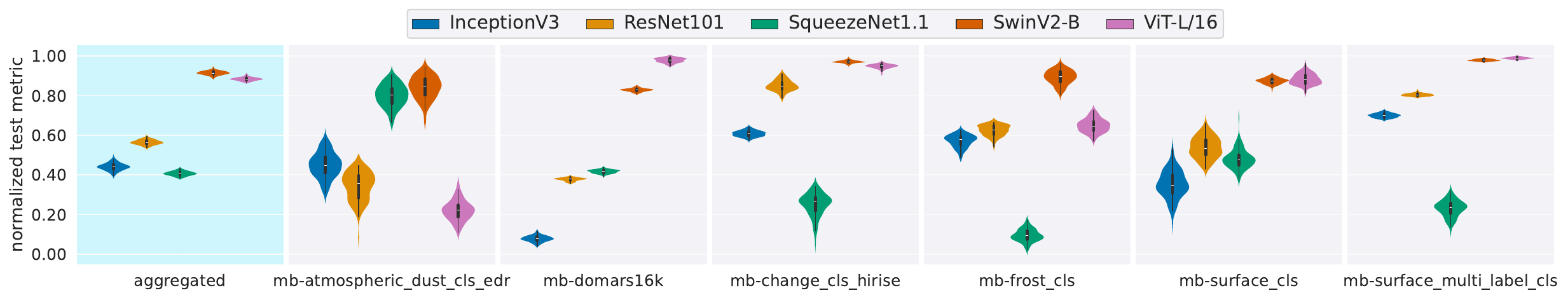}
    \caption{\textbf{Classification Benchmark under Feature Extraction setting:} Normalized F1-score of all baselines across six datasets (higher the better). Aggregated plot shows the average over all datasets.}
    \label{fig:analysis_1_classification}
\end{figure*}

\begin{figure*}[h!]
    \centering
    \includegraphics[width=0.99\textwidth]{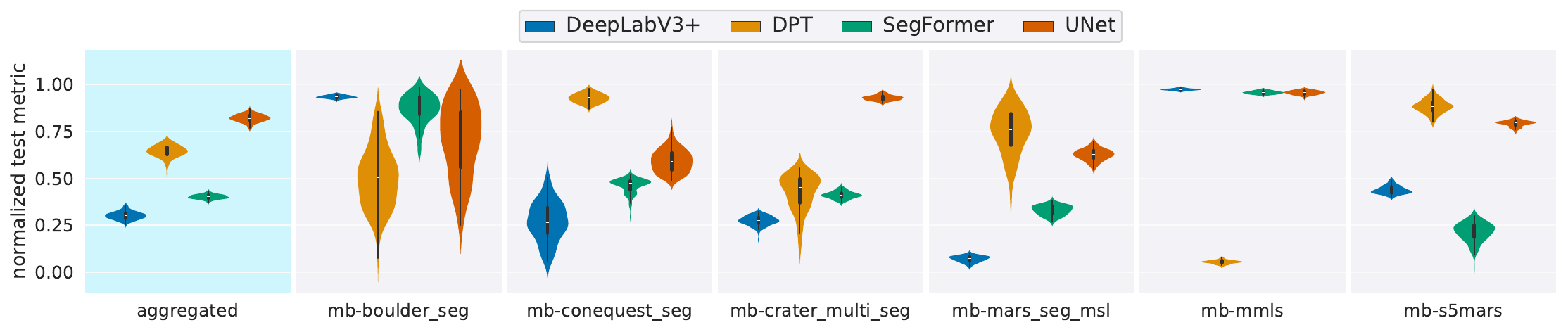}
    \caption{\textbf{Segmentation Benchmark under Feature Extraction setting:} Normalized IoU of all baselines across six datasets (higher the better). Aggregated plot shows the average over all datasets.}
    \label{fig:analysis_1_segmentation}
\end{figure*}

\begin{figure}
    \centering
    \includegraphics[width=0.99\textwidth]{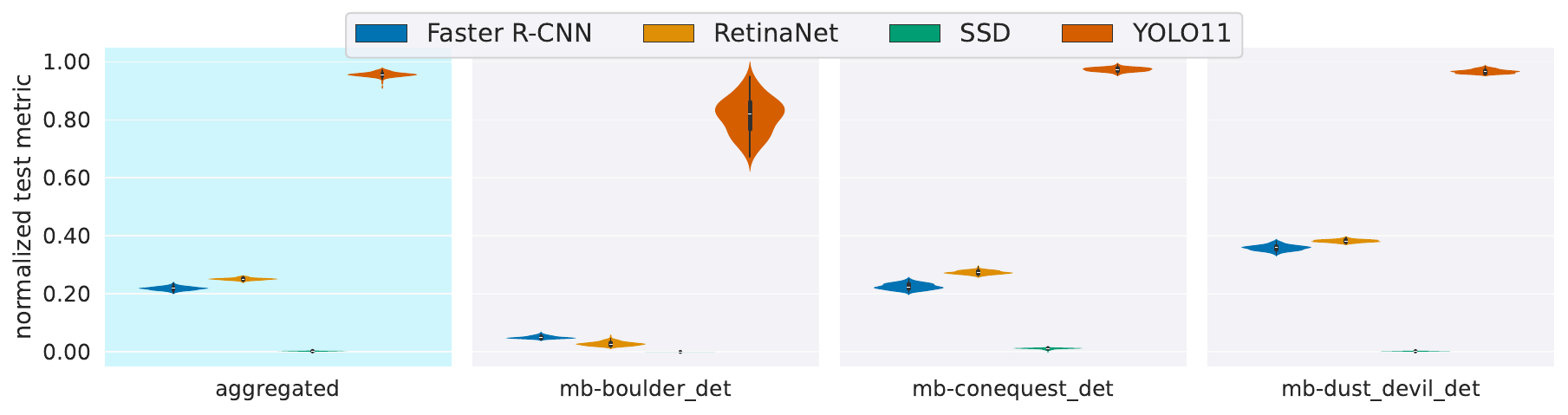}
    \caption{\textbf{Object Detection Benchmark under Feature Extraction setting:} Normalized mAP of all baselines across three datasets (higher is better). Aggregated plot shows the average over all datasets.}
    \label{fig:analysis_1_detection}
\end{figure}

Figures \ref{fig:analysis_1_classification}, \ref{fig:analysis_1_segmentation}, and \ref{fig:analysis_1_detection} show the bootstrapped IQM of normalized performance metric (as defined in Section \ref{subsec:reporting_results}) across six classification, six segmentation, and all three object detection datasets and one training strategy (feature extraction with frozen backbone), along with aggregated results. We report the F1-score for classification tasks, IoU for segmentation tasks, and mAP for object detection tasks. For classification and segmentation, the datasets are selected in a way that ensures a diverse set of geologic features. For example, if two datasets cover the same feature type (e.g., landmarks), we report results for only one of them. Additional results, including those for alternative training regimes and other datasets, are reported in Appendix \ref{sec:extended_results} for all datasets spanning the three tasks..

In classification tasks, SqueezeNet1.1 consistently underperforms relative to other architectures, likely due to its small parameter count. In contrast, ViT-L/16 and SwinV2-B Transformer exhibit competitive performance, with both showing strong generalization across datasets. Notably, some models display narrower confidence intervals than others, suggesting they are more stable and better suited to specific tasks.

For segmentation, U-Net achieves the highest overall performance despite having a relatively wide confidence interval in some datasets. It outperforms both transformer-based models (SegFormer and DPT) on nearly all datasets as well as in aggregate metrics. The DPT model, in particular, shows highly unstable results with large confidence intervals, making it less reliable. These results suggest that, despite its simplicity, U-Net remains a strong baseline for segmentation tasks in Mars science applications.

For object detection, YOLO11 shows the best performance for all three datasets and even in aggregated results. Detection performance is particularly weak on mb-boulder\_det and mb-dust\_devil\_det.

These challenges are primarily due to several factors:

\begin{itemize}
    \item The overall dataset size is significantly smaller for all three object detection datasets compared to several classification and segmentation datasets.
    \item The number of objects per image is low, with many images containing only one or even zero target objects.
    \item The grayscale nature of the imagery limits visual cues, and low object–background contrast (e.g., in dust devil detection) further complicates learning.
\end{itemize}

\subsection{What is the effect of training set size on the performance of each model?}
\label{subsec:analysis_2}

\begin{figure*}[h!]
    \centering
    \includegraphics[width=0.99\textwidth]{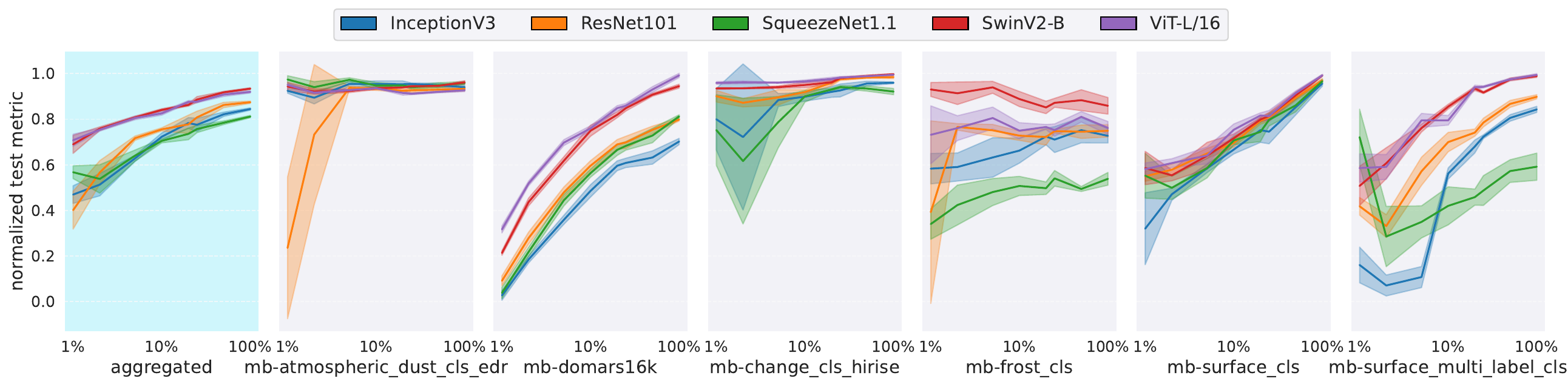}
    \caption{\textbf{Classification vs Train size:} Normalized F1-score of baselines with a growing size (from 1\% to 100\%) of the training set. Shaded regions indicate confidence intervals over multiple runs.}
    \label{fig:analysis_2_classification}
\end{figure*}

\begin{figure*}[h!]
    \centering
    \includegraphics[width=0.99\textwidth]{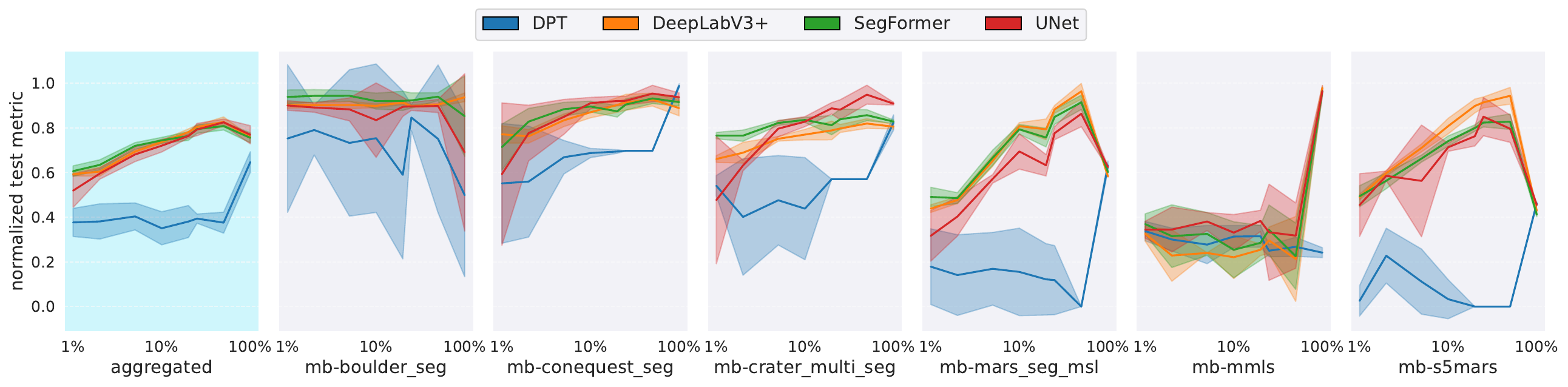}
    \caption{\textbf{Segmentation vs Train size:} Normalized IoU of baselines with a growing size (from 1\% to 100\%) of the training set. Shaded regions indicate confidence intervals over multiple runs.}
    \label{fig:analysis_2_segmentation}
\end{figure*}

To assess how training set size impacts model performance, we conducted experiments by varying the amount of labeled training data. Specifically, we trained each model using 1\%, 2\%, 5\%, 10\%, 20\%, 25\%, 50\%, and 100\% of the available training data, while keeping the validation and test sets fixed. For each configuration, we performed multiple runs and report the average normalized test metric, as shown in Figures \ref{fig:analysis_2_classification} and \ref{fig:analysis_2_segmentation}.

From the aggregated results, we observe a consistent trend: increasing the training set size generally leads to improved performance in both classification and segmentation tasks. However, dataset-level analysis reveals that the rate of improvement and error margins vary significantly depending on the model and dataset. This shows the differing levels of difficulty among datasets in \dataset, highlighting the benchmark's overall challenge. 

In classification, transformer-based models such as SwinV2-B and ViT-L/16 consistently outperform smaller convolutional models like SqueezeNet1.1. In contrast, for segmentation tasks, U-Net outperforms transformer-based models such as DPT and SegFormer across most training sizes. DPT not only shows lower overall performance but also exhibits high variance across runs, as reflected in wide confidence intervals.

\subsection{How do models that are trained for EO tasks perform on \dataset?}
\label{subsec:analysis_3}

\begin{wrapfigure}{r}{0.7\linewidth}
    \begin{center}
        \includegraphics[width=0.7\textwidth]{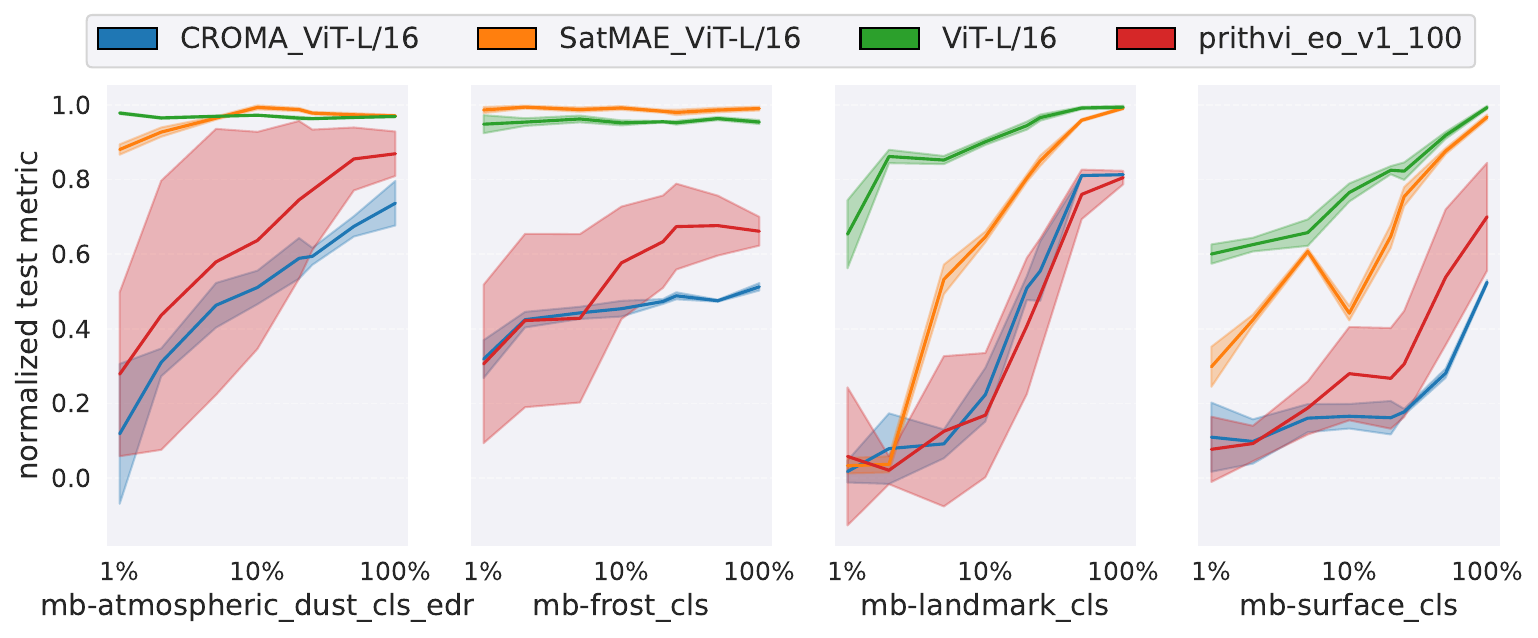}
        \caption{\textbf{Classification vs Train size for EO baselines:} Normalized F1-score with a growing size (from 1\% to 100\%) of the training set. Shaded regions indicate confidence intervals over multiple runs.}
        % \vspace{-2mm}
        \label{fig:analysis_3}
    \end{center}
\end{wrapfigure}

Although there are no published foundation models for Mars orbital or surface imagery, there are many foundation models for Earth orbital imagery. To assess cross-domain generalization, we evaluated foundation models pre-trained on EO data. Specifically, SatMAE \cite{reed2023scale}, CROMA \cite{fuller2024croma}, and Prithvi \cite{jakubik2023prithvi} on selected \dataset{} classification tasks (see Appendix \ref{subsec:eo_experiments_details} for experimental details). These models were originally trained on Earth satellite data that vary in geography, scale, and semantics but share the overhead imaging perspective found in many Mars datasets. We compare them to a ViT-L/16 model pre-trained on ImageNet to establish a general-domain baseline (Figure \ref{fig:analysis_3}).

Although EO pre-trained models performed well on all datasets, the ImageNet pre-trained ViT performed better. One possible explanation is that although ViT is pre-trained on natural images and EO models are pre-trained on satellite data, ViT is pre-trained on 14 million images, while SatMAE, CROMA, and Prithvi are pre-trained on 1 million or less than 1 million images. Additionally, diversity in ImageNet, because as discussed in the literature, diversity and/or geographical coverage of pre-training data can affect the performance of the model \cite{entezari2023role, nguyen2022quality, purohit2025how, ramanujan2023connection}. Among EO foundation models, the Prithvi model in particular consistently showed low performance and large error bars. All these results show that, despite EO models pre-trained on satellite data, Earth and Mars orbital imagery differ significantly in ways that likely impact model transferability. For instance, Martian imagery lacks vegetation, water bodies, and human-made structures, which are common in EO datasets. Additionally, Mars exhibits unique geological formations, color distributions, and atmospheric conditions that are totally different than Earth imagery. These domain gaps suggest that while EO-pretrained models can offer a reasonable starting point, foundation models specifically trained on Mars data are likely to yield more robust and generalizable performance for Martian tasks.

% Although EO pre-trained models performed well on all datasets, the ImageNet pre-trained ViT performed better. The Prithvi model in particular consistently showed low performance and large error bars. These results show that ... although it looks like one pre-trained EO model can help to generalize, but although model is pre-trained on satellite data, there is 

% I would say something about how the Earth and Mars orbital imagery domains might seem similar but they are actually very different (briefly list some differences), so it's likely customized Mars orbital foundation models would perform better

% 3
% EO pre-trained (maybe SatMAE)
% feature extractor

% Earth models do not work for Mars science tasks

% select few datasets --> landmark_classfication, 
% CTX --> 100k
% HiRISE --> 100K
% THEMIS -> 100k

\subsection{How do proprietary VLMs, such as Gemini and GPT, perform on \dataset?}
\label{subsec:analysis_4}

With the rapid advancement of vision-language models (VLMs), such as Gemini \cite{team2023gemini} and GPT \cite{brown2020language}, there is increasing interest in evaluating their effectiveness beyond general-purpose tasks. These models, trained on diverse multimodal datasets, have demonstrated strong performance on various open-domain vision benchmarks with minimal supervision. However, their applicability to Mars science, has not been explored. Evaluating VLMs on \dataset{} provides valuable insight into their ability to generalize to planetary science tasks without domain-specific fine-tuning. 

We focused on evaluating the reasoning capabilities of these models by explicitly prompting them with context-rich instructions, rather than relying solely on direct answer generation. We used the Gemini 2.0 Flash and GPT-4o Mini models, both from their May 2025 checkpoints.

\begin{wraptable}{r}{0.62\linewidth}
    \centering
    \resizebox{\linewidth}{!}{
        \begin{tabular}{c|cc|cc}
            \toprule[1.5pt]
            \multirow{2}{*}{Task}          & \multicolumn{2}{c|}{Gemini} & \multicolumn{2}{c}{GPT} \\
                                           & Accuracy     & F1-score     & Accuracy   & F1-score   \\
            \midrule[1pt]
            mb-domars16k                   & 0.34       & 0.32       & 0.36     & 0.30     \\
            mb-surface\_cls                & 0.43       & 0.44       & 0.42     & 0.41     \\
            mb-frost\_cls                  & 0.50       & 0.55       & 0.43     & 0.54     \\
            mb-atmospheric\_dust\_cls\_edr & 0.43       & 0.50       & 0.68     & 0.56     \\
            \midrule
            mb-crater\_multi\_seg          & 0.37       & 0.41       & 0.49     & 0.51     \\
            mb-mars\_seg\_msl              & 0.86       & 0.84       & 0.79     & 0.70     \\
            \bottomrule
        \end{tabular}
    }
    \caption{Performance of Gemini and GPT on \dataset{}.}
    \label{tab:vlm_results}
\end{wraptable}

% Performance of Gemini and GPT on six \dataset{} datasets.

% Classification and segmentation tasks are evaluated using a prompting-based approach with class definitions provided via system instructions.

% \begin{table*}[t]
% \centering
% \resizebox{\linewidth}{!}{
%     \begin{tabular}{ccccccccc|cccc}
%        \toprule[1.5pt]
%        & \multicolumn{8}{c|}{\textbf{Classification}} & \multicolumn{4}{c}{\textbf{Segmentation}} \\
%        \midrule
%        & \multicolumn{2}{c}{mb-domars16k} & \multicolumn{2}{c}{mb-surface\_classification} & \multicolumn{2}{c}{mb-frost\_classification} & \multicolumn{2}{c|}{mb-atmospheric\_dust\_classification\_edr} & \multicolumn{2}{c}{mb-crater\_multi\_segmentation} & \multicolumn{2}{c}{mb-mars\_seg\_msl} \\
%        % \midrule
%        & Accuracy & F1-Score & Accuracy & F1-Score & Accuracy & F1-Score & Accuracy & F1-Score & Accuracy & F1-Score & Accuracy & F1-Score \\
%        \midrule[1pt]
%        \textbf{Gemini} & 0.3440 & 0.3202 & 0.4286 & 0.4409 & 0.4940 & 0.5458 & 0.4277 & 0.5027 & 0.3736 & 0.4120 & 0.8606 & 0.8395 \\
%        \textbf{GPT} & 0.3627 & 0.2971  & 0.4177 & 0.4057 & 0.4280 & 0.5402 & 0.6814 & 0.5571 & 0.4929 & 0.5094 & 0.7874 & 0.6955 \\
%        \bottomrule
%     \end{tabular}
% }
% \caption{VLM results}
% \label{tab:vlm_results}
% \end{table*}

We selected six \dataset{} datasets spanning classification and segmentation tasks. The selected tasks cover a range of geologic features to evaluate how well the models generalize across different scientific concepts. From each dataset, we randomly sampled 500 test images, ensuring the label distribution in the sampled subset matched that of the original dataset. This sample size was chosen to balance evaluation fidelity with the computational cost associated with API-based model usage, particularly for GPT. We reformulated segmentation as a multi-label classification task. For both classification and segmentation, we provided system instructions defining each class and prompted the models to predict the relevant classes for each image. Full prompts and system instructions for all tasks are included in Appendix \ref{sec:prompts}.

% For classification tasks, we provided system instructions defining each class and asked the models to predict the class for each image. For segmentation, we reformulated the problem as a multi-label classification task by extracting all class labels present in the ground-truth mask. Instructions included definitions for all relevant classes, and models were prompted to list all classes they could identify in the image. Full prompts and system instructions for all tasks are included in the Appendix.

Both Gemini and GPT achieved reasonable performance on some tasks, but their results are inconsistent across datasets (Table~\ref{tab:vlm_results}). Notably, both models perform well on the \texttt{mb-mars\_seg\_msl} dataset, achieving an F1-score of 0.84 (Gemini) and 0.70 (GPT). This dataset involves terrain segmentation with classes such as sand, rock, and sky, classes that are also common in natural images and likely well-represented in the models’ pre-training data. In contrast, performance drops significantly on datasets such as \texttt{mb-crater\_multi\_seg} and \texttt{mb-domars16k}, which require identification of fine-grained geologic structures like crater types and Martian landmarks.

With this, we also conducted experiments on smaller vision-language models (CLIP \cite{radford2021learning}, SigLIP \cite{tschannen2025siglip}, and SmolVLM \cite{marafioti2025smolvlm}), and these models also show similar trends observed for Gemini and GPT (see Appendix \ref{sec:vlm_evaluation} for details). Our results suggest that current VLMs lack sufficient specialized knowledge for accurate interpretation. As noted in Section \ref{subsec:tasks_and_datasets}, many of these tasks demand domain expertise. These findings highlight the gap between general-purpose vision-language capabilities and the needs of Mars science, further reinforcing the importance of domain-specific model development.

\section{Research opportunities}
\label{sec:research_opportunity}

% Research Oppr

% what do these results show? -->
% 1) We need some in-domain pre-trained models
% 2) Identifying fine-grained objects/classes
% 3) Highly imbalanced (distribution of classes makes harder get good enough performance)

\dataset{} provides valuable research opportunities, not only for the planetary science and remote sensing communities but also for the broader machine learning and computer vision community. \dataset{} creates the following key research opportunities:

\begin{itemize}[leftmargin=*]
    \item \dataset{} will accelerate the development of foundation models specifically tailored to Mars orbital and surface-related tasks by facilitating a systematic evaluation of model performance. It provides essential infrastructure for benchmarking diverse models within a unified framework, mirroring the influential role benchmarks have historically played in other specialized domains.
    \item The benchmark comprises several challenging datasets that introduce unique complexities to computer vision tasks. For instance, dust devil detection is particularly challenging due to the subtle contrast differences between dust devils and the Martian terrain. ConeQuest presents difficulties stemming from significant visual variability among cones collected from various Martian regions, challenging models to generalize across high intra-class variance. In addition, many datasets included in \dataset{} are small-scale and highly imbalanced, i.e., mb-change\_cls\_ctx, mb-boulder\_seg, and mb-boulder\_det.
    \item \dataset{} significantly expands research opportunities focused on addressing distribution shifts and out-of-distribution generalization. These challenges are closely aligned with contemporary methodological advancements such as those proposed by \cite{ilharco2022editing, wortsman2022model, koh2021wilds, von2020neural, cole2022does, ramanujan2024connection, entezari2023role, miller2021accuracy, learningdataset, taori2020measuring, chen2023understanding, humblot2024noisy, shi2023effective, naganuma2023empirical}, which emphasize robust model evaluation across diverse domains to enhance real-world applicability and to advance understanding of model robustness, generalization, and failure modes when exposed to out-of-distribution (OOD) data.
\end{itemize}

\section{Conclusion}
\label{sec:conclusion}

We introduced the first benchmark for evaluating models on a wide range of Mars science tasks using both orbital and surface imagery. \dataset{} standardizes diverse datasets into a unified, machine-learning-ready format and provides code for fine-tuning and evaluating across classification, segmentation, and object detection tasks. Datasets in \dataset{} also include a wide variety of geologic features that have been extensively studied in the literature and remain of high interest to the scientific community. We believe that \dataset{} will drive the development of Mars-specific foundation models, improve generalization across planetary tasks, and open new research directions in planetary science and beyond.

\paragraph{Limitations} A key limitation of \dataset{} is the absence of georeferencing for most datasets. This arises from the fact that the original sources of these datasets do not provide spatial metadata (e.g., latitude and longitude coordinates), mapping the samples to the Martian surface. As a result, it is currently not possible to assess the spatial distribution or coverage of \dataset{} across different regions of Mars. Lack of georeferencing is a known challenge in remote sensing benchmarks, as it restricts the ability to conduct spatial analysis or regional generalization studies. There are a few exceptions within the \dataset{} collection that include geolocation information. For instance, the ConeQuest dataset already provides georeferenced samples, and we retain this metadata in our release. Additionally, both crater segmentation datasets (binary and multi-class) were prepared by us from scratch, and therefore also include geolocation metadata. Both crater datasets are derived from the THEMIS sensor; however, the current version is based on an older THEMIS release from 2010. A newer version of the THEMIS dataset (released in 2017) \cite{hill2017well} is now available and can be utilized in the future to generate updated versions of these two crater segmentation datasets.

% There few datasets exceptions in \dataset{} which include geolocation information, i.e., ConeQuest already include georeferencing samples and we retrain this metadata in the our release. In addition, as both crater segmentation datasets (binary and multi-class) are prepared by us from scratch, we did include geolocation information. Apart from this, both crater datasets are created from THEMIS sensor. However, the current version is created from the older version of THEMIS from 2010, whereas new version of THEMIS is available from 207 which can be compiled to create more up-to-date version for these 2 crater datasets.

% Additionally, we did not explore techniques to address class imbalance in datasets, such as re-sampling or loss reweighting. Investigating methods to handle imbalance and its effect on model performance remains an important direction for future work.

\textbf{Acknowledgment:} Part of this research was carried out at the Jet Propulsion Laboratory, California Institute of Technology, under a contract with the National Aeronautics and Space Administration. We acknowledge Mihir Parmar for his support with the proprietary VLM experiments. We also thank Shrey Malvi and Hitansh Shah for their initial assistance in building and curating the Crater datasets.

% We extend our gratitude to the Research Computing (RC) at ASU for providing computing resources for experiments.

\newpage
\bibliographystyle{plain}
\bibliography{references}

%%%%%%%%%%%%%%%%%%%%%%%%%%%%%%%%%%%%%%%%%%%%%%%%%%%%%%%%%%%%

\newpage
\section*{NeurIPS Paper Checklist}

\begin{enumerate}

\item {\bf Claims}
    \item[] Question: Do the main claims made in the abstract and introduction accurately reflect the paper's contributions and scope?
    \item[] Answer: \answerYes{} %, \answerNo{}, or \answerNA{}.
    \item[] Justification: See Section \ref{sec:analysis}. % \justificationTODO{}
    % \item[] Guidelines:
    % \begin{itemize}
    %     \item The answer NA means that the abstract and introduction do not include the claims made in the paper.
    %     \item The abstract and/or introduction should clearly state the claims made, including the contributions made in the paper and important assumptions and limitations. A No or NA answer to this question will not be perceived well by the reviewers. 
    %     \item The claims made should match theoretical and experimental results, and reflect how much the results can be expected to generalize to other settings. 
    %     \item It is fine to include aspirational goals as motivation as long as it is clear that these goals are not attained by the paper. 
    % \end{itemize}

\item {\bf Limitations}
    \item[] Question: Does the paper discuss the limitations of the work performed by the authors?
    \item[] Answer: \answerYes{}
    \item[] Justification: See Section \ref{sec:conclusion}

\item {\bf Theory assumptions and proofs}
    \item[] Question: For each theoretical result, does the paper provide the full set of assumptions and a complete (and correct) proof?
    \item[] Answer: \answerNA{}.
    % \item[] Justification: \justificationTODO{}
    % \item[] Guidelines:
    % \begin{itemize}
    %     \item The answer NA means that the paper does not include theoretical results. 
    %     \item All the theorems, formulas, and proofs in the paper should be numbered and cross-referenced.
    %     \item All assumptions should be clearly stated or referenced in the statement of any theorems.
    %     \item The proofs can either appear in the main paper or the supplemental material, but if they appear in the supplemental material, the authors are encouraged to provide a short proof sketch to provide intuition. 
    %     \item Inversely, any informal proof provided in the core of the paper should be complemented by formal proofs provided in appendix or supplemental material.
    %     \item Theorems and Lemmas that the proof relies upon should be properly referenced. 
    % \end{itemize}

    \item {\bf Experimental result reproducibility}
    \item[] Question: Does the paper fully disclose all the information needed to reproduce the main experimental results of the paper to the extent that it affects the main claims and/or conclusions of the paper (regardless of whether the code and data are provided or not)?
    \item[] Answer: \answerYes{}
    \item[] Justification: See Section \ref{sec:experiments}.

\item {\bf Open access to data and code}
    \item[] Question: Does the paper provide open access to the data and code, with sufficient instructions to faithfully reproduce the main experimental results, as described in supplemental material?
    \item[] Answer: \answerYes{}
    % \item[] Justification: \justificationTODO{}
    % \item[] Guidelines:
    % \begin{itemize}
    %     \item The answer NA means that paper does not include experiments requiring code.
    %     \item Please see the NeurIPS code and data submission guidelines (\url{https://nips.cc/public/guides/CodeSubmissionPolicy}) for more details.
    %     \item While we encourage the release of code and data, we understand that this might not be possible, so “No” is an acceptable answer. Papers cannot be rejected simply for not including code, unless this is central to the contribution (e.g., for a new open-source benchmark).
    %     \item The instructions should contain the exact command and environment needed to run to reproduce the results. See the NeurIPS code and data submission guidelines (\url{https://nips.cc/public/guides/CodeSubmissionPolicy}) for more details.
    %     \item The authors should provide instructions on data access and preparation, including how to access the raw data, preprocessed data, intermediate data, and generated data, etc.
    %     \item The authors should provide scripts to reproduce all experimental results for the new proposed method and baselines. If only a subset of experiments are reproducible, they should state which ones are omitted from the script and why.
    %     \item At submission time, to preserve anonymity, the authors should release anonymized versions (if applicable).
    %     \item Providing as much information as possible in supplemental material (appended to the paper) is recommended, but including URLs to data and code is permitted.
    % \end{itemize}

\item {\bf Experimental setting/details}
    \item[] Question: Does the paper specify all the training and test details (e.g., data splits, hyperparameters, how they were chosen, type of optimizer, etc.) necessary to understand the results?
    \item[] Answer: \answerYes{}
    \item[] Justification: See Appendix \ref{sec:experimental_setup}.
    % \item[] Guidelines:
    % \begin{itemize}
    %     \item The answer NA means that the paper does not include experiments.
    %     \item The experimental setting should be presented in the core of the paper to a level of detail that is necessary to appreciate the results and make sense of them.
    %     \item The full details can be provided either with the code, in appendix, or as supplemental material.
    % \end{itemize}

\item {\bf Experiment statistical significance}
    \item[] Question: Does the paper report error bars suitably and correctly defined or other appropriate information about the statistical significance of the experiments?
    \item[] Answer: \answerYes{}
    \item[] Justification: See Section \ref{sec:analysis}
    % \item[] Guidelines:
    % \begin{itemize}
    %     \item The answer NA means that the paper does not include experiments.
    %     \item The authors should answer "Yes" if the results are accompanied by error bars, confidence intervals, or statistical significance tests, at least for the experiments that support the main claims of the paper.
    %     \item The factors of variability that the error bars are capturing should be clearly stated (for example, train/test split, initialization, random drawing of some parameter, or overall run with given experimental conditions).
    %     \item The method for calculating the error bars should be explained (closed form formula, call to a library function, bootstrap, etc.)
    %     \item The assumptions made should be given (e.g., Normally distributed errors).
    %     \item It should be clear whether the error bar is the standard deviation or the standard error of the mean.
    %     \item It is OK to report 1-sigma error bars, but one should state it. The authors should preferably report a 2-sigma error bar than state that they have a 96\% CI, if the hypothesis of Normality of errors is not verified.
    %     \item For asymmetric distributions, the authors should be careful not to show in tables or figures symmetric error bars that would yield results that are out of range (e.g. negative error rates).
    %     \item If error bars are reported in tables or plots, The authors should explain in the text how they were calculated and reference the corresponding figures or tables in the text.
    % \end{itemize}

\item {\bf Experiments compute resources}
    \item[] Question: For each experiment, does the paper provide sufficient information on the computer resources (type of compute workers, memory, time of execution) needed to reproduce the experiments?
    \item[] Answer: \answerYes{}
    \item[] Justification: See Appendix \ref{sec:experimental_setup}.
    % \item[] Guidelines:
    % \begin{itemize}
    %     \item The answer NA means that the paper does not include experiments.
    %     \item The paper should indicate the type of compute workers CPU or GPU, internal cluster, or cloud provider, including relevant memory and storage.
    %     \item The paper should provide the amount of compute required for each of the individual experimental runs as well as estimate the total compute. 
    %     \item The paper should disclose whether the full research project required more compute than the experiments reported in the paper (e.g., preliminary or failed experiments that didn't make it into the paper). 
    % \end{itemize}

\item {\bf Code of ethics}
    \item[] Question: Does the research conducted in the paper conform, in every respect, with the NeurIPS Code of Ethics \url{https://neurips.cc/public/EthicsGuidelines}?
    \item[] Answer: \answerYes{}
    % \item[] Justification: \justificationTODO{}
    % \item[] Guidelines:
    % \begin{itemize}
    %     \item The answer NA means that the authors have not reviewed the NeurIPS Code of Ethics.
    %     \item If the authors answer No, they should explain the special circumstances that require a deviation from the Code of Ethics.
    %     \item The authors should make sure to preserve anonymity (e.g., if there is a special consideration due to laws or regulations in their jurisdiction).
    % \end{itemize}

\item {\bf Broader impacts}
    \item[] Question: Does the paper discuss both potential positive societal impacts and negative societal impacts of the work performed?
    \item[] Answer: \answerYes{}
    \item[] Justification: See Appendix \ref{sec:societal_impact}.

\item {\bf Safeguards}
    \item[] Question: Does the paper describe safeguards that have been put in place for responsible release of data or models that have a high risk for misuse (e.g., pretrained language models, image generators, or scraped datasets)?
    \item[] Answer: \answerNA{}
    % \item[] Justification: \justificationTODO{}
    % \item[] Guidelines:
    % \begin{itemize}
    %     \item The answer NA means that the paper poses no such risks.
    %     \item Released models that have a high risk for misuse or dual-use should be released with necessary safeguards to allow for controlled use of the model, for example by requiring that users adhere to usage guidelines or restrictions to access the model or implementing safety filters. 
    %     \item Datasets that have been scraped from the Internet could pose safety risks. The authors should describe how they avoided releasing unsafe images.
    %     \item We recognize that providing effective safeguards is challenging, and many papers do not require this, but we encourage authors to take this into account and make a best faith effort.
    % \end{itemize}

\item {\bf Licenses for existing assets}
    \item[] Question: Are the creators or original owners of assets (e.g., code, data, models), used in the paper, properly credited and are the license and terms of use explicitly mentioned and properly respected?
    \item[] Answer: \answerYes{}
    \item[] Justification: See Section \ref{subsec:design_process}
    % \item[] Guidelines:
    % \begin{itemize}
    %     \item The answer NA means that the paper does not use existing assets.
    %     \item The authors should cite the original paper that produced the code package or dataset.
    %     \item The authors should state which version of the asset is used and, if possible, include a URL.
    %     \item The name of the license (e.g., CC-BY 4.0) should be included for each asset.
    %     \item For scraped data from a particular source (e.g., website), the copyright and terms of service of that source should be provided.
    %     \item If assets are released, the license, copyright information, and terms of use in the package should be provided. For popular datasets, \url{paperswithcode.com/datasets} has curated licenses for some datasets. Their licensing guide can help determine the license of a dataset.
    %     \item For existing datasets that are re-packaged, both the original license and the license of the derived asset (if it has changed) should be provided.
    %     \item If this information is not available online, the authors are encouraged to reach out to the asset's creators.
    % \end{itemize}

\item {\bf New assets}
    \item[] Question: Are new assets introduced in the paper well documented and is the documentation provided alongside the assets?
    \item[] Answer: \answerYes{}
    \item[] Justification: See Section \ref{subsec:design_process}
    % \item[] Guidelines:
    % \begin{itemize}
    %     \item The answer NA means that the paper does not release new assets.
    %     \item Researchers should communicate the details of the dataset/code/model as part of their submissions via structured templates. This includes details about training, license, limitations, etc. 
    %     \item The paper should discuss whether and how consent was obtained from people whose asset is used.
    %     \item At submission time, remember to anonymize your assets (if applicable). You can either create an anonymized URL or include an anonymized zip file.
    % \end{itemize}

\item {\bf Crowdsourcing and research with human subjects}
    \item[] Question: For crowdsourcing experiments and research with human subjects, does the paper include the full text of instructions given to participants and screenshots, if applicable, as well as details about compensation (if any)? 
    \item[] Answer: \answerNA{}
    % \item[] Justification: \justificationTODO{}
    % \item[] Guidelines:
    % \begin{itemize}
    %     \item The answer NA means that the paper does not involve crowdsourcing nor research with human subjects.
    %     \item Including this information in the supplemental material is fine, but if the main contribution of the paper involves human subjects, then as much detail as possible should be included in the main paper. 
    %     \item According to the NeurIPS Code of Ethics, workers involved in data collection, curation, or other labor should be paid at least the minimum wage in the country of the data collector. 
    % \end{itemize}

\item {\bf Institutional review board (IRB) approvals or equivalent for research with human subjects}
    \item[] Question: Does the paper describe potential risks incurred by study participants, whether such risks were disclosed to the subjects, and whether Institutional Review Board (IRB) approvals (or an equivalent approval/review based on the requirements of your country or institution) were obtained?
    \item[] Answer: \answerNA{}
    % \item[] Justification: \justificationTODO{}
    % \item[] Guidelines:
    % \begin{itemize}
    %     \item The answer NA means that the paper does not involve crowdsourcing nor research with human subjects.
    %     \item Depending on the country in which research is conducted, IRB approval (or equivalent) may be required for any human subjects research. If you obtained IRB approval, you should clearly state this in the paper. 
    %     \item We recognize that the procedures for this may vary significantly between institutions and locations, and we expect authors to adhere to the NeurIPS Code of Ethics and the guidelines for their institution. 
    %     \item For initial submissions, do not include any information that would break anonymity (if applicable), such as the institution conducting the review.
    % \end{itemize}

\item {\bf Declaration of LLM usage}
    \item[] Question: Does the paper describe the usage of LLMs if it is an important, original, or non-standard component of the core methods in this research? Note that if the LLM is used only for writing, editing, or formatting purposes and does not impact the core methodology, scientific rigorousness, or originality of the research, declaration is not required.
    %this research? 
    \item[] Answer: \answerNA{}
    % \item[] Justification: \justificationTODO{}
    % \item[] Guidelines:
    % \begin{itemize}
    %     \item The answer NA means that the core method development in this research does not involve LLMs as any important, original, or non-standard components.
    %     \item Please refer to our LLM policy (\url{https://neurips.cc/Conferences/2025/LLM}) for what should or should not be described.
    % \end{itemize}

\end{enumerate}

\newpage

\appendix

\clearpage

% Create appendix-only TOC
\begingroup
  \renewcommand{\contentsname}{Appendix Contents}
  \setcounter{tocdepth}{3}
  \startcontents[appendix]       % Start collecting appendix entries
  \section*{Appendix}
  \printcontents[appendix]{}{1}{}% Print the appendix-only TOC
\endgroup

\clearpage

\section{\dataset{} Resources}
\label{sec:marsbench_resources}

\begin{itemize}%[noitemsep]
    \item Project Page - \href{https://mars-bench.github.io/}{mars-bench.github.io/}
    \item HuggingFace - \href{https://huggingface.co/collections/Mirali33/mars-bench}{huggingface.co/collections/Mirali33/mars-bench}
    \item Zenodo - \href{https://zenodo.org/communities/mars-bench/records}{zenodo.org/communities/mars-bench/records}
    \item Github - \href{https://github.com/kerner-lab/Mars-Bench}{github.com/kerner-lab/Mars-Bench}
    \item LeaderBoard - \href{https://huggingface.co/spaces/Mirali33/Mars-Bench}{huggingface.co/spaces/Mirali33/Mars-Bench}
    \item Baseline Models - \href{https://huggingface.co/collections/Mirali33/mars-bench-models}{huggingface.co/collections/Mirali33/mars-bench-models}
\end{itemize}

\section{\dataset{} Details}
\label{sec:appendix_marsbench}

% Alongside the Hugging Face release, we also release \dataset{} available via Zenodo\footnote{\url{https://zenodo.org/communities/mars-bench/records}}.
This appendix provides detailed documentation of the \dataset{} benchmark. We describe the dataset naming conventions used (\ref{subsec:naming_convention}), task details of each of the 20 datasets included (\ref{subsec:details_of_marsbench}), our process for preparing few-shot and partitioned versions (\ref{subsec:few_shot_partition}), corrections and improvements made to original datasets with expert input (\ref{subsec:correction}), and finally, a list of relevant datasets that were excluded from this release and the reasons for their exclusion (\ref{subsec:removed_datasets}).

\subsection{Dataset Naming Convention}
\label{subsec:naming_convention}

For datasets with well-established names in the original literature (which are DoMars16k, ConeQuest, MMLS, Mars-Seg, S5Mars), we retain the original names and prepend the prefix “mb-”. For other datasets, we adopt a consistent naming scheme based on geologic feature (e.g., atmospheric\_dust), followed by task type (e.g., \texttt{cls} for classification), and optionally by sensor or data source (e.g., \texttt{edr}) as defined in Section 3.1. The suffixes \texttt{cls}, \texttt{seg}, and \texttt{det} indicate classification, segmentation, and object detection tasks, respectively. This naming scheme allows for easy categorization and filtering within the benchmark.

% For the datasets which already has unique name provided/defined in the original paper, we use those names only, i.e., domars16k, conequest, mmls, mars\_seg, s5mars. We still apply prefic "mb-" in all datasets and separate them with mission, instruments, or task type wherever required as defined in Section \ref{subsec:design_process}.

% For the datasets where actual dataset name is not defined, the way we define name of the datasets are we write the geologic feature name (e.g., atmospheric\_dust) followed by task type always (e.g., cls), and followed by instrument/mission name (e.g., edr) if required.

% Here, cls, det, and seg stand for classification, segmentation, and detection.

\subsection{Dataset Descriptions}
\label{subsec:details_of_marsbench}

This section provides brief descriptions of the 20 datasets included in \dataset{}, including their task type, targeted geologic features, class structure, and observation modality.

\subsubsection{Classification}
\label{subsubsec:classification}

In all classification datasets, the data is organized in a structured format. Each split, train, validation, and test contains subfolders corresponding to each class specific to that dataset. Additionally, we provide an \texttt{annotation.csv} file containing metadata for every sample, including the following fields: \texttt{file\_id} (a unique identifier for each sample), \texttt{split} (indicating the data partition), \texttt{feature\_name} (a 3-letter acronym representing the class), and \texttt{label} (the numerical class ID). To aid interpretability, each dataset also includes a \texttt{mapping.json} file that maps both the class IDs and acronyms to their corresponding full feature names.

\begin{figure}[htbp]
  \centering

  \begin{subfigure}[b]{0.48\textwidth}
    \centering
    \includegraphics[width=0.40\linewidth]{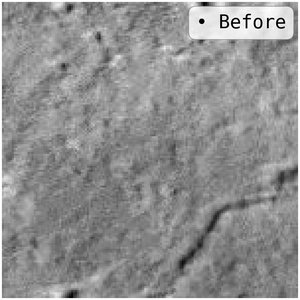}
    \includegraphics[width=0.40\linewidth]{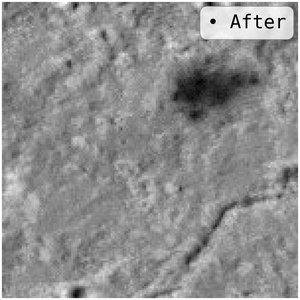}
    \caption{Change}
    \label{fig:change_ctx}
  \end{subfigure}
  % \hfill
  \hspace{-10mm}
  \begin{subfigure}[b]{0.48\textwidth}
    \centering
    \includegraphics[width=0.40\linewidth]{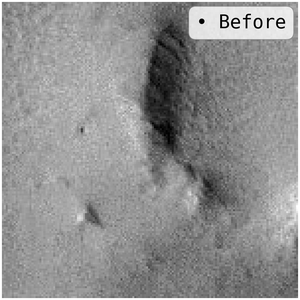}
    \includegraphics[width=0.40\linewidth]{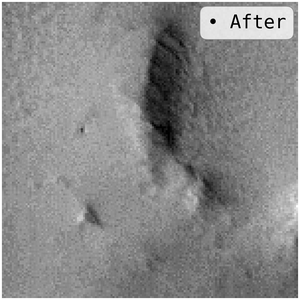}
    \caption{No change}
    \label{fig:no_change_ctx}
  \end{subfigure}

  \caption{mb-change\_cls\_ctx}
  \label{fig:change_cls_ctx}
\end{figure}

\begin{figure}[htbp]
  \centering

  \begin{subfigure}[b]{0.48\textwidth}
    \centering
    \includegraphics[width=0.40\linewidth]{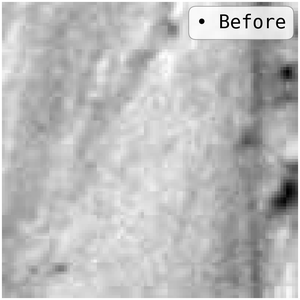}
    \includegraphics[width=0.40\linewidth]{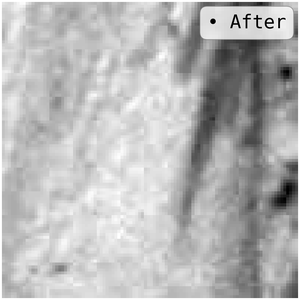}
    \caption{Change}
    \label{fig:change_hirise}
  \end{subfigure}
  % \hfill
  \hspace{-10mm}
  \begin{subfigure}[b]{0.48\textwidth}
    \centering
    \includegraphics[width=0.40\linewidth]{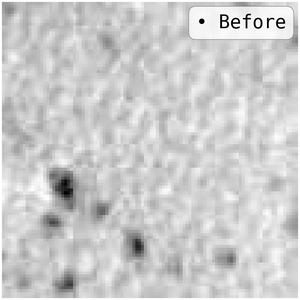}
    \includegraphics[width=0.40\linewidth]{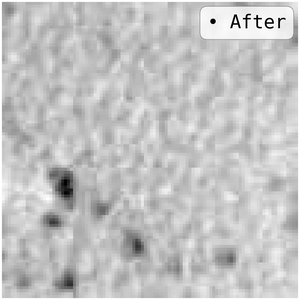}
    \caption{No change}
    \label{fig:no_change_hirise}
  \end{subfigure}

  \caption{mb-change\_cls\_hirise}
  \label{fig:change_cls_hirise}
\end{figure}

\paragraph{mb-change\_cls\_ctx and mb-change\_cls\_hirise} These datasets are designed for binary classification of surface changes using temporal image pairs; specifically, one image taken before and another after some time period, from the \textit{same} Martian location. The task involves identifying whether meaningful surface change has occurred and classifying between ``\textbf{Change}'' and ``\textbf{No change}''. The dataset includes two versions based on different sensors: CTX and HiRISE. Unlike standard single-image classification, this task requires forming a composite input from two grayscale images (Figure \ref{fig:change_cls_ctx} and \ref{fig:change_cls_hirise}). Following the approach outlined by Kerner et. al. \cite{kerner2019toward}, we adopt the composite grayscale method: the blue channel encodes the ``before'' image, the green channel encodes the ``after'' image, and the red channel is set to zero. \texttt{mb-change\_cls\_ctx} is the smallest dataset included in \dataset{}, in terms of the number of samples. Since the original datasets do not provide standard splits, we generated consistent train, validation, and test sets for both CTX and HiRISE versions.

% This is binary classification where model is provided with 2 images from 2 different timestamp (labeled as \_before and \_after) but from the same region on Mars. It has 2 versions from 2 different sensors, i.e., CTX and HiRISE. This is one of the unique task as it involves creating a sample for model from 2 different images. The way original paper approaches this problem is by handling the 2 images from different timestamp in a different way, e.g., by taking signed different, absolute difference, using composite grayscale, using features from trained autoencoders. For all the experiments, we have done in this paper, we have used composite grayscale approach where, we created a composite of the before and after grayscale images in which the blue channel contains the before image, the green channel contains the after image, and the red channel contains all zeros. Both CTX and HiRISE version of the data do not provide the train, val, and test split, so, we created it here.

\newpage

\begin{figure}[htbp]
  \centering

  \begin{subfigure}[b]{0.48\textwidth}
    \centering
    \includegraphics[width=0.40\linewidth]{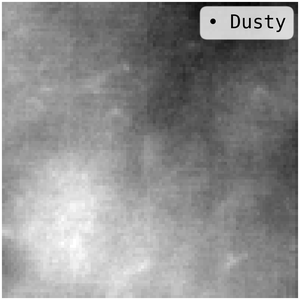}
    \includegraphics[width=0.40\linewidth]{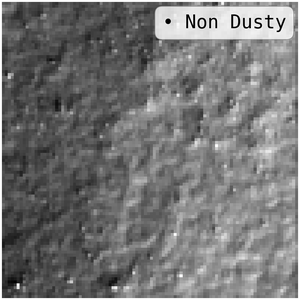}
    \caption{mb-atmospheric\_dust\_cls\_edr}
    \label{fig:atmospheric_dust_cls_edr}
  \end{subfigure}
  % \hfill
  \hspace{-10mm}
  \begin{subfigure}[b]{0.48\textwidth}
    \centering
    \includegraphics[width=0.40\linewidth]{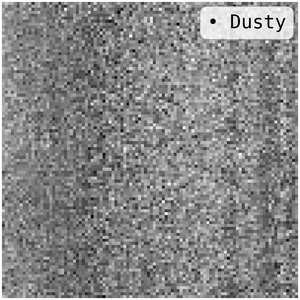}
    \includegraphics[width=0.40\linewidth]{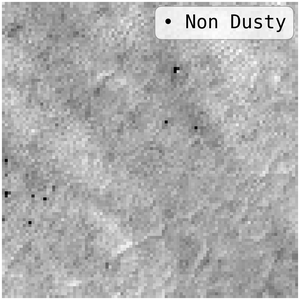}
    \caption{mb-atmospheric\_dust\_cls\_rdr}
    \label{fig:atmospheric_dust_cls_rdr}
  \end{subfigure}

  \caption{mb-atmospheric\_dust\_cls datasets}
  \label{fig:atmospheric_dust_cls}
\end{figure}

\paragraph{mb-atmospheric\_dust\_cls\_edr \& mb-atmospheric\_dust\_cls\_rdr} 
These are binary classification tasks focused on classifying between ``\textbf{Dusty}'' and ``\textbf{Non dusty}'' (Figure \ref{fig:atmospheric_dust_cls}) regions in Mars surface imagery captured by the HiRISE camera on the Mars Reconnaissance Orbiter. The \texttt{EDR} (Experimental Data Record) refers to raw images from the instrument that have not been calibrated or stitched together; while the \texttt{RDR} (Reduced Data Record) is a downsampled or processed version of the EDR, typically used for quick viewing or initial analysis. Both versions are balanced in terms of class distribution and come with predefined train, validation, and test splits.

% This is binary classification task which distinguish between "dusty" and "not dusty" regions in Mars' surface images captured by the HiRISE. Here, EDR means Experimental Data Record which refers raw images from the instrument that have not been calibrated or stitched together. RDR means Reduced Data Record which is a downsampled or processed version of the EDR, typically used for quick viewing or initial analysis. Both these versions of originally provided train, val, and test splits.

\begin{figure}[htbp]

  \centering
  \begin{subfigure}{\columnwidth}
    \centering
    \includegraphics[width=0.18\columnwidth]{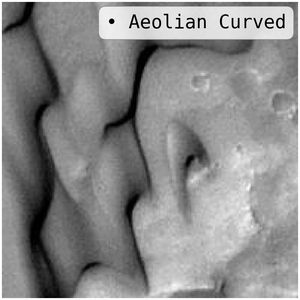}
    \includegraphics[width=0.18\columnwidth]{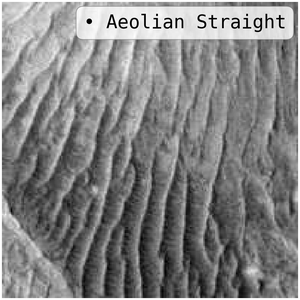}
    \includegraphics[width=0.18\columnwidth]{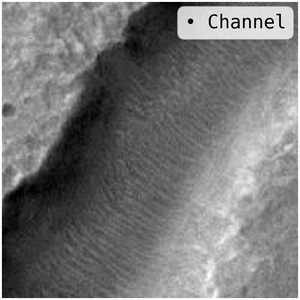}
    \includegraphics[width=0.18\columnwidth]{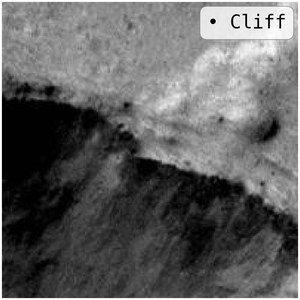}
    \includegraphics[width=0.18\columnwidth]{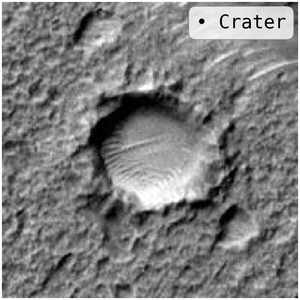}
  \end{subfigure}

  \centering
  \begin{subfigure}{\columnwidth}
    \centering
    \centering
    \includegraphics[width=0.18\columnwidth]{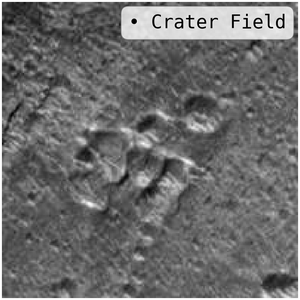}
    \includegraphics[width=0.18\columnwidth]{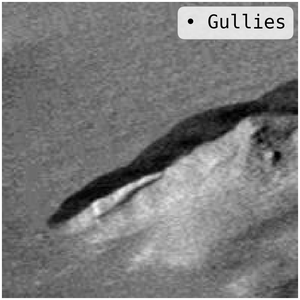}
    \includegraphics[width=0.18\columnwidth]{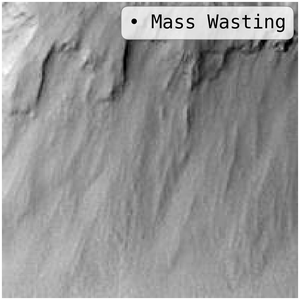}
    \includegraphics[width=0.18\columnwidth]{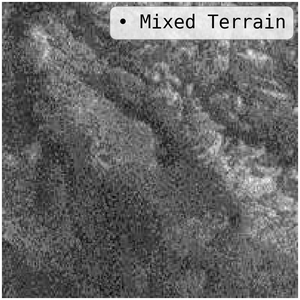}
    \includegraphics[width=0.18\columnwidth]{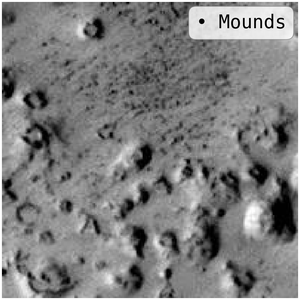}
  \end{subfigure}

  \centering
  \begin{subfigure}{\columnwidth}
    \centering
    \centering
    \includegraphics[width=0.18\columnwidth]{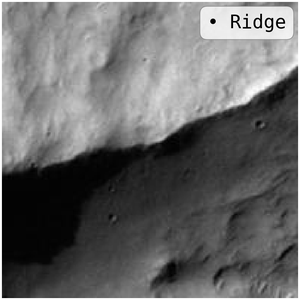}
    \includegraphics[width=0.18\columnwidth]{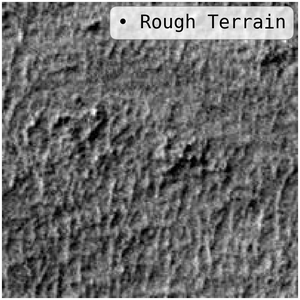}
    \includegraphics[width=0.18\columnwidth]{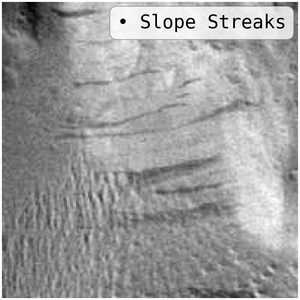}
    \includegraphics[width=0.18\columnwidth]{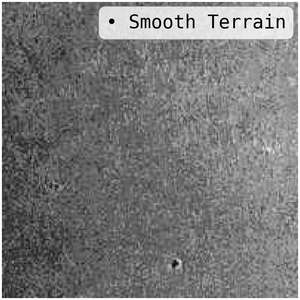}
    \includegraphics[width=0.18\columnwidth]{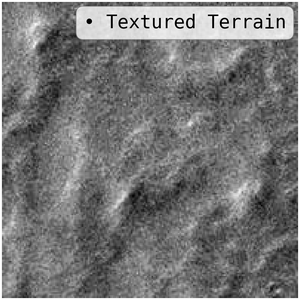}
  \end{subfigure}

  \caption{mb-domars16k}
  \label{fig:domars16k}

\end{figure}

\paragraph{mb-domars16k} This is a multi-class classification dataset designed for geomorphologic feature recognition on Mars using imagery from the CTX sensor. It consists of 15 classes (Figure \ref{fig:domars16k}) grouped into five thematic categories: (1) \textbf{Aeolian Bedforms:} Aeolian Curved, Aeolian Straight; (2) \textbf{Topographic Landforms:} Channel, Cliff, Mounds, Ridge; (3) \textbf{Slope Features:} Gullies, Mass Wasting, Slope Streaks; (4) \textbf{Impact Landforms:} Crater, Crater Field; and (5) \textbf{Basic Terrain:} Mixed Terrain, Rough Terrain, Smooth Terrain, Textured Terrain. This is one of the largest and diverse \textit{orbital} datasets in terms of a number of classes. Hence, the dataset presents a unique challenge due to its class granularity, significant variability within classes, and subtle differences between classes, making it valuable for evaluating models on fine-grained classification and generalization. The original version includes train, validation, and test splits; and the dataset is balanced in terms of class distribution.

% This is a multi-class classification dataset designed for geomorphologic feature recognition on Mars based on the CTX sensor. The dataset contains 15 classes, which are based on 5 thematic groups: Aeolian Bedforms (corresponding classes: Aeolian Curved, Aeolian Straight), Topographic Landforms (corresponding classes: Channel, Cliff, Mounds, Ridge), Slope Feature Landforms (corresponding classes: Gullies, Mass Wasting, Slope Streaks), Impact Landforms (corresponding classes: Crater, Crater Field), and Basic Terrain Landforms (corresponding classes: Mixed Terrain, Rough Terrain, Smooth Terrain, Textured Terrain). This is one of the biggest and diverse orbital datasets in terms of a number of classes. This dataset presents a unique challenge due to its class granularity, significant variability within classes, and subtle differences between classes, making it valuable for evaluating models on fine-grained classification and generalization. The original version includes train, validation, and test splits; and the dataset is balanced in terms of class distribution.

\newpage

\begin{figure}[htbp]

  \centering
  \begin{subfigure}{\columnwidth}
    \centering
    \includegraphics[width=0.2\columnwidth]{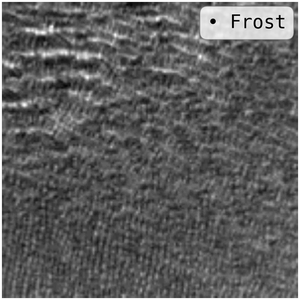}
    \includegraphics[width=0.2\columnwidth]{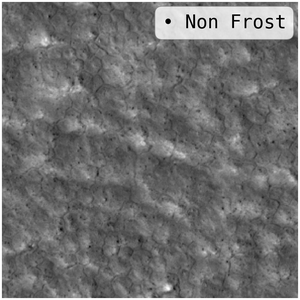}
  \end{subfigure}

  \caption{mb-frost\_cls}
  \label{fig:frost_cls}

\end{figure}

% \cite{doran2024evaluating}
\paragraph{mb-frost\_cls} This is a binary classification dataset designed to detect the presence or absence of surface frost in Mars satellite imagery. The dataset consists of HiRISE image patches labeled as either ``\textbf{Frost}'' or ``\textbf{Non Frost}'' (Figure \ref{fig:frost_cls}). Among all datasets in \dataset{}, this is the largest in terms of the number of samples. The dataset is well-balanced in terms of class distribution and includes predefined train, validation, and test splits, as provided by the original authors.

% and is curated to evaluate how frost detection performance varies across different terrain types. It captures terrain diversity, including plains, craters, and polar regions, making it particularly valuable for analyzing domain-specific model performance and generalization.

\begin{figure}[htbp]

  \centering
  \begin{subfigure}{\columnwidth}
    \centering
    \includegraphics[width=0.15\columnwidth]{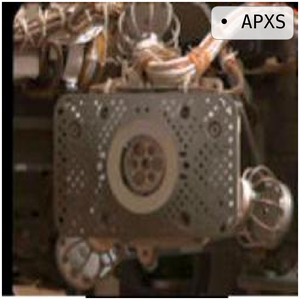}
    \includegraphics[width=0.15\columnwidth]{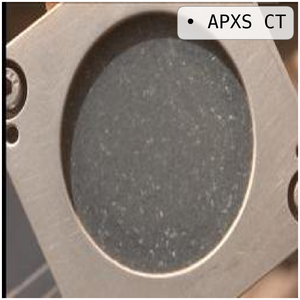}
    \includegraphics[width=0.15\columnwidth]{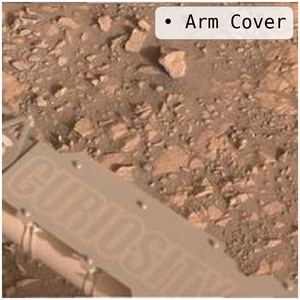}
    \includegraphics[width=0.15\columnwidth]{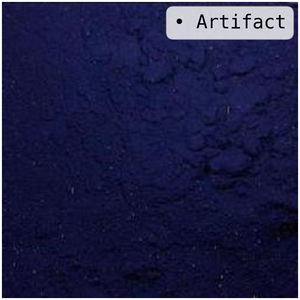}
    \includegraphics[width=0.15\columnwidth]{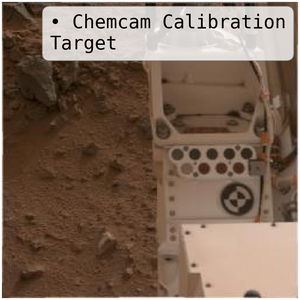}
    \includegraphics[width=0.15\columnwidth]{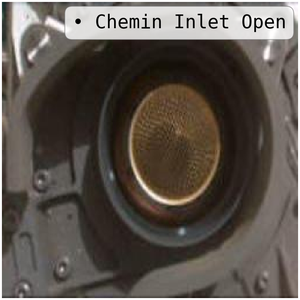}
  \end{subfigure}

  \centering
  \begin{subfigure}{\columnwidth}
    \centering
    \centering
    \includegraphics[width=0.15\columnwidth]{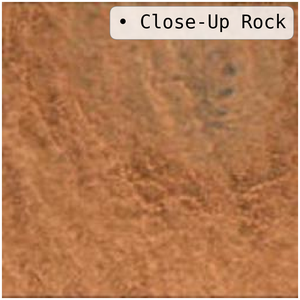}
    \includegraphics[width=0.15\columnwidth]{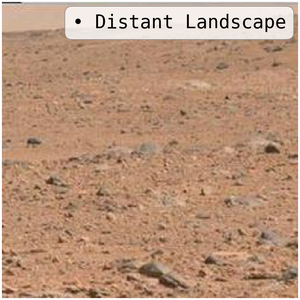}
    \includegraphics[width=0.15\columnwidth]{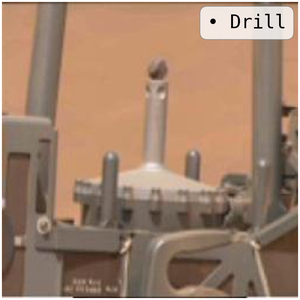}
    \includegraphics[width=0.15\columnwidth]{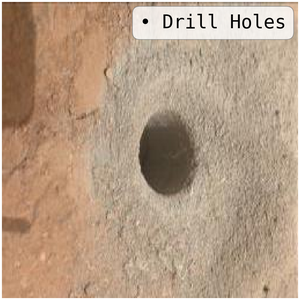}
    \includegraphics[width=0.15\columnwidth]{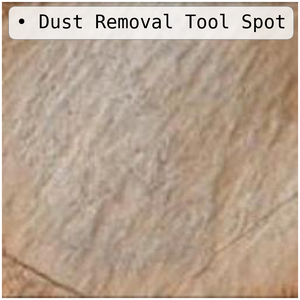}
    \includegraphics[width=0.15\columnwidth]{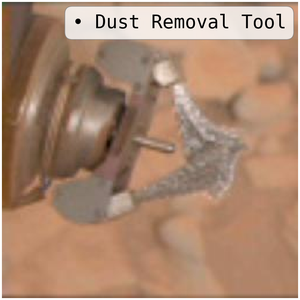}
  \end{subfigure}

  \centering
  \begin{subfigure}{\columnwidth}
    \centering
    \centering
    \includegraphics[width=0.15\columnwidth]{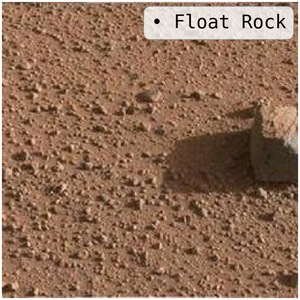}
    \includegraphics[width=0.15\columnwidth]{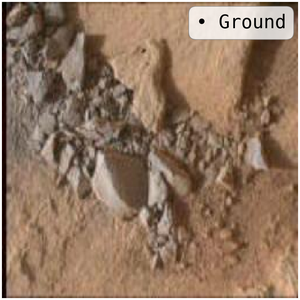}
    \includegraphics[width=0.15\columnwidth]{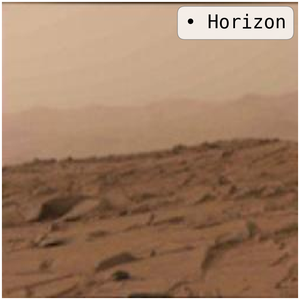}
    \includegraphics[width=0.15\columnwidth]{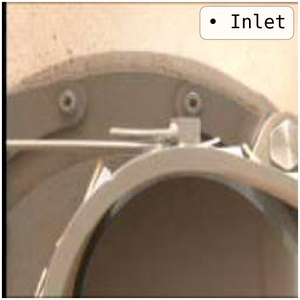}
    \includegraphics[width=0.15\columnwidth]{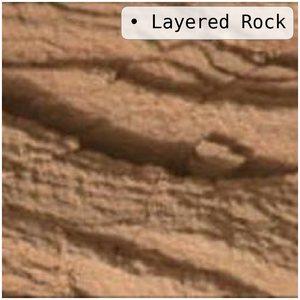}
    \includegraphics[width=0.15\columnwidth]{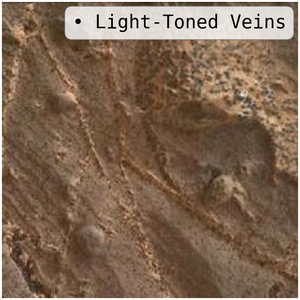}
  \end{subfigure}

  \centering
  \begin{subfigure}{\columnwidth}
    \centering
    \centering
    \includegraphics[width=0.15\columnwidth]{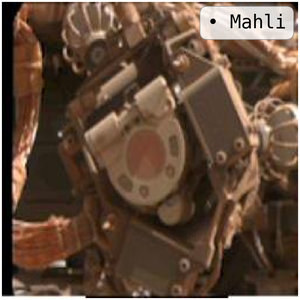}
    \includegraphics[width=0.15\columnwidth]{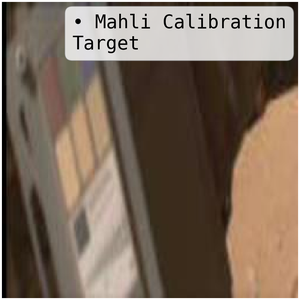}
    \includegraphics[width=0.15\columnwidth]{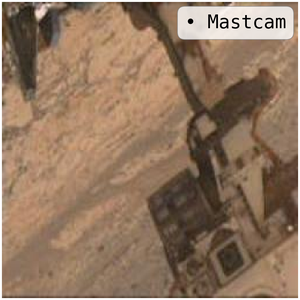}
    \includegraphics[width=0.15\columnwidth]{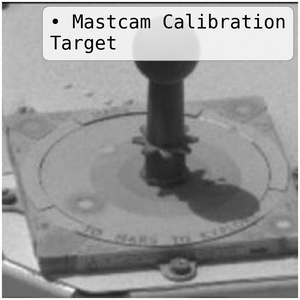}
    \includegraphics[width=0.15\columnwidth]{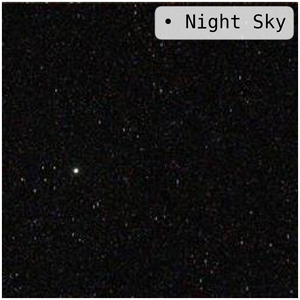}
    \includegraphics[width=0.15\columnwidth]{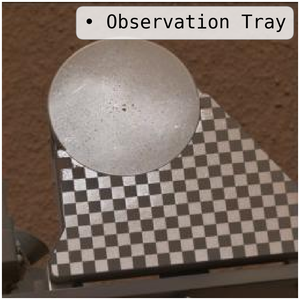}
  \end{subfigure}

  \centering
  \begin{subfigure}{\columnwidth}
    \centering
    \centering
    \includegraphics[width=0.15\columnwidth]{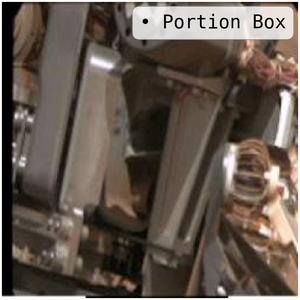}
    \includegraphics[width=0.15\columnwidth]{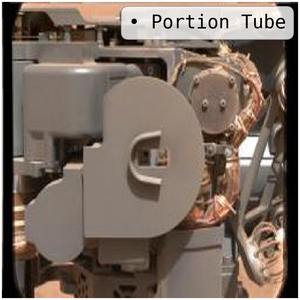}
    \includegraphics[width=0.15\columnwidth]{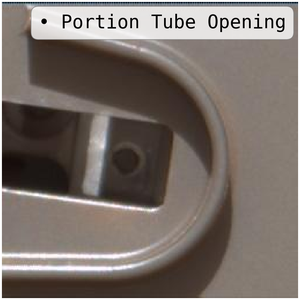}
    \includegraphics[width=0.15\columnwidth]{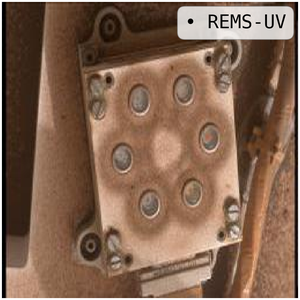}
    \includegraphics[width=0.15\columnwidth]{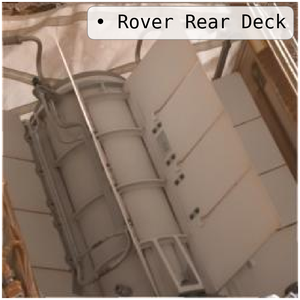}
    \includegraphics[width=0.15\columnwidth]{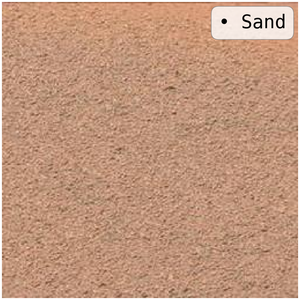}
  \end{subfigure}

  \centering
  \begin{subfigure}{\columnwidth}
    \centering
    \centering
    \includegraphics[width=0.15\columnwidth]{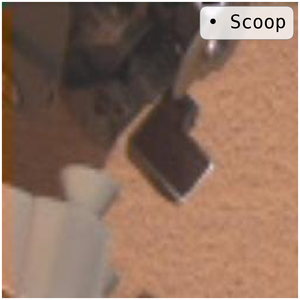}
    \includegraphics[width=0.15\columnwidth]{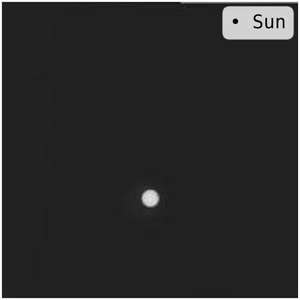}
    \includegraphics[width=0.15\columnwidth]{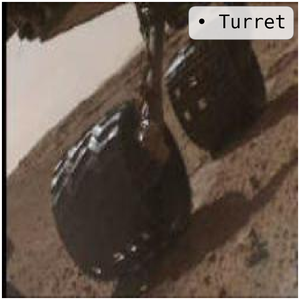}
    \includegraphics[width=0.15\columnwidth]{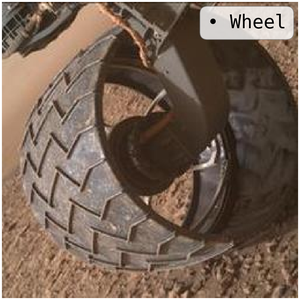}
    \includegraphics[width=0.15\columnwidth]{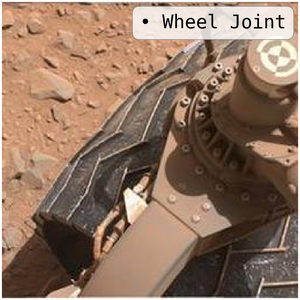}
    \includegraphics[width=0.15\columnwidth]{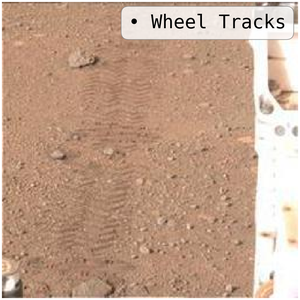}
  \end{subfigure}

  \caption{mb-surface\_cls}
  \label{fig:surface_cls}

\end{figure}

\paragraph{mb-surface\_cls} This is a multi-class classification dataset consisting of surface imagery captured by the Mastcam and MAHLI instruments aboard the Curiosity rover. It comprises 36 classes (Figure \ref{fig:surface_cls}), making it the largest and most diverse surface imagery dataset included in \dataset{}. The dataset was created by combining two previously released versions of the surface classification dataset, following consultation with the original authors (see Section~\ref{subsec:correction} for details).

The classes span a wide range of surface elements, including rover components, scientific instruments, geologic features, and environmental elements. The full list includes: Alpha Particle X-Ray Spectrometer (APXS), APXS Calibration Target, Arm Cover, Artifact, ChemCam Calibration Target, CheMin Inlet Open, Close-Up Rock, Distant Landscape, Drill, Drill Holes, Dust Removal Tool (DRT), DRT Spot, Float Rock, Ground, Horizon, Inlet, Layered Rock, Light-Toned Veins, MAHLI, MAHLI Calibration Target, Mastcam, Mastcam Calibration Target, Night Sky, Observation Tray, Portion Box, Portion Tube, Portion Tube Opening, REMS-UV, Rover Rear Deck, Sand, Scoop, Sun, Turret, Wheel, Wheel Joint, Wheel Tracks.

Images were labeled using the IDAR tool, incorporating annotations from both domain experts and volunteers. For ambiguous samples, class prioritization rules were applied to assign the most representative label. One part of the dataset (\cite{wagstaff2021mars}) includes predefined train, validation, and test splits based on Mars sol ranges. For the earlier version \cite{wagstaff2018deep}, which did not include original splits, we created consistent splits before merging with the newer version. As with many real-world planetary datasets, \texttt{mb-surface\_cls} is highly imbalanced, with a large proportion of samples belonging to the \texttt{Ground} class.

\begin{figure}[htbp]

  \centering
  \begin{subfigure}{\columnwidth}
    \centering
    \includegraphics[width=0.18\columnwidth]{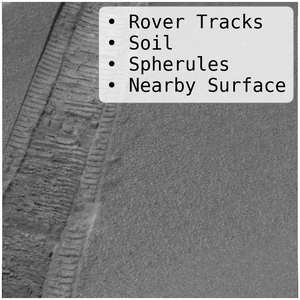}
    \includegraphics[width=0.18\columnwidth]{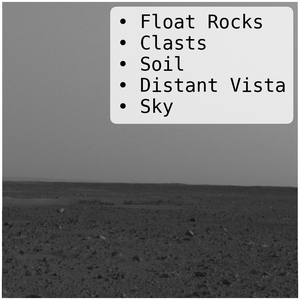}
    \includegraphics[width=0.18\columnwidth]{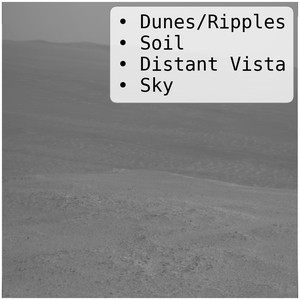}
  \end{subfigure}

  \caption{mb-surface\_multi\_label\_cls}
  \label{fig:surface_multi_label_cls}

\end{figure}

% \cite{cole2020identifying}
\paragraph{mb-surface\_multi\_label\_cls} This is the only multi-label classification dataset in \dataset{}, based on imagery captured by the Pancam instruments aboard the Opportunity and Spirit rovers. Hence, each image in this dataset can be associated with one or multiple labels (Figure \ref{fig:surface_multi_label_cls}). It includes 25 unique classes encompassing a broad range of surface, environmental, and rover-related features. The class list covers geologic and contextual elements such as: RAT Hole, Clasts, Dunes/Ripples, Soil, Rock Outcrops, Close-Up Rock, RAT Brushed Target, Distant Vista, Rover Deck, Bright Soil, Float Rocks, Artifacts, Pancam Calibration Target, Arm Hardware, Round Rock Features, Spherules, Other Hardware, Astronomy, Nearby Surface, Miscellaneous Rocks, Rover Tracks, Sky, Rover Parts, Linear Rock Features, and Soil Trench. The dataset includes pre-defined train, validation, and test splits.

\newpage

\begin{figure}[htbp]

  \centering
  \begin{subfigure}{\columnwidth}
    \centering
    \includegraphics[width=0.18\columnwidth]{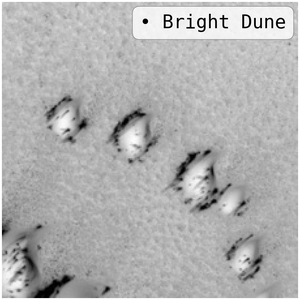}
    \includegraphics[width=0.18\columnwidth]{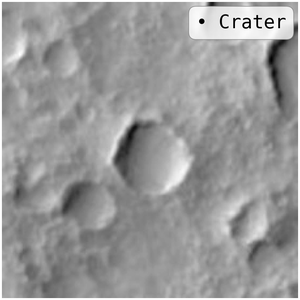}
    \includegraphics[width=0.18\columnwidth]{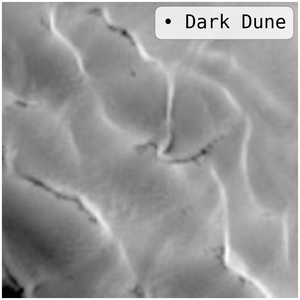}
    \includegraphics[width=0.18\columnwidth]{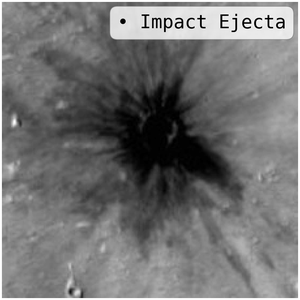}
  \end{subfigure}

  \centering
  \begin{subfigure}{\columnwidth}
    \centering
    \centering
    \includegraphics[width=0.18\columnwidth]{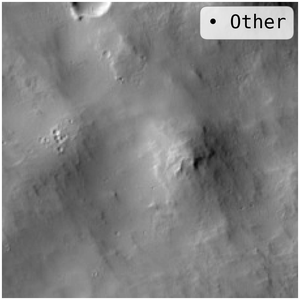}
    \includegraphics[width=0.18\columnwidth]{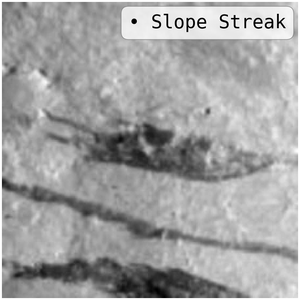}
    \includegraphics[width=0.18\columnwidth]{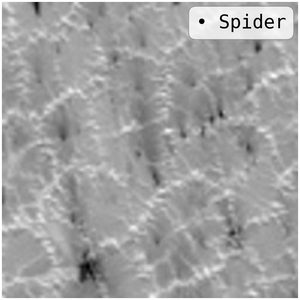}
    \includegraphics[width=0.18\columnwidth]{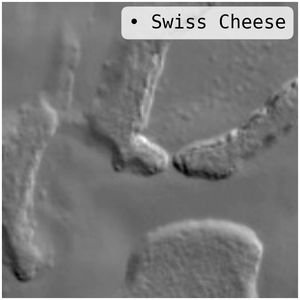}
  \end{subfigure}

  \caption{mb-landmark\_cls}
  \label{fig:landmark_cls}

\end{figure}

% \cite{wagstaff2021mars}
\paragraph{mb-landmark\_cls} This is a multi-class classification dataset derived from orbital HiRISE imagery. It classifies into 8 surface feature classes: \textbf{Bright Dune, Crater, Dark Dune, Impact Ejecta, Slope Streak, Spider, Swiss Cheese}, and \textbf{Other} (Figure \ref{fig:landmark_cls}). The class distribution is highly imbalanced, with “Other” comprising the majority of samples and Impact Ejecta being the minority class. Landmarks were extracted using a dynamic salience-based method. Labels were generated via a mix of volunteer crowdsourcing and expert validation, with additional calibration techniques applied to improve reliability. The dataset includes predefined train, validation, and test splits.

\subsubsection{Segmentation}
\label{subsubsec:segmentation}

All segmentation datasets in \dataset{} are provided as image–mask pairs, where each mask represents the ground truth labels for semantic segmentation. The masks are encoded as single-channel images, with each pixel assigned a discrete class ID. Across all datasets, the class ID \texttt{0} consistently denotes the background. Additionally, we include a \texttt{mapping.json} file with each dataset that specifies the mapping between class IDs and their corresponding semantic class names, ensuring clarity and ease of use for downstream tasks.

\begin{figure}[htbp]
  \centering

  \begin{subfigure}[b]{0.48\textwidth}
    \centering
    \includegraphics[width=0.40\linewidth]{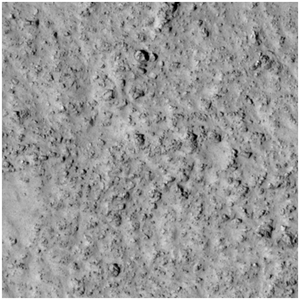}
    \includegraphics[width=0.40\linewidth]{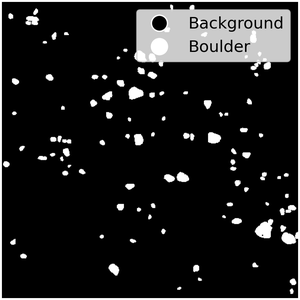}
  \end{subfigure}
  \hspace{-10mm}
  \begin{subfigure}[b]{0.48\textwidth}
    \centering
    \includegraphics[width=0.40\linewidth]{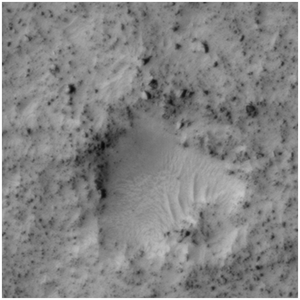}
    \includegraphics[width=0.40\linewidth]{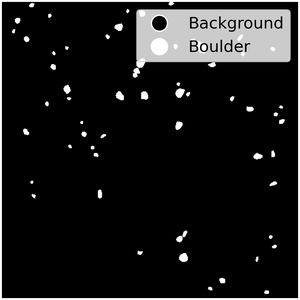}
  \end{subfigure}

  \caption{mb-boulder\_seg}
  \label{fig:boulder_seg}
\end{figure}

% \cite{prieur2023automatic}
\paragraph{mb-boulder\_seg} This is a binary segmentation dataset focused on segmenting boulders on the Martian surface using high-resolution orbital imagery from the HiRISE camera. The dataset comprises manually annotated binary masks indicating the presence or absence of boulders within each image (Figure \ref{fig:boulder_seg}). Boulders were annotated by planetary scientists using precise polygon outlines, ensuring high-quality labels. The dataset originally provides train, validation, and test splits. This is one of the smallest datasets in \dataset{} with only tens of samples, and that makes it challenging for the computer vision community.

% \newpage

\begin{figure}[htbp]
  \centering

  \begin{subfigure}[b]{0.48\textwidth}
    \centering
    \includegraphics[width=0.40\linewidth]{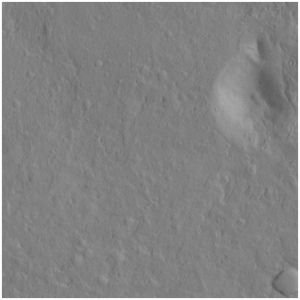}
    \includegraphics[width=0.40\linewidth]{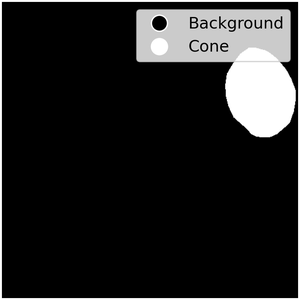}
  \end{subfigure}
  \hspace{-10mm}
  \begin{subfigure}[b]{0.48\textwidth}
    \centering
    \includegraphics[width=0.40\linewidth]{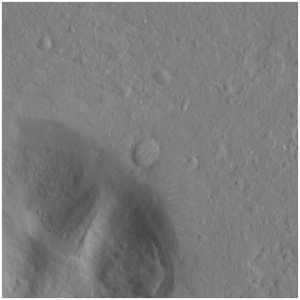}
    \includegraphics[width=0.40\linewidth]{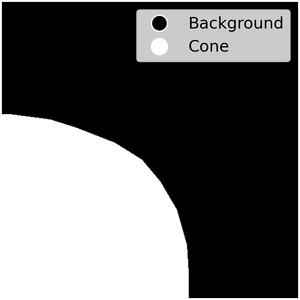}
  \end{subfigure}

  \caption{mb-conequest\_seg}
  \label{fig:conequest_seg}
\end{figure}

% \cite{purohit2024conequest}
\paragraph{mb-conequest\_seg} This is a binary segmentation dataset focused on identifying volcanic cones on the Martian surface using CTX imagery. It was developed to support global mapping and morphologic analysis of small-scale volcanic landforms. The dataset spans six geographically diverse regions on Mars, capturing substantial variation in cone shape, size, and appearance, making it a challenging benchmark for model generalization. Each sample consists of an image and its corresponding binary mask (Figure \ref{fig:conequest_seg}), with all annotations created and validated by expert geologists to ensure scientific accuracy.

Notably, the dataset includes negative samples (images without any cones), which introduces additional complexity by requiring models to correctly predict true negatives rather than detecting cones in every image. We provide metadata indicating which samples are negative, allowing users the flexibility to include or exclude them during training. This information can also be verified using simple image processing techniques. The dataset comes with pre-defined train, validation, and test splits.

\begin{figure}[htbp]
  \centering

  \begin{subfigure}[b]{0.48\textwidth}
    \centering
    \includegraphics[width=0.40\linewidth]{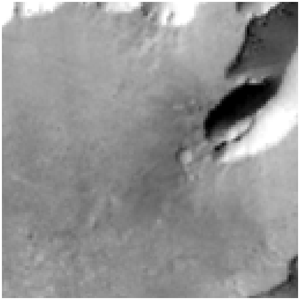}
    \includegraphics[width=0.40\linewidth]{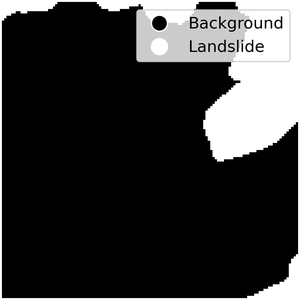}
  \end{subfigure}
  \hspace{-10mm}
  \begin{subfigure}[b]{0.48\textwidth}
    \centering
    \includegraphics[width=0.40\linewidth]{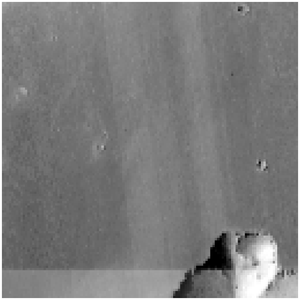}
    \includegraphics[width=0.40\linewidth]{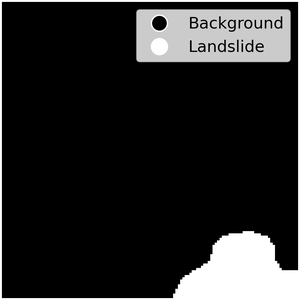}
  \end{subfigure}

  \caption{mb-mmls}
  \label{fig:mb-mmls}
\end{figure}

% \cite{paheding2024marsls}
\paragraph{mb-mmls} This is a binary segmentation dataset designed to identify landslides on the Martian surface, with a focus on the Valles Marineris region from the CTX sensor. All annotations were manually created by expert geologists, ensuring high-quality, scientifically accurate labels. Each image sample includes multi-modal satellite data comprising 7 channels: RGB (3), Digital Elevation Model (DEM), thermal inertia, slope, and grayscale intensity (in Figure \ref{fig:mb-mmls}, we have visualized grayscale channels). This rich set of modalities captures the complex geomorphology of landslide-prone regions, making the dataset especially valuable for developing and benchmarking robust segmentation models in planetary science. All experiments in this paper utilize only the RGB channels for training and evaluation. The dataset includes predefined train, validation, and test splits to support standardized evaluation.

% This is a binary segmentation dataset, focused on identifying volcanic cones on the Martian surface using CTX imagery. The dataset was created to support global mapping and morphologic analysis of small-scale volcanic landforms. It covers six geographically diverse regions on Mars, capturing high variability in cone shape, size, and visual appearance, posing a significant challenge for model generalization. The dataset consists of image-mask pairs. The dataset is also challenging as it contains negative samples as well, meaning that samples which does not contain even a single cone, which challenges the model to predict true negatives as well, instead of predicting cones in all images. We provide information about which data samples are negative patches, so, user to decide to exclude them during training, and also it can be checked via simple image processing tools. The dataset already provides train, validation, and test splits.

\begin{figure}[htbp]
  \centering

  \begin{subfigure}[b]{\textwidth}
    \centering
    \begin{subfigure}[b]{0.48\textwidth}
      \centering
      \includegraphics[width=0.40\linewidth]{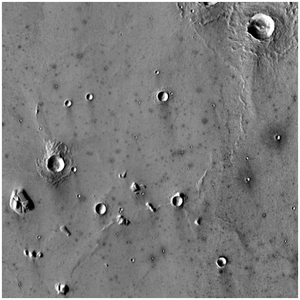}
      \includegraphics[width=0.40\linewidth]{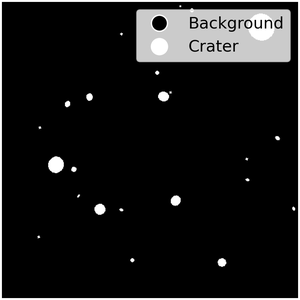}
    \end{subfigure}
    \hspace{-10mm}
    \begin{subfigure}[b]{0.48\textwidth}
      \centering
      \includegraphics[width=0.40\linewidth]{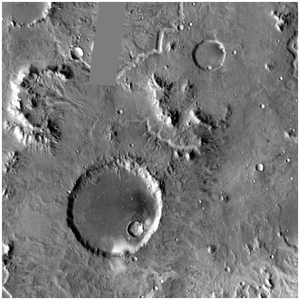}
      \includegraphics[width=0.40\linewidth]{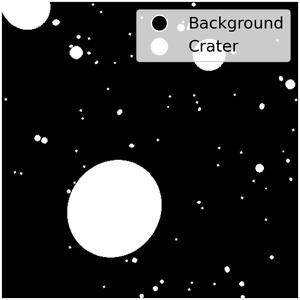}
    \end{subfigure}
    \caption{mb-crater\_binary\_seg}
    \label{fig:crater_binary_seg}
  \end{subfigure}

  \vspace{5mm}

  \begin{subfigure}[b]{\textwidth}
    \centering
    \begin{subfigure}[b]{0.48\textwidth}
      \centering
      \includegraphics[width=0.40\linewidth]{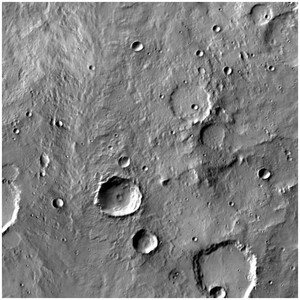}
      \includegraphics[width=0.40\linewidth]{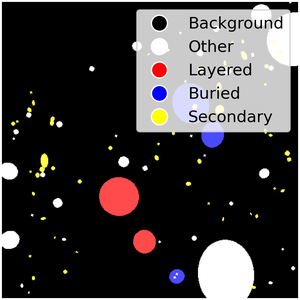}
    \end{subfigure}
    \hspace{-10mm}
    \begin{subfigure}[b]{0.48\textwidth}
      \centering
      \includegraphics[width=0.40\linewidth]{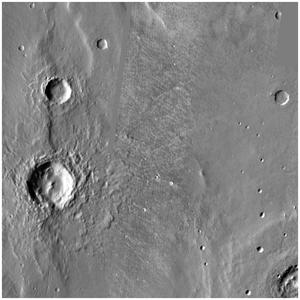}
      \includegraphics[width=0.40\linewidth]{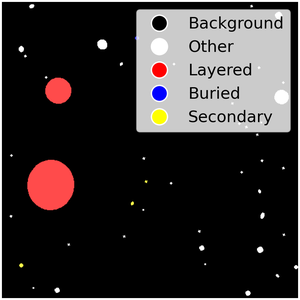}
    \end{subfigure}
    \caption{mb-crater\_multi\_seg}
    \label{fig:crater_multi_seg}
  \end{subfigure}

  \caption{mb-crater\_seg datasets}
  \label{fig:crater_seg}
\end{figure}

\paragraph{mb-crater\_binary\_seg \& mb-crater\_multi\_seg} These two datasets focus on crater segmentation using THEMIS imagery. \texttt{mb-crater\_binary\_seg} is a binary segmentation dataset that distinguishes crater vs. non-crater regions, while \texttt{mb-crater\_multi\_seg} is a multi-class segmentation dataset with four crater types: Other, Layered, Buried, and Secondary (Figure \ref{fig:crater_seg}). Craters are crucial for understanding the geological history, surface age, and impact processes of planetary bodies \cite{robbins2012new}. Moreover, classifying craters into distinct morphological types enables researchers to assess which crater types are more informative for scientific analysis \cite{lagain2021mars}.

Although craters are often described as bowl-shaped, they are not always perfectly circular. To address this, we provide annotations in elliptical form, marking the \textit{first} known release of crater segmentation using elliptical geometry, in contrast to the circular annotations commonly used in prior datasets \cite{malvi2023automated, delatte2019segmentation, benedix2020deriving}. As the original release consisted only of metadata, we generated the full image-mask dataset using open-source THEMIS data\footnote{\url{https://www.mars.asu.edu/data/thm_dir_100m/}}. Due to significant missing pixels near the poles, we restricted the dataset to within approximately $\pm30^\circ$ latitude of the Martian equator.

To prevent spatial data leakage, we created geographically disjoint train, validation, and test splits. Specifically, images from longitudes $-180^\circ$ to $60^\circ$ are assigned to the training set, $60^\circ$ to $120^\circ$ to the test set, and $120^\circ$ to $180^\circ$ to the validation set.

\newpage

\begin{figure}[htbp]
  \centering

  \begin{subfigure}[b]{\textwidth}
    \centering
    \begin{subfigure}[b]{0.48\textwidth}
      \centering
      \includegraphics[width=0.40\linewidth]{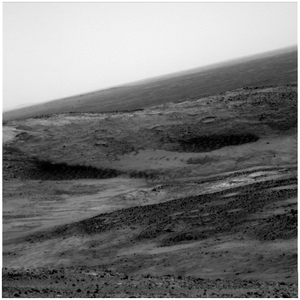}
      \includegraphics[width=0.40\linewidth]{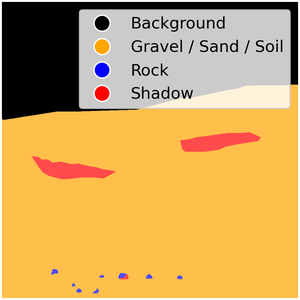}
    \end{subfigure}
    \hspace{-10mm}
    \begin{subfigure}[b]{0.48\textwidth}
      \centering
      \includegraphics[width=0.40\linewidth]{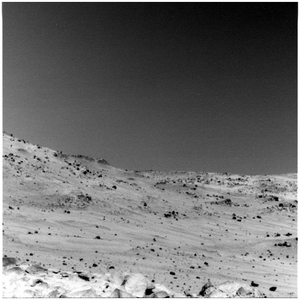}
      \includegraphics[width=0.40\linewidth]{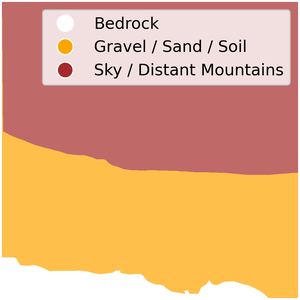}
    \end{subfigure}
    \caption{mb-mars\_seg\_mer}
    \label{fig:mb-mars_seg_mer}
  \end{subfigure}

  \vspace{5mm}

  \begin{subfigure}[b]{\textwidth}
    \centering
    \begin{subfigure}[b]{0.48\textwidth}
      \centering
      \includegraphics[width=0.40\linewidth]{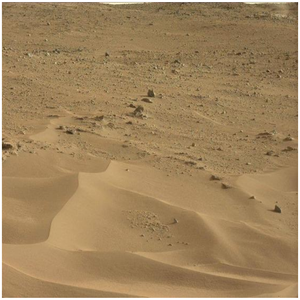}
      \includegraphics[width=0.40\linewidth]{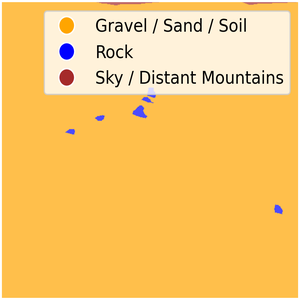}
    \end{subfigure}
    \hspace{-10mm}
    \begin{subfigure}[b]{0.48\textwidth}
      \centering
      \includegraphics[width=0.40\linewidth]{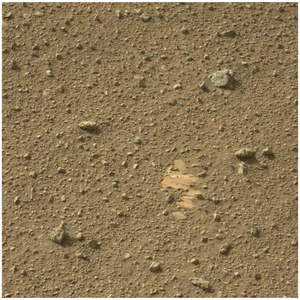}
      \includegraphics[width=0.40\linewidth]{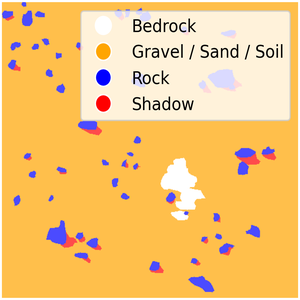}
    \end{subfigure}
    \caption{mb-mars\_seg\_msl}
    \label{fig:mb-mars_seg_msl}
  \end{subfigure}

  \caption{mb-mars\_seg datasets}
  \label{fig:mb-mars_seg}
\end{figure}

% \cite{li2022stepwise}
\paragraph{mb-mars\_seg\_mer \& mb-mars\_seg\_msl} These are multi-class segmentation datasets designed to support terrain understanding on Mars using imagery from two distinct rover missions. The \texttt{MSL} dataset corresponds to the Mars Science Laboratory (Curiosity) mission and includes imagery captured by Mastcam, while the \texttt{MER} dataset is sourced from the Mars Exploration Rover missions (Opportunity and Spirit), using Navcam and Pancam sensors. Both datasets are annotated with six terrain-related classes: Bedrock, Gravel/Sand/Soil, Rock, Shadow, Sky/Distant Mountains, and Track, representing typical surface elements observed during rover operations (Figure \ref{fig:mb-mars_seg}).

While the original datasets were annotated by planetary science experts, we applied additional refinement in \dataset{} by consolidating visually similar or ambiguous categories, such as different granular terrain types, based on expert consultation, aiming to reduce annotation inconsistencies and improve evaluation reliability (see Section \ref{subsec:correction}). Since the original datasets do not provide standard splits, we generated consistent train, validation, and test sets for both MER and MSL versions.

\begin{figure}[htbp]
  \centering

  \begin{subfigure}[b]{0.48\textwidth}
    \centering
    \includegraphics[width=0.40\linewidth]{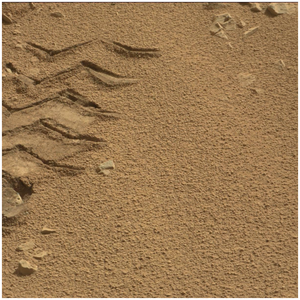}
    \includegraphics[width=0.40\linewidth]{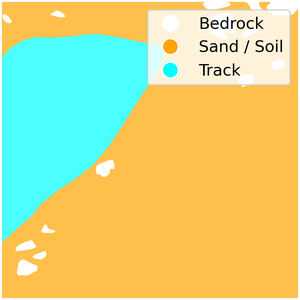}
  \end{subfigure}
  \hspace{-10mm}
  \begin{subfigure}[b]{0.48\textwidth}
    \centering
    \includegraphics[width=0.40\linewidth]{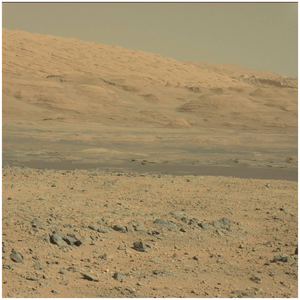}
    \includegraphics[width=0.40\linewidth]{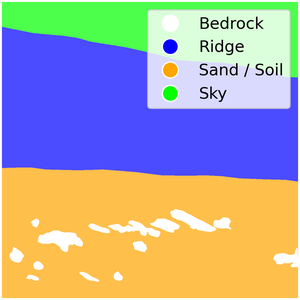}
  \end{subfigure}

  \caption{mb-s5mars}
  \label{fig:mb-s5mars}
\end{figure}

% \cite{zhang2022s}
\paragraph{mb-s5mars} This is a multi-class segmentation dataset developed to enable semantic understanding of Martian surface terrain using imagery captured by the Mastcam camera aboard the Curiosity rover. It contains 8 classes: Bedrock, Hole, Ridge, Rock, Rover, Sand/Soil, Sky, and Track, representing features commonly encountered during rover-based navigation and scientific exploration (Figure \ref{fig:mb-s5mars}). Although the original dataset is reported to be annotated by domain experts, in \dataset{} we further refine it by merging visually ambiguous classes based on expert analysis to reduce label noise and enhance the robustness of model evaluation (see Section \ref{subsec:correction} for details). The dataset includes predefined train, validation, and test splits.

\subsubsection{Object Detection}
\label{subsubsec:object_detection}

As described in Section 4, all object detection datasets in \dataset{} are provided in multiple annotation formats to support a broad range of models and frameworks. Specifically, we include annotations in \texttt{COCO}, \texttt{Pascal VOC}, and \texttt{YOLO} formats. This ensures compatibility with most object detection pipelines and reduces the effort and time required for format conversion by end users.

\begin{figure}[htbp]

  \centering
  \begin{subfigure}{\columnwidth}
    \centering
    \includegraphics[width=0.25\columnwidth]{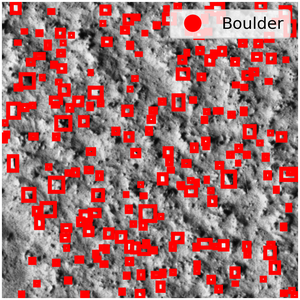}
    \includegraphics[width=0.25\columnwidth]{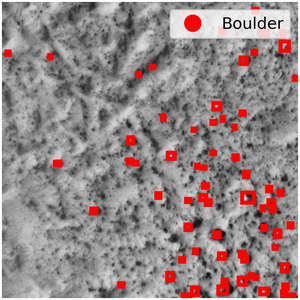}
  \end{subfigure}

  \caption{mb-boulder\_det}
  \label{fig:boulder_det}

\end{figure}

% \cite{prieur2023automatic}
\paragraph{mb-boulder\_det} This is the object detection version of the Boulder dataset, designed to localize boulders on the Martian surface using high-resolution orbital imagery from the HiRISE camera. Each image is annotated with manually curated bounding boxes that delineate individual boulders (Figure \ref{fig:boulder_det}), with annotations created by planetary scientists to ensure high-quality and scientifically accurate labels. The dataset includes predefined train, validation, and test splits. This is one of the smallest datasets in \dataset{} with only tens of samples. Given the small-object nature of the task, this dataset presents a valuable benchmark for evaluating object detection models in low-data regimes, a setting of growing interest in the computer vision community.

% \newpage

\begin{figure}[htbp]

  \centering
  \begin{subfigure}{\columnwidth}
    \centering
    \includegraphics[width=0.25\columnwidth]{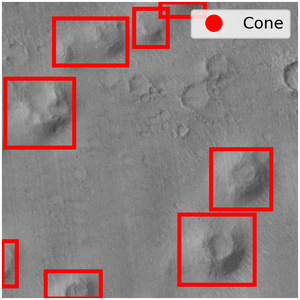}
    \includegraphics[width=0.25\columnwidth]{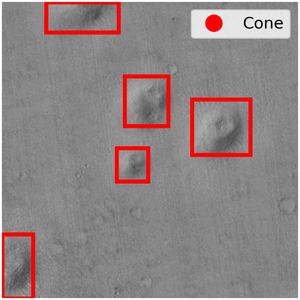}
  \end{subfigure}

  \caption{mb-conequest\_det}
  \label{fig:conequest_det}

\end{figure}

% \cite{purohit2024conequest}
\paragraph{mb-conequest\_det} This is the object detection version of the ConeQuest dataset, formulated to localize cones on the Martian surface using CTX imagery. It was developed to support global mapping and morphologic analysis of small-scale volcanic landforms. The dataset spans six geographically diverse regions on Mars, capturing substantial variation in cone shape, size, and appearance, making it a challenging benchmark for model generalization. Each sample consists of an image and its bounding boxes (Figure \ref{fig:conequest_det}), with all annotations created by expert geologists to ensure scientific accuracy.

As the original ConeQuest contains negative samples, we have removed it from this version of the dataset as many detection models do not support training with image samples that do not have any objects. The dataset comes with pre-defined train, validation, and test splits.

\newpage

\begin{figure}[htbp]

  \centering
  \begin{subfigure}{\columnwidth}
    \centering
    \includegraphics[width=0.25\columnwidth]{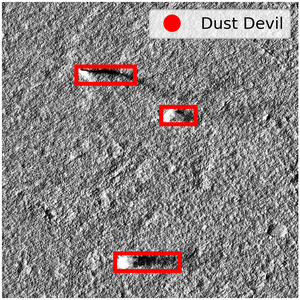}
    \includegraphics[width=0.25\columnwidth]{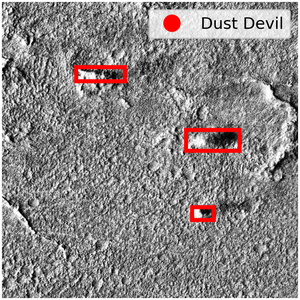}
  \end{subfigure}

  \caption{mb-dust\_devil\_det}
  \label{fig:dust_devil_det}

\end{figure}

% \cite{guo2024martian}
\paragraph{mb-dust\_devil\_det} This task focuses on identifying dust devils in Martian orbital imagery. Dust devils are small-scale, short-lived whirlwinds that play a significant role in Martian atmospheric dynamics, surface modification, and dust transport (Figure \ref{fig:dust_devil_det}). The dataset consists of CTX images with manually annotated bounding boxes around visible dust devils. Detecting dust devils presents a considerable challenge due to their faint visibility, small size, and the similarity in texture to surrounding terrain. It includes predefined training, validation, and test splits.

\subsection{Few-Shot and Partitioned Data Preparation}
\label{subsec:few_shot_partition}

To facilitate benchmarking and enable analysis of model performance across varying data regimes, we release both partitioned and few-shot versions of the datasets. These versions are particularly useful for studying how different methods perform with respect to training dataset size and how quickly they reach performance saturation. Importantly, only the training sets are modified in these variants; the validation and test sets remain unchanged across all versions to ensure fair comparisons.

\paragraph{Partitioned Datasets} We provide partitioned training sets for all datasets across all task types in \dataset. Specifically, we generate pre-defined subsets using the following proportions of the original training data: 1\%, 2\%, 5\%, 10\%, 20\%, 25\%, and 50\%. This results in a total of 131 partitioned datasets. These subsets enable systematic evaluation of how models scale with increasing amounts of data.

\paragraph{Few-Shot Datasets} Few-shot versions are provided for all \textit{classification} tasks (except mb-change\_cls\_ctx) in \dataset. We include the following few-shot configurations: 1-shot, 2-shot, 5-shot, 10-shot, 15-shot, and 20-shot, totaling 54 additional datasets.

For the multi-label classification dataset (mb-surface\_multi\_label\_cls), special care is taken to ensure that each class appears at least the specified number of times in the few-shot setting. For example, in the 2-shot version, every class is represented by at least two instances across the dataset.

\paragraph{Special Cases and Exceptions}
\begin{itemize}
    \item For mb-change\_cls\_ctx, we do not provide few-shot versions, as the original dataset is already very small.
    \item For mb-boulder\_seg, mb-boulder\_det, and mb-change\_cls\_ctx; partitioned datasets are only provided starting from 10\%, as smaller subsets resulted in empty or unusable training splits.
\end{itemize}

These curated few-shot and partitioned datasets are designed to support robust, reproducible research in data-scarce scenarios and facilitate the development and comparison of new methods.

% in planetary science.

% We provide and partition and few shot version of the datasets. The purpose of providing a partition and a few shots is to study how a particular method performs with respect to training dataset size and how early it can saturate. With this, we also believe that for a methodology that will be proposed in the future, providing this partition makes it easier to compare with each other and find state-of-the-art results on each dataset. Here, it is important to note that for these released versions, only the training set is getting affected, validation and testing sets are still the same.

% We provide partition datasets for all the datasets of all task types in \dataset. 
% Specifically, we generate pre-defined subsets using the following ratios of the training set: 1\%, 2\%, 5\%, 10\%, 20\%, 25\%, and 50\%. These resulted in 140 datasets.

% With this, we also provide a few-shot dataset version in all classification datasets. Particularly, we provide 1-shot, 2-shot, 5-shot, 10-shot, 15-shot, and 20-shot. That results in an additional 54 datasets. For mb-surface\_multi\_label\_cls, as this is the multi-label dataset, the way we prepared the dataset is that if it is 2-shot, we make sure that all the classes are covered \textit{at least} 2 times in the whole dataset.

% It is important to note that we did not provide a few shots of data for mb-change\_cls\_ctx because the data itself is very small. For partition in mb-change\_cls\_ctx and mb-boulder\_seg, we only provide data starting from 5\% because the lower partition end up having no training data.

\subsection{Expert-Driven Corrections and Refinements}
\label{subsec:correction}

To ensure the quality, usability, and consistency of the released datasets, we made several corrections and modifications. These changes are aimed at improving clarity, removing redundancies, and aligning the datasets with modern machine learning practices:

\begin{itemize}
    \item Removal of Augmented Samples: For datasets such as mb-landmark\_cls and mb-surface\_multi\_label\_cls, we removed pre-augmented versions of samples that were originally included. We believe it is more appropriate to provide only the raw samples, allowing users to apply their own augmentation strategies using state-of-the-art computer vision techniques. Pre-included augmentations can often be redundant or incompatible with newer workflows.
    \item Harmonization of Surface and Landmark Classification Datasets: Wagstaff et al. released two versions of the surface and landmark classification datasets \cite{wagstaff2018deep, wagstaff2021mars}. After consulting the authors, we merged the surface classification datasets (released as mb-surface\_cls) by:
    \begin{itemize}
        \item Combining identical classes (e.g., wheel present in both versions).
        \item Merging semantically similar classes (e.g., nearby surface and ground).
        \item Removing duplicate samples (keeping only one if identical samples existed).
        \item Eliminating the ambiguous class Other rover part, which originally served as a catch-all category. The merged dataset now includes a broader and more clearly defined set of surface-related classes.
    \end{itemize}
    For the landmark classification datasets, we retained only the newer version, as recommended by the authors.
    \item Balancing in ConeQuest Dataset: The original ConeQuest dataset \cite{purohit2024conequest} included a significant class imbalance, with only $\sim 12\%$ of samples containing cones (positive samples), and the rest being negative. While the intent was to include broad geographic coverage and true negative learning, we found this imbalance suboptimal for general use.
    \begin{itemize}
        \item In mb-conequest\_seg, we balanced the positive and negative samples across each region while preserving the geographic diversity. We followed the exact methodology described in the original experimental setup.
        \item In mb-conequest\_det, we excluded all negative samples, as many object detection models do not support samples without any annotated objects.
    \end{itemize}
    \item Correction of Ambiguous Annotations in Terrain Segmentation Datasets: We performed expert reviews of the terrain segmentation datasets \cite{li2022stepwise, zhang2022s} after observing inconsistencies. Experts identified several annotation ambiguities, especially between visually similar classes such as soil, sand, and gravel. These challenges are common in pixel-level annotation tasks due to the fine granularity and visual overlap.
    \begin{itemize}
        \item In mb-s5mars, we merged the classes sand, soil, and gravel into a single category.
        \item In mb-mars\_seg\_mer and mb-mars\_seg\_msl, we merged sand and soil.
    \end{itemize}
    These refinements aim to provide cleaner, more consistent datasets that are easier to use, compare, and extend for downstream machine learning tasks in planetary exploration.
\end{itemize}

\subsection{Excluded Datasets}
\label{subsec:removed_datasets}

While our paper includes a carefully curated selection of datasets, there are several others in the literature that we chose not to include for various reasons, detailed below:

\begin{itemize}
    \item AI4MARS \cite{swan2021ai4mars}: Upon expert review, we identified annotation errors in this dataset. Since the annotations were produced via crowdsourcing, we found them unsuitable for a highly specialized domain like planetary science, where expert validation is crucial. Consequently, we excluded this dataset.
    \item Cone Detection (Mills et. al.) \cite{mills2024global}: The authors did not release the actual training data. Instead, they shared outputs generated by their own pipeline, global mappings of cones as bounding boxes with latitude and longitude coordinates. This data is not validated by experts and cannot directly support downstream machine learning tasks.
    \item Cone Detection (Chen et. al.) \cite{chen2024global}: The dataset was not released in a format suitable for machine learning applications. Additionally, no instructions were provided to preprocess or structure the data for ML pipelines, making its use impractical.
    \item Cone Detection and Segmentation \cite{yang2024mapping}: Although the authors indicated the dataset would be available upon request, we reached out and never received a response, leaving us unable to include the data.
    \item Novelty Detection \cite{kerner2019novelty} and Outlier Detection \cite{kerner2022domain}: These datasets do not fall into the task categories we currently support, i.e., classification, segmentation, or object detection. Furthermore, significant preprocessing would be required. We may consider including them in a future extended version of \dataset.
    \item Rockfall Detection \cite{bickel2021labeled}: Our analysis of the training and test sets revealed a significant number of false negatives (FNs). Although the paper acknowledges this possibility, such inconsistencies hinder reliable model evaluation, especially when FNs are present in the test set, so we excluded this dataset.
    \item SPOC \cite{rothrock2016spoc}: The dataset link was not provided in the paper. Upon contacting the authors, we learned that SPOC is an earlier and slightly different version of the AI4MARS dataset \cite{swan2021ai4mars}, and that it is significantly smaller. The authors recommended using AI4MARS instead.
\end{itemize}

\section{Experiments Details}
\label{sec:experimental_setup}

\subsection{Details of Earth Observation Baselines}
\label{subsec:eo_experiments_details}

For Earth Observation (EO) baselines, we follow the same experimental protocol used for models pre-trained on natural images. Specifically, we perform hyperparameter tuning for each EO model and then train and evaluate the models across all dataset partitions as well as the full training set, using seven random seeds to ensure robust evaluation.

Among the four datasets evaluated with EO-based models, \texttt{mb-surface\_cls} is an RGB dataset, while the others are grayscale. EO foundation models such as SatMAE, CROMA, and Prithvi are pre-trained on multi-spectral inputs and thus expect input images with multiple channels, often significantly more than standard RGB data. For example, CROMA requires 12 channels for its optical encoder and 2 channels for its radar encoder. As there are multiple versions of these models available, we selected ViT-L from all of them. Since CROMA offers two encoder options, we selected the optical encoder because all the datasets used in our evaluation are from optical sensors. Similarly, for SatMAE, which provides multiple pre-trained encoder choices, we chose the encoder that was pre-trained on the non-temporal subset of the fMoW dataset, as it aligns best with our data characteristics instead of the multi-spectral encoder.

To adapt single-channel or RGB Mars datasets for these multi-spectral models, we replicate the available channels as needed. For grayscale images, the single channel is duplicated to match the required number of input channels. This is a standard practice suggested in prior works and consistent with recommendations in the timm library documentation and repositories such as TorchSeg \cite{torchseggithub}. While this approach does not introduce new spectral information, it allows for compatibility with the pre-trained architecture without retraining encoders from scratch.

% For EO baselines, we follow the same experiment style as we it for other models which are pre-trained on natural images data. We first do the hyperparameter tuning, and then train and evaluate the model on all partitions data along with full dataset for 7 different random seed. From all the 4 datasets which we have evaluated on EO-based models, mb-landmark\_cls is RGB dataset, and all other grayscale. As EO-based models are pre-trained on multi-spectral data, it expects data to have multi-channel depending on model architecture. For example, CROMA needs 12-channels in optical encoder, and 2-channels in radar encoder. Hence, the way we feed data to the model is just by duplicating channels as suggested in \cite{torchseggithub} and in the documentation of timm library.

\newpage

\subsection{Pipeline and Hyperparameters}
\label{subsec:pipeline_and_hyperparameters}

\begin{wraptable}{r}{0.62\linewidth}
    % \begin{table*}[ht]
      \centering
      % \small
      \begin{tabular}{@{} l l @{}}
        \toprule
        \textbf{config}         & \textbf{value}                 \\
        \midrule
        seed                    & 0, 1, 10, 42, 123, 1000, 1234  \\
        learning rate schedule  & w/o, cosine, plateau, step     \\
        base learning rate      & 1e-3, 1e-4, 1e-5               \\
        weight decay            & 0.05                           \\
        batch size              & 16, 32, 64                     \\
        optimizer               & Adam, AdamW, SGD               \\
        max training epochs     & 50, 100, 200                   \\
        patience                & 5, 10                          \\
        \bottomrule
      \end{tabular}
      \caption{Training hyperparameters}
      \label{tab:hyperparams}
    % \end{table*}
\end{wraptable}

We provide a user-friendly and scalable training and inference pipeline for classification, segmentation, and object detection tasks. The pipeline supports running experiments via command-line arguments, allowing easy configuration of core parameters such as dataset, model architecture, training type, and hyperparameters.

It includes modular support for logging with options for Weights \& Biases (Wandb), TensorBoard, and CSV; model checkpointing; early stopping; and other PyTorch Lightning-compatible callbacks. For reproducibility, we fix the random seed across all relevant libraries and save Hydra configuration files and logs locally. Since \dataset{} is released on both Hugging Face and Zenodo, the pipeline supports loading data from either platform.

\begin{wraptable}{r}{0.58\linewidth}
    % \begin{table*}[ht]
      \centering
      \small
      \begin{tabular}{@{} l l @{}}

        \multicolumn{2}{l}{\textbf{Classification}} \\
        \toprule
        \textbf{config}         & \textbf{value} \\
        \midrule
        criterion               & \makecell{cross entropy, binary cross entropy\\(only for binary classification)} \\
        \midrule

        \\

        \multicolumn{2}{l}{\textbf{Segmentation}} \\
        \toprule
        \textbf{config}         & \textbf{value} \\
        \midrule \vspace{2mm}
        criterion               & \makecell{generalized\_dice (square, simple,\\linear), cross entropy, combined} \\
        smoothing value         & 1e-5 (only for generalized\_dice) \\
        \bottomrule

      \end{tabular}
      \caption{Configuration for loss function}
      \label{tab:criterion}
    % \end{table*}
\end{wraptable}

As described in Section 4, for each combination of model, dataset, and training strategy, we first perform hyperparameter tuning. We tune the learning rate, learning rate scheduler, weight decay, batch size, optimizer, and maximum number of training epochs. The full search space is listed in Table \ref{tab:hyperparams}.

We also experiment with different loss functions, summarized in Table~\ref{tab:criterion}. For binary classification, we try both a one-node output with binary cross-entropy and a two-node output with standard cross-entropy. For segmentation, we evaluate three loss types: generalized Dice loss, cross-entropy, and a weighted combination of both. We also explore three different weighting schemes. For object detection, we use the default loss returned by each model implementation.

% We recognize our user would have different requirements based on their background, therefore we have kept our pipe as abstract and easy to use to allow no code training, benchmarking and inference via complementing command line builder and customizable such that each element of our pipe can be modified via command line or extending current offerings via template design pattern. We have utilized various creational, structural, and behavioral class design patterns to keep a smooth user experience while understanding and maintaining code. We appreciate public contributions to add Datasets, Models, or pipeline elements. We also share HFDataset, which allows users to use their choice of Hugging Face dataset as long as they follow our nomenclature.

% All the experiments that we have done in this paper, 

% All of our tasks support cross-entropy loss (BCE for binary/multi-label), generalized dice loss (segmentation), and multiple optimizers (Adam, Adamw, and SGD), extendible to more in the future. To support a smoother training experience, we have boosted logging quantitatively(all global and class-wise core metrics reported in the paper and more) and qualitatively (task-specific multiple format reporting for set samples each n epochs).

\subsection{Number of Experiments}
\label{subsec:number_of_experiments}

Due to the large number of models, datasets, training strategies, and data splits involved in our benchmark, we summarize here the scale of experiments conducted.

As described in Section 4, we begin by performing hyperparameter tuning for every unique combination of model, dataset, and training type. This includes 13 models, 20 datasets, and 3 training strategies, resulting in 780 hyperparameter tuning runs. Each of these runs was repeated multiple times, depending on the number of configurations in the search space. Once the best hyperparameters were selected, we retrained each configuration using 7 different random seeds to ensure robust and stable performance reporting.

\paragraph{Classification} We evaluated 5 models across 9 datasets, under 3 training types and 7 random seeds, totaling \textbf{945} classification experiments. In addition, we evaluated all of these combinations on 7 partitioned versions of each dataset, resulting in 6,615 runs. However, for the \texttt{mb-change\_cls\_ctx} dataset, we excluded 1\%, 2\%, and 5\% partitions due to insufficient training samples, which led to the exclusion of 315 experiments. The final count for partitioned classification experiments is \textbf{6,300}. We also included few-shot evaluation, using 6 different configurations (1-shot to 20-shot) across all classification datasets except \texttt{mb-change\_cls\_ctx}, which adds \textbf{5,040} few-shot experiments.

\paragraph{Segmentation} We evaluated 4 models on 8 datasets using 3 training strategies and 7 random seeds, resulting in \textbf{672} standard segmentation experiments. We also performed partitioned experiments using 7 training set splits. However, for \texttt{mb-boulder\_seg}, we excluded the 1\%, 2\%, and 5\% partitions due to extremely limited data, resulting in 252 experiments being skipped. This leads to a total of \textbf{4,452} partitioned segmentation runs.

\paragraph{Object detection} We ran experiments using 4 models across 3 datasets, again under 3 training strategies and 7 random seeds, totaling \textbf{252} standard experiments. We did not perform partition experiments in object detection due to lower performance on the full dataset (more details in Section \ref{subsec:object_detection_results}).

% We repeated these experiments across 7 partitioned versions of the training sets, but excluded the lowest three partitions for \texttt{mb-boulder\_det}, removing 252 experiments. This resulted in \textbf{1,512} partitioned object detection runs.

% \bigskip

% In summary, we conducted:
% \begin{itemize}
%     \item 780 hyperparameter tuning runs
%     \item 945 standard classification experiments
%     \item 6,300 classification partition experiments
%     \item 5,094 few-shot classification experiments
%     \item 672 standard segmentation experiments
%     \item 4,452 segmentation partition experiments
%     \item 252 standard detection experiments
%     \item 1,512 detection partition experiments
% \end{itemize}

In addition, for EO baselines, we conducted experiments on 4 datasets using 3 EO-pretrained models, evaluated over 7 random seeds and 8 training sizes (7 partitions plus the full dataset), resulting in a total of \textbf{672} additional runs.

In summary, we conducted over \textbf{19,000 total model runs}, making \dataset{} one of the most comprehensively evaluated benchmarks for Mars science and planetary vision research. All the experiments were conducted on NVIDIA A100-SXM4 and NVIDIA A30 based on availability on the ASU Sol supercomputer \cite{asusol}.

\subsection{Reporting Results}
\label{subsec:reporting_results_appendix}

As mentioned in Section 4.1 and inspired by the methodologies in \cite{agarwal2021deep} and \cite{lacoste2023geo}, we follow a consistent procedure to report results across thousands of experiments. We report outcomes separately for each training setting (random initialization, frozen pre-trained feature extractor, and pre-training with full fine-tuning), which allows for direct performance comparisons across different training settings.

First, we perform hyperparameter tuning for each model–dataset combination, selecting the best configuration based on validation loss using early stopping. Once the optimal hyperparameters are determined, we train and evaluate each model–dataset combination for 7 times with different random seeds, as recommended in prior work \cite{agarwal2021deep, lacoste2023geo}. For each combination, we compute the InterQuartile Mean (IQM) by discarding the top and bottom 25\% of scores and averaging the remaining values. This approach helps reduce both bias and variance in the reported performance. Before aggregating results across tasks, we normalize the scores within each task to account for differences in scale.

To quantify uncertainty, we perform 1,000 rounds of stratified bootstrapping. In each round, we sample (with replacement) one trial from each dataset, recompute the IQM across all datasets, and build a distribution of IQM values. From this distribution, we calculate 95\% confidence intervals. In our final results, we present per-task baselines and overall model performance (aggregated across all tasks) via violin plots. This process is repeated independently for each training setting, and results are reported separately to maintain clarity and consistency.

The results shown in Figures 2, 3, 4, and 5  \textit{in the main paper} are normalized only, without any aggregation. While the main paper and the appendix report both normalized and aggregated results for the feature extraction setting, we also include the corresponding raw results: F1-score for classification, IoU for segmentation, and mAP for object detection. For all other training types, we report only the raw results without normalization or aggregation.

\newpage

\section{Extended Results}
\label{sec:extended_results}

In this section, we present key observations derived from the full set of experiments conducted in this study.

\subsection{Classification Results}
\label{subsec:classification_results}

Figures \ref{fig:results_classification_actual_fe}, \ref{fig:results_classification_actual_tl}, and \ref{fig:results_classification_actual_st} present the classification results (F1-score) for all datasets under feature extraction, transfer learning, and training from scratch settings, respectively. We exclude results for \texttt{mb-change\_cls\_ctx} as it shows negligible variation across different models and training strategies. Overall, feature extraction consistently achieves the highest F1-scores across all models and datasets, followed by transfer learning. Training from scratch performs the worst, with noticeably higher variance. The performance gap between transfer learning and feature extraction is generally smaller than that between transfer learning and scratch, particularly for larger models like SwinV2-B and ViT-L/16.

% Figure \ref{fig:results_classification_actual_fe}, \ref{fig:results_classification_actual_tl}, and \ref{fig:results_classification_actual_st} show results of all classification datasets based on F1-score for feature extraction, transfer learning, and trained from scratch training, respectively. Please note that we do not show results for mb-change\_cls\_ctx as it does not show a significant difference across different models or training settings, and we defer showing results for this dataset. These results show that feature Extraction consistently achieves the highest F1-scores across all models and datasets, followed by transfer learning, while training from Scratch performs the worst with the largest variance. The performance gap between transfer learning and feature extraction is smaller than between transfer learning and scratch, especially for larger models like SwinV2-B and ViT-L/16.

\begin{figure}[!h]
    \centering
    \includegraphics[width=0.99\textwidth]{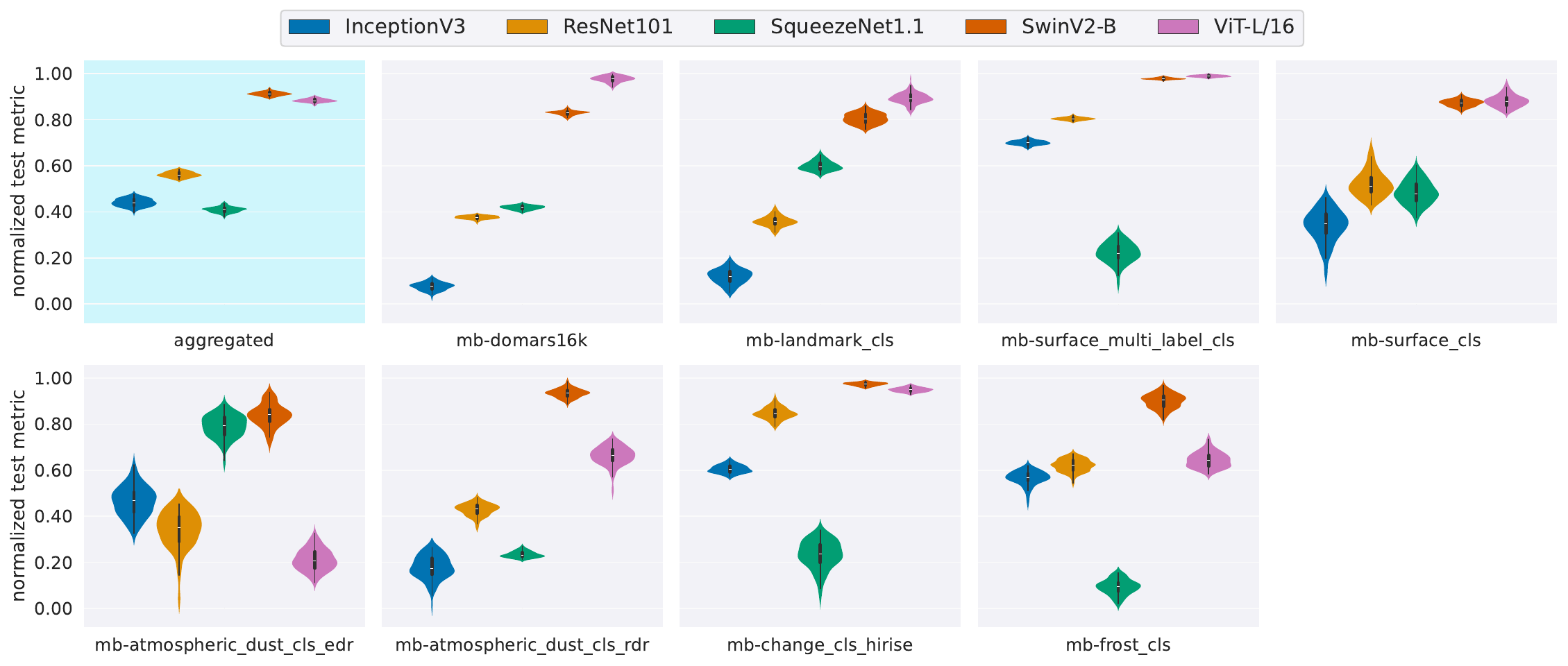}
    \caption{\textbf{Classification Benchmark under Feature Extraction setting:} Normalized F1-score of various baselines (higher is better). Violin plots are obtained from bootstrap samples of normalized IQM (Section \ref{subsec:reporting_results_appendix}). The left plot reports the average across all tasks.}
    \label{fig:results_classification_normalized_fe}
\end{figure}

\begin{figure}
    \centering
    \includegraphics[width=0.99\textwidth]{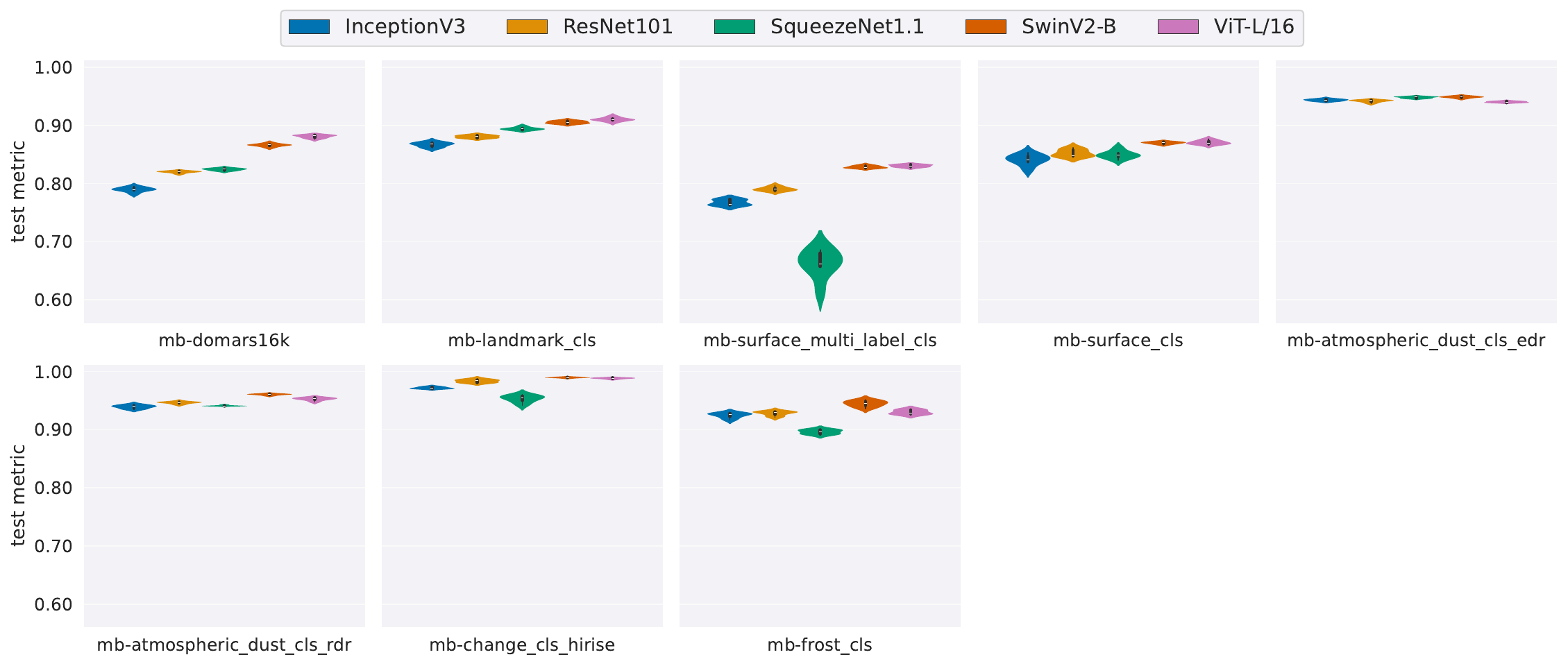}
    \caption{\textbf{Classification Benchmark under Feature Extraction setting:} Raw F1-score of various baselines (higher is better). Violin plots represent the distribution of seeds.}
    \label{fig:results_classification_actual_fe}
\end{figure}

\begin{figure}
    \centering
    \includegraphics[width=0.99\textwidth]{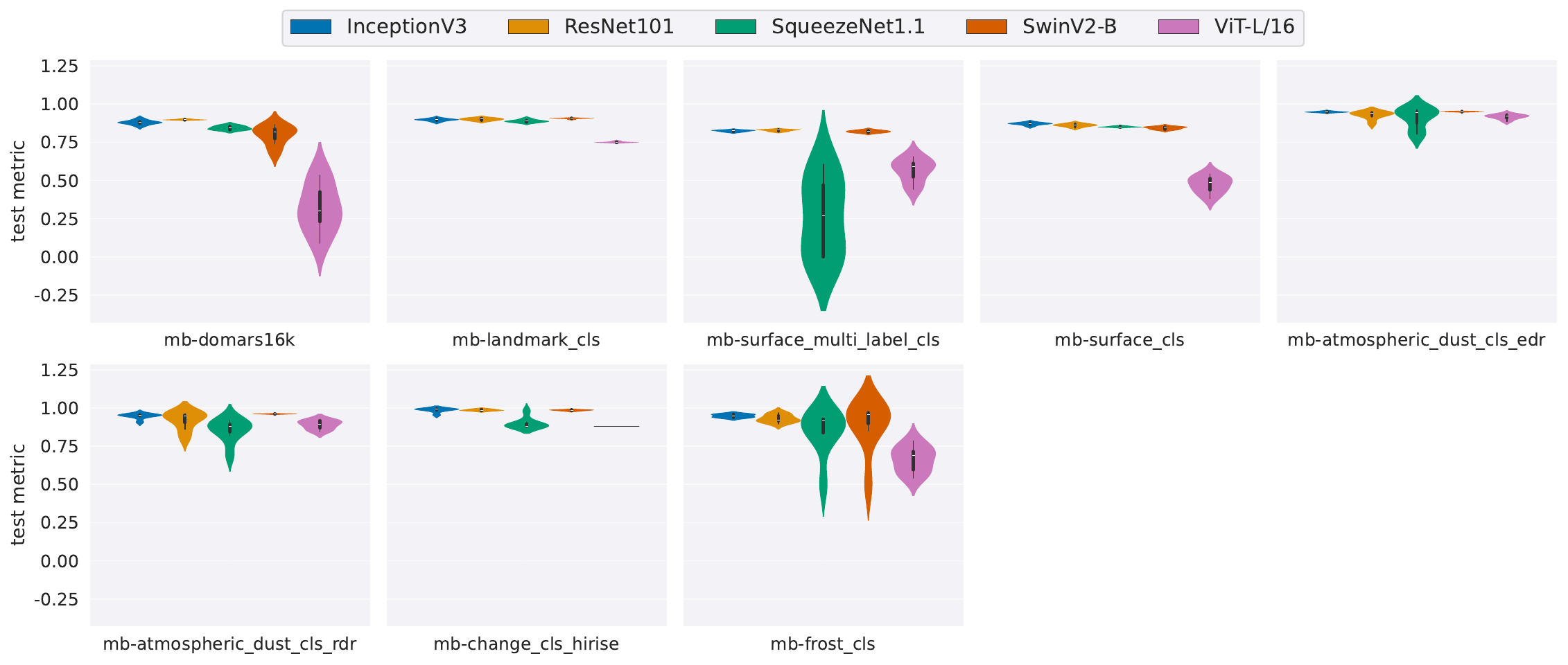}
    \caption{\textbf{Classification Benchmark under Transfer Learning setting:} Raw F1-score of various baselines (higher is better). Violin plots represent the distribution of seeds.}
    \label{fig:results_classification_actual_tl}
\end{figure}

\begin{figure}
    \centering
    \includegraphics[width=0.99\textwidth]{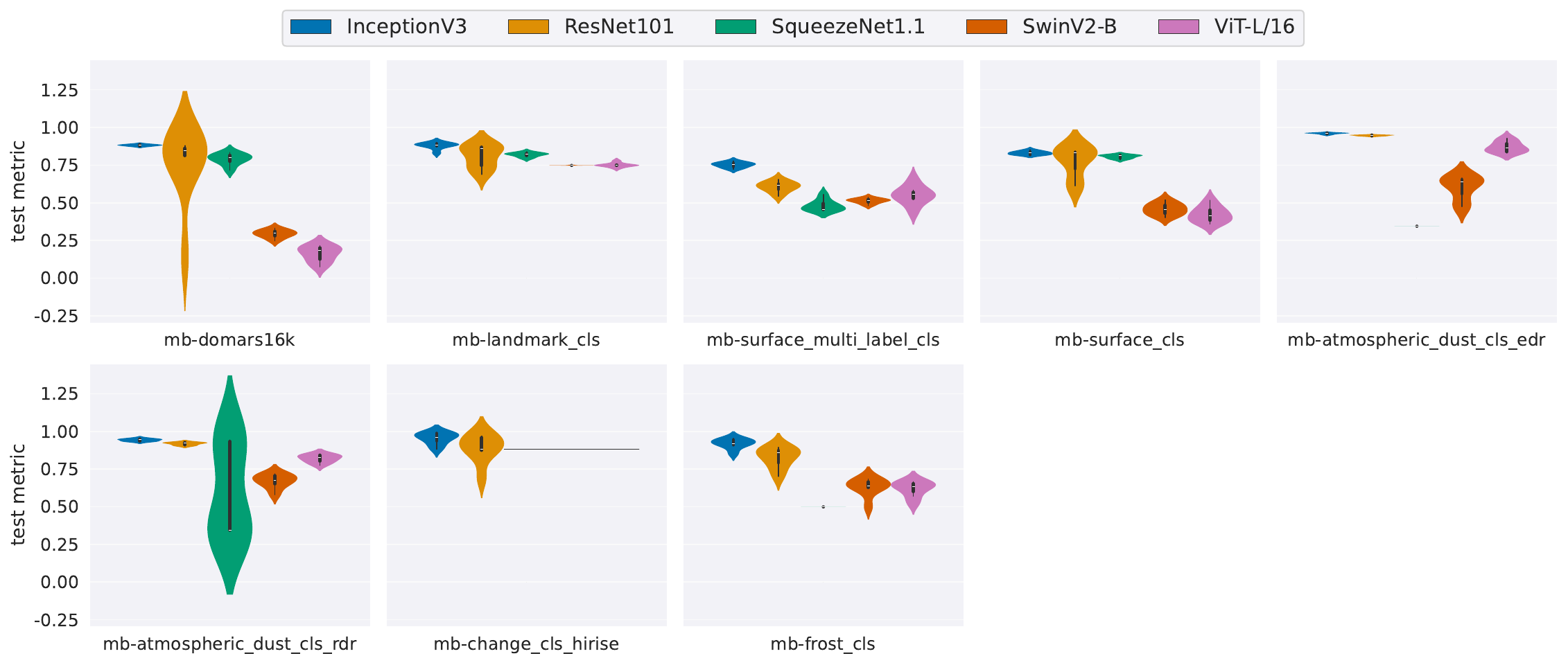}
    \caption{\textbf{Classification benchmark with models trained from scratch:} Raw F1-score of various baselines (higher is better). Violin plots represent the distribution of seeds.}
    \label{fig:results_classification_actual_st}
\end{figure}

\newpage

\subsubsection{Addressing Data Imbalance}
\label{subsubsec:data_imbalance}

Several datasets in \dataset{} have significant class imbalance, which can negatively impact model performance, especially for minority classes. In this section, we evaluate the effectiveness of three common strategies; under-sampling, over-sampling, and loss weighting to mitigate data imbalance and assess their impact on classification performance. All experiments are conducted under the feature extraction setting.

\begin{itemize}
    \item \textbf{Under-sampling:} For each dataset, we identify the class with the fewest samples (the minority class) and randomly sample an equal number of instances from all other classes to match this count. This results in a balanced dataset with uniform class distribution, though with a reduced overall training size.
    \item \textbf{Over-sampling:} Rather than directly upsampling all minority classes to match the largest class, which can lead to excessive duplication, we first modestly down-sample the majority class and then apply data augmentation to upsample the minority classes. This approach helps avoid extreme repetition (e.g., scaling a minority class 200×) while still achieving a more balanced class distribution.
    \item \textbf{Loss weighting:} Instead of modifying the dataset directly, we adjust the loss function to give higher weight to minority classes. Class weights are computed inversely proportional to class frequencies and incorporated into the loss calculation, encouraging the model to pay more attention to underrepresented classes during training.
\end{itemize}

We applied these techniques across four highly imbalanced datasets from \dataset: (1) mb-landmark\_cls, (2) mb-surface\_cls, (3) mb-change\_cls\_ctx, and (4) mb-change\_cls\_hirise. Experiments were conducted under feature extraction settings using all 5 classification models: SqueezeNet, ResNet, Inception, Vision Transformer (ViT), and Swin Transformer (SwinT). We compare these three balancing strategies against the baseline (standard training without any data manipulation). Results demonstrate how each technique affects overall model performance and performance on minority classes.

\begin{table*}[htbp]
    \centering
    
    \resizebox{0.99\linewidth}{!}{
      \begin{tabular}{lcccc}
        \toprule
         & \textbf{mb-landmark\_cls}
         & \textbf{mb-surface\_cls}
         & \textbf{mb-change\_cls\_ctx} 
         & \textbf{mb-change\_cls\_hirise} \\
        \midrule
        \textbf{Standard}       & 0.84 & 0.77 & 0.78 & 0.86 \\
        \textbf{Oversampling}   & 0.60 & 0.54 & 0.54 & 0.76 \\
        \textbf{Undersampling}  & 0.41 & 0.34 & 0.58 & 0.79 \\
        \textbf{Loss weighting} & 0.68 & 0.58 & 0.68 & 0.81 \\
        \bottomrule
      \end{tabular}
    }
  \caption{Comparison of class imbalance handling strategies against the standard baseline training, based on weighted F1-score.}
  \label{tab:imbalance_methods}
\end{table*}

\begin{itemize}
    \item \textbf{Overall performance:} From the table \ref{tab:imbalance_methods}, it can be observed that while data manipulation techniques help balance classes, they often result in a decrease in overall accuracy compared to the standard setup. Among the techniques, loss weighting generally maintains better performance, but even it does not consistently outperform the standard baseline across all scenarios.
\end{itemize}

However, the aggregated results do not provide a complete picture of how each technique impacts the performance of the minority and majority classes individually. To interpret these results better, we analyzed the class-wise performance by comparing each data manipulation technique with the baseline (standard setup):

\begin{itemize}
    \item \textbf{Loss weighting:} Same as aggregated results, loss weighting is the most consistent technique in improving minority class performance across all datasets and models, while also maintaining relatively stable performance for the majority class.
    \begin{itemize}
        \item Example: In the mb-change\_cls\_hirise dataset using ResNet, the minority class (Change) accuracy improved from 0.17 to 0.48, while the majority class (No change) performance dropped only slightly from 0.96 to 0.94. Similarly, in the mb-surface\_cls dataset, the minority class (Arm Cover) improved from 0.00 to 0.50 with ResNet.
    \end{itemize}
    \item \textbf{Undersampling:} Although balancing data, a decreased number of training samples shows a negative effect on performance. Performance for the minority class does not show significant improvement and significantly degrades the performance of the majority class, particularly for transformer-based models, which typically require more data to converge.
    \begin{itemize}
        \item Example: In mb-landmark\_cls, the minority class (Impact Ejecta) improved marginally from 0.00 to 0.03 with Inception, while the majority class (Others) accuracy dropped from 0.93 to 0.35. A similar trend was observed in mb-surface\_cls, where the minority class (Arm Cover) improved only from 0.00 to 0.01, while the majority class (Ground) performance dropped from 0.72 to 0.03.
    \end{itemize}
    \item \textbf{Oversampling:} Oversampling shows mixed results. It shows significant effectiveness in transformer-based models but is less consistent in convolutional models.
    \begin{itemize}
        \item Example: In the mb-change\_cls\_hirise dataset, ViT showed improvement in the minority class from 0.39 to 0.52, while the majority class remained almost unaffected (0.96 to 0.95). SwinT shows similarly stable behavior across datasets. In contrast, ResNet benefited from oversampling in some datasets (e.g., from 0.00 to 0.62 in mb-surface\_cls and 0.17 to 0.35 in mb-change\_hirise). However, in SqueezeNet, although it retains the performance of the minority class, it shows performance degradation in the majority class, dropping from 0.84 to 0.51 in mb-surface\_cls and from 0.32 to 0.18 in mb-change\_cls\_hirise.
    \end{itemize}
\end{itemize}

In summary, all these results indicate that there is no single data manipulation technique that is universally effective across all models and dataset combinations for handling class imbalance. This suggests that the choice of technique depends heavily on the specific characteristics of the dataset and the model being used. We believe this presents a significant opportunity for the community to develop more specialized solutions for imbalanced data in niche domains like planetary science, where class distributions can be highly diverse.

\begin{figure}
    \centering
    \includegraphics[width=0.99\textwidth]{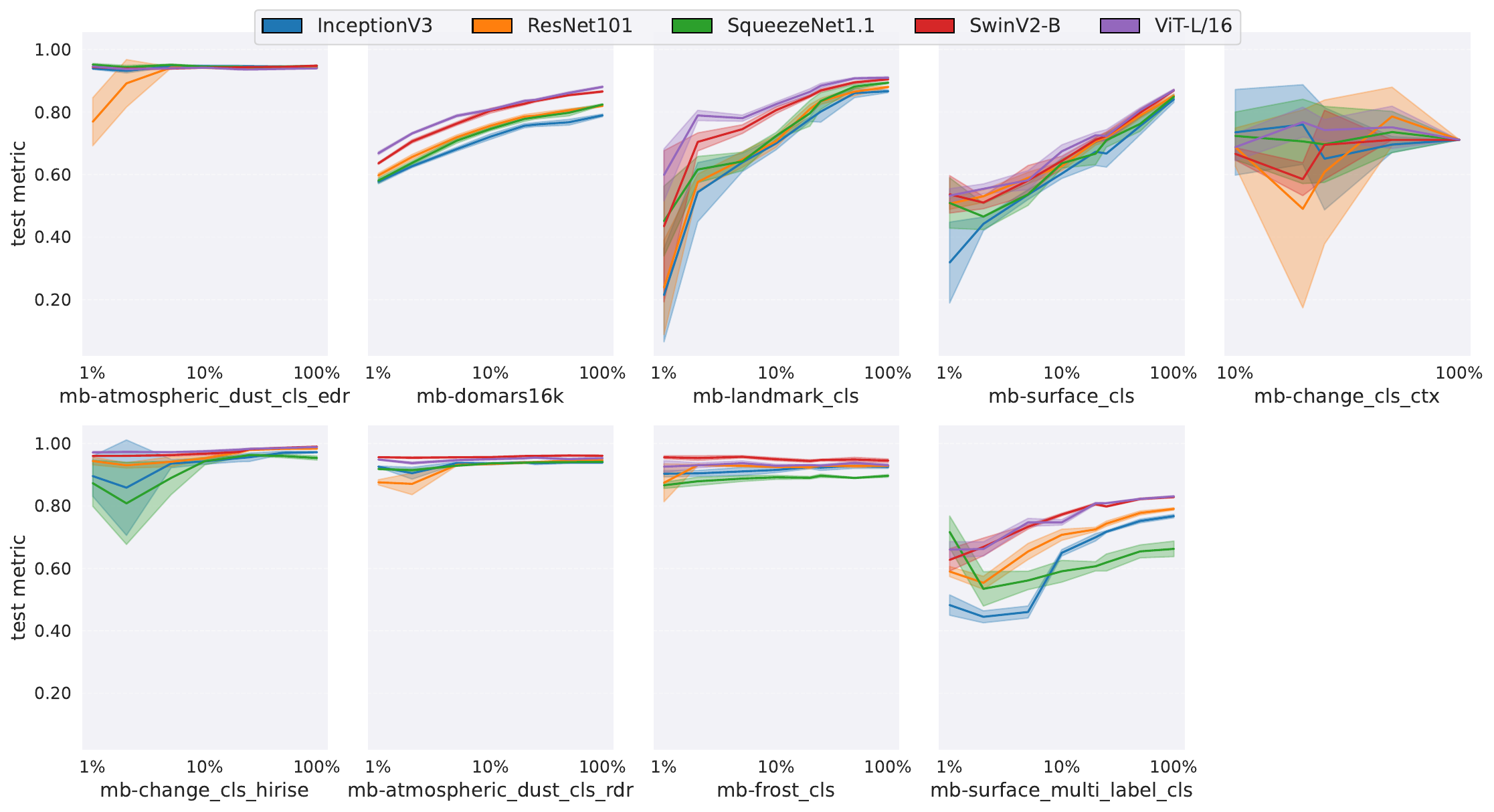}
    \caption{\textbf{Classification vs Train size under Feature Extraction setting:} Raw F1-score of baselines with a growing size (from 1\% to 100\%) of the training set. Shaded regions indicate confidence intervals over multiple runs. Note: Partitions in mb-change\_cls\_ctx start at 10\%.}
    \label{fig:results_classification_partition_actual_fe}
\end{figure}

\begin{figure}
    \centering
    \includegraphics[width=0.99\textwidth]{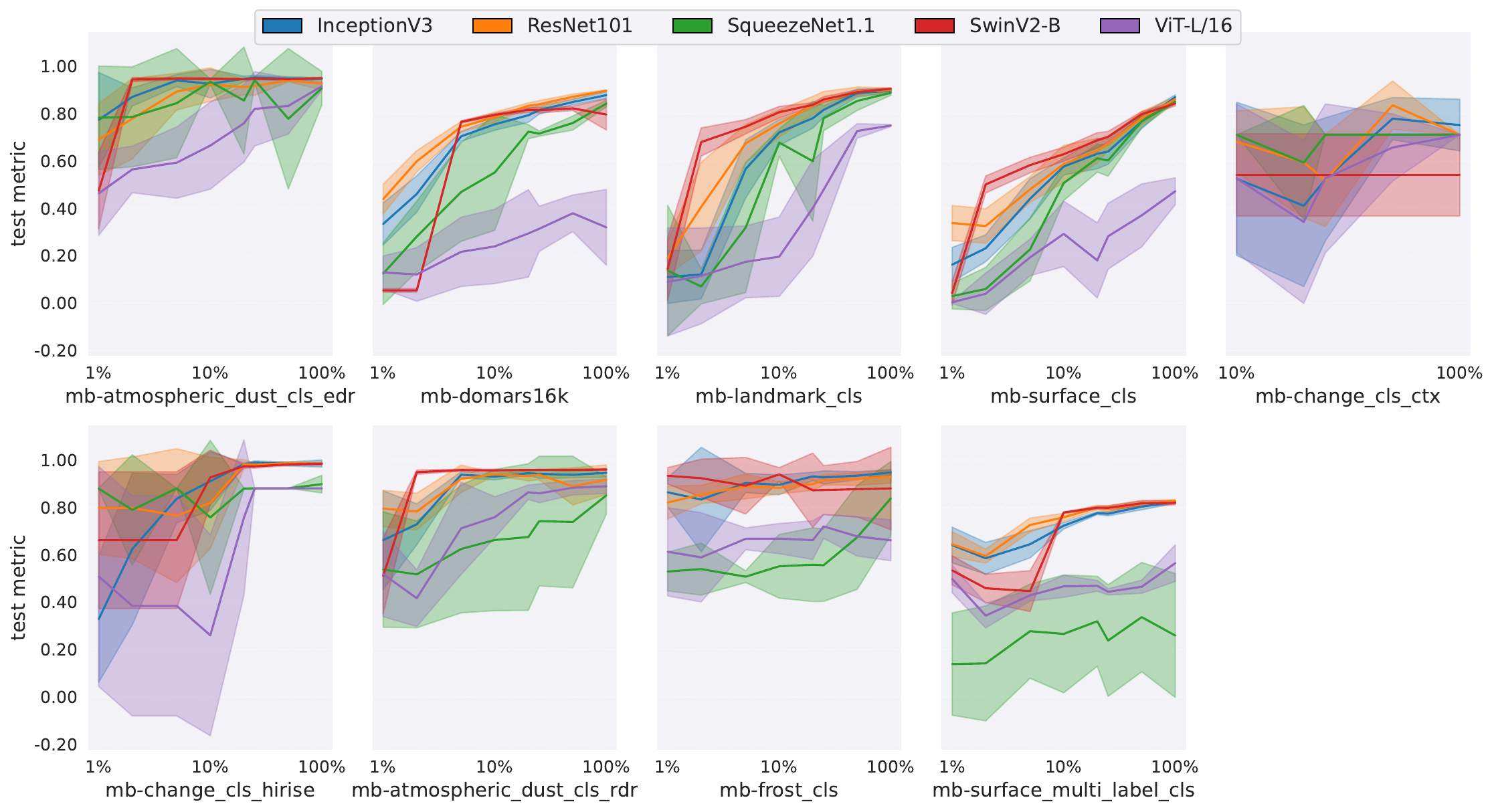}
    \caption{\textbf{Classification vs Train size under Transfer Learning setting:} Raw F1-score of baselines with a growing size (from 1\% to 100\%) of the training set. Shaded regions indicate confidence intervals over multiple runs. Note: Partitions in mb-change\_cls\_ctx start at 10\%.}
    \label{fig:results_classification_partition_actual_tl}
\end{figure}

\begin{figure}
    \centering
    \includegraphics[width=0.99\textwidth]{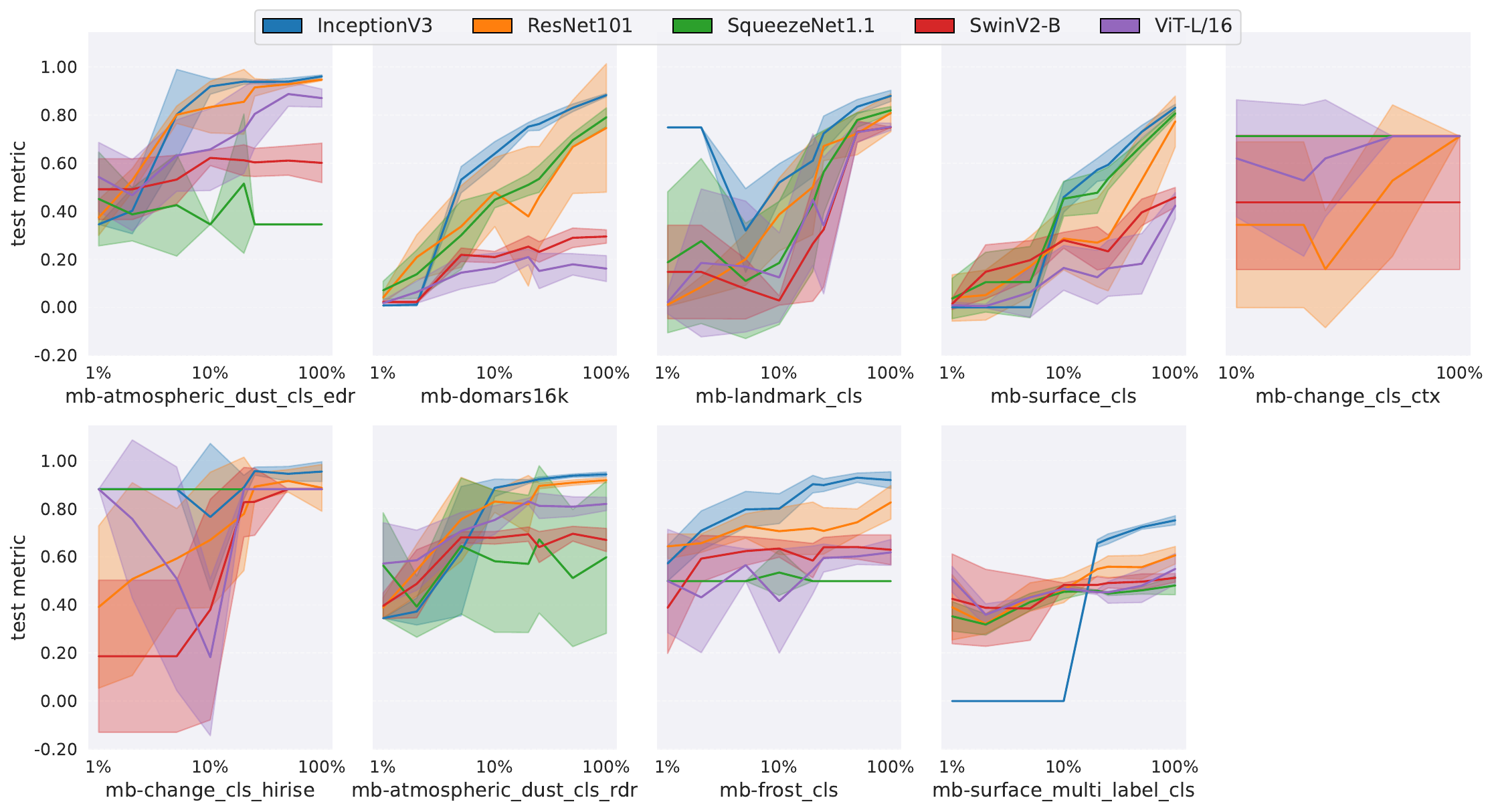}
    \caption{\textbf{Classification vs Train size with models trained from scratch:} Raw F1-score of baselines with a growing size (from 1\% to 100\%) of the training set. Shaded regions indicate confidence intervals over multiple runs. Note: Partitions in mb-change\_cls\_ctx start at 10\%.}
    \label{fig:results_classification_partition_actual_st}
\end{figure}

Figures \ref{fig:results_classification_partition_actual_fe}, \ref{fig:results_classification_partition_actual_tl}, and \ref{fig:results_classification_partition_actual_st} illustrate how classification performance varies with training set size across the three training strategies. Feature extraction shows rapid performance gains with increasing data and tends to saturate earlier. Transfer learning improves more gradually and typically requires more data to catch up. Training from scratch exhibits slower improvement and higher variability, especially on small datasets like \texttt{mb-change\_cls\_ctx}, where it often fails to generalize.

% Figure \ref{fig:results_classification_partition_actual_fe}, \ref{fig:results_classification_partition_actual_tl}, and \ref{fig:results_classification_partition_actual_st} show results of all classification datasets for training size vs performance based on F1-score for feature extraction, transfer learning, and trained from scratch training, respectively. Across all partitions, feature extraction shows faster performance gain with increasing data and saturates earlier, while transfer learning requires more data to catch up, and scratch improves slowly with high variability. On smaller datasets (e.g., mb-change\_cls\_ctx), only feature extraction achieves stable gains; scratch often fails to improve meaningfully.

\begin{figure}
    \centering
    \includegraphics[width=0.99\textwidth]{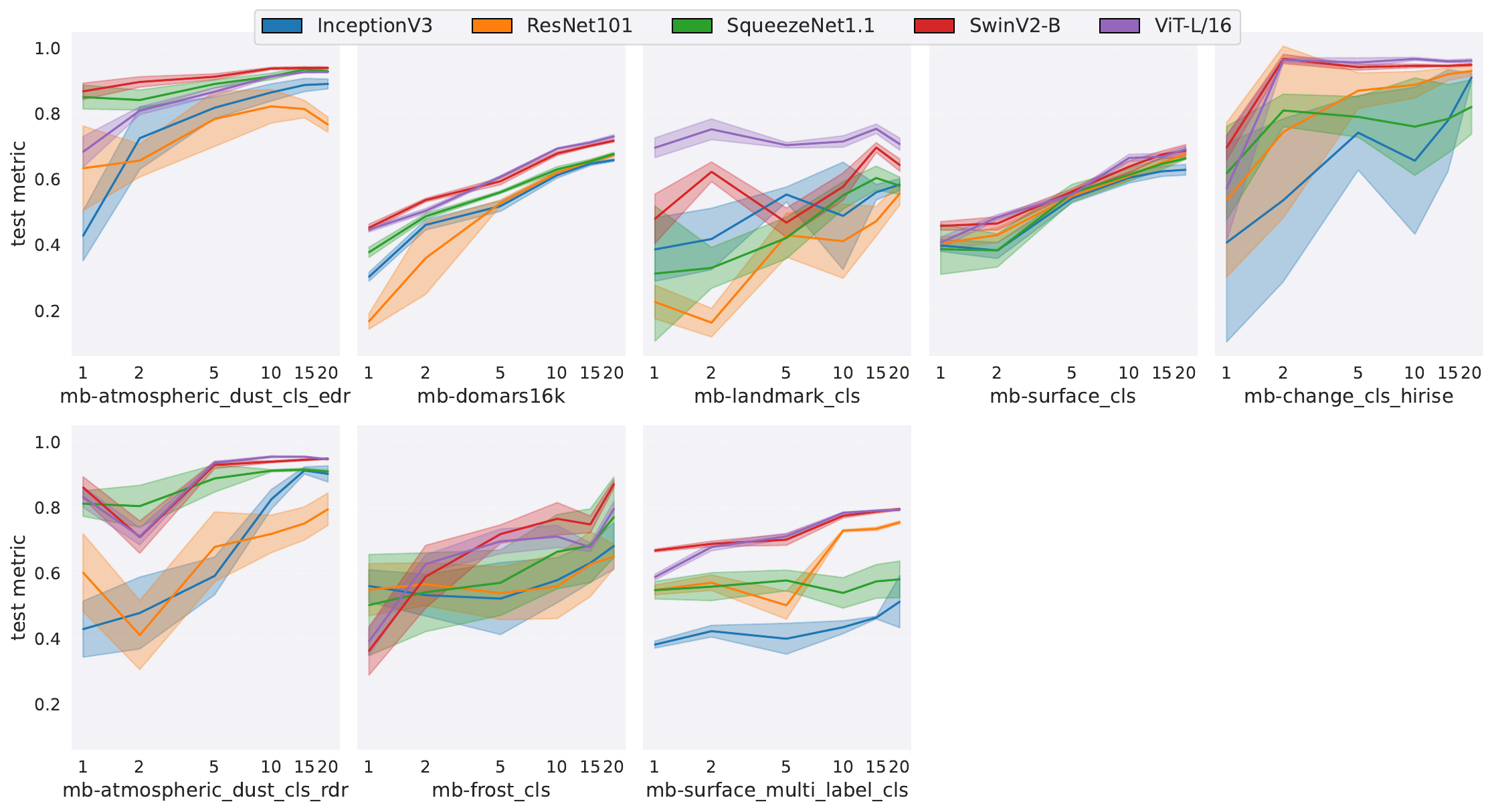}
    \caption{\textbf{Classification vs Few-shot under Feature Extraction setting:} Raw F1-score of baselines on few-shot setting. Shaded regions indicate confidence intervals over multiple runs.}
    \label{fig:results_classification_fewshot_actual_fe}
\end{figure}

\begin{figure}
    \centering
    \includegraphics[width=0.99\textwidth]{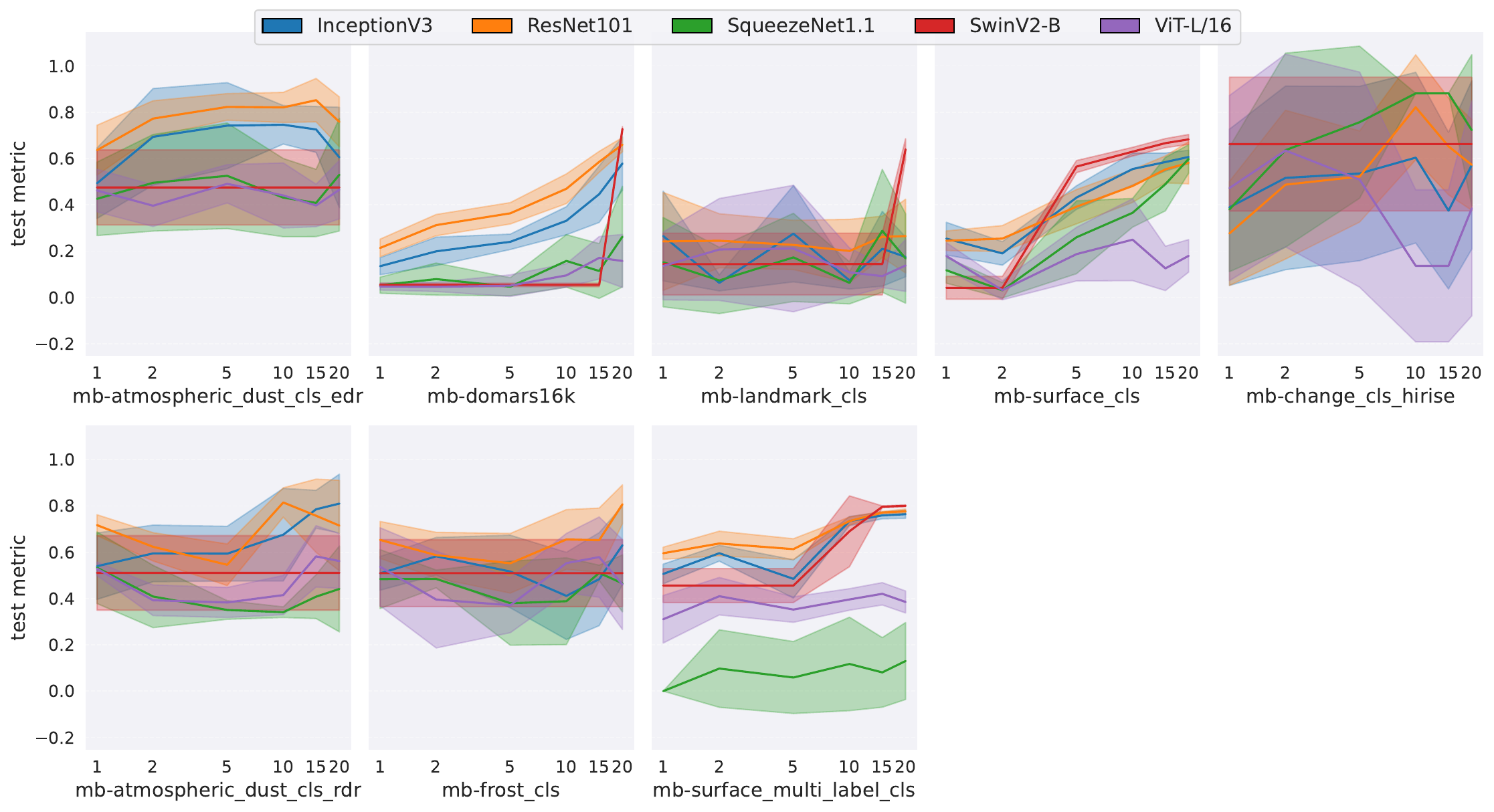}
    \caption{\textbf{Classification vs Few-shot under Transfer Learning setting:} Raw F1-score of baselines on few-shot setting. Shaded regions indicate confidence intervals over multiple runs.}
    \label{fig:results_classification_fewshot_actual_tl}
\end{figure}

\begin{figure}
    \centering
    \includegraphics[width=0.99\textwidth]{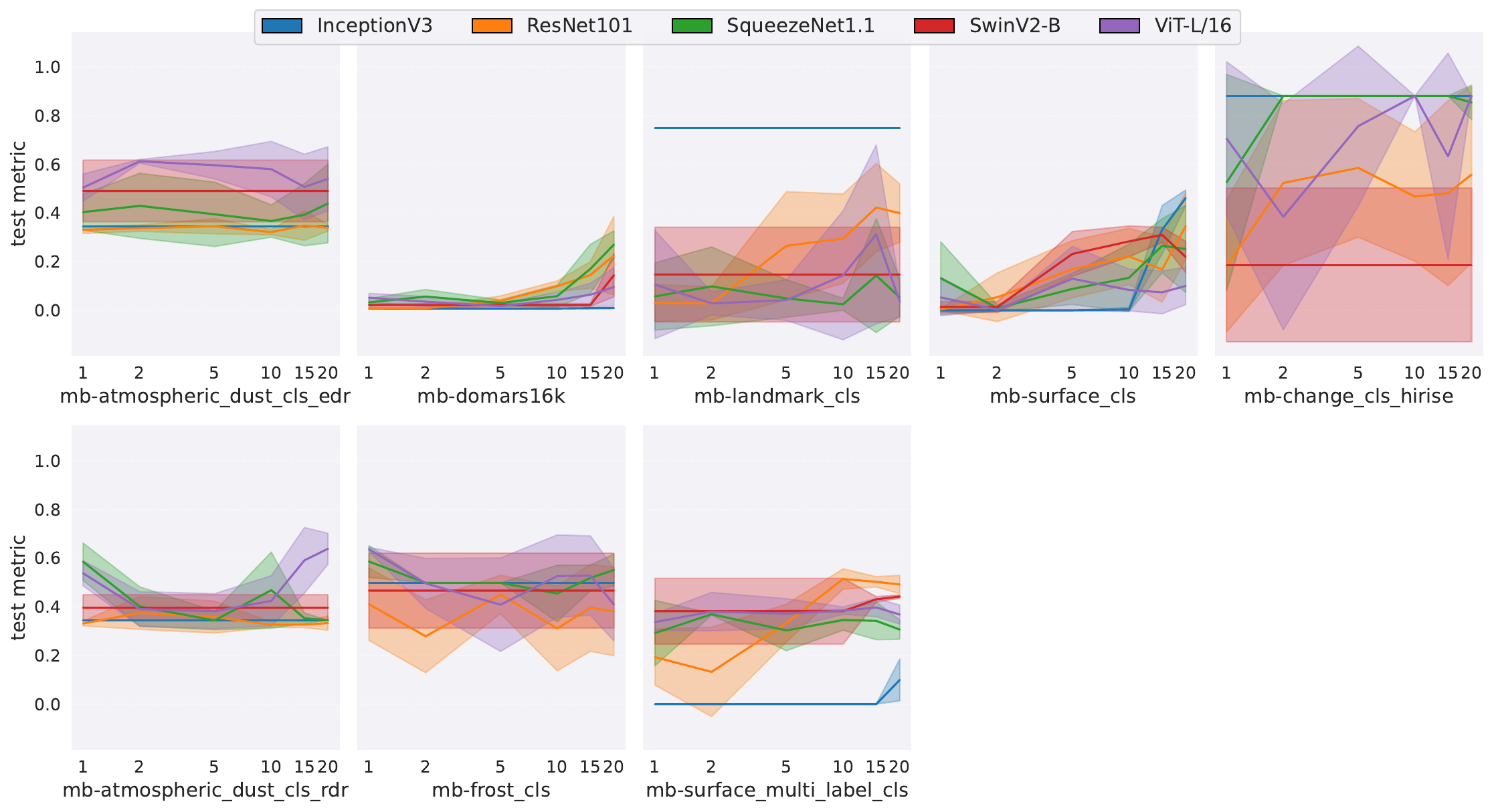}
    \caption{\textbf{Classification vs Few-shot with models trained from scratch:} Raw F1-score of baselines on few-shot setting. Shaded regions indicate confidence intervals over multiple runs.}
    \label{fig:results_classification_fewshot_actual_st}
\end{figure}

Figures \ref{fig:results_classification_fewshot_actual_fe}, \ref{fig:results_classification_fewshot_actual_tl}, and \ref{fig:results_classification_fewshot_actual_st} show few-shot learning results across the same training strategies. Consistent with earlier findings, feature extraction significantly outperforms the other approaches across all shot counts, demonstrating strong performance even with as few as 1–2 examples. Transfer learning performs moderately but remains inconsistent across datasets. Training from scratch struggles with very limited data and only starts to improve at 10+ shots, typically lagging far behind the other methods. Note that mb-change\_cls\_ctx does not have few-shot data due to its already limited dataset size.

% Note that \texttt{mb-change\_cls\_ctx} is excluded from this analysis due to its already limited dataset size.

% Figure \ref{fig:results_classification_fewshot_actual_fe}, \ref{fig:results_classification_fewshot_actual_tl}, and \ref{fig:results_classification_fewshot_actual_st} show results of all classification datasets for few-shot vs performance based on F1-score for feature extraction, transfer learning, and trained from scratch training, respectively. Same as the above 2 analyses, feature extraction clearly outperforms the other two across all shot counts, with acceptable performance even at 1–2 shots. Transfer learning performs moderately but is inconsistent across datasets, while scratch struggles at low shots and only shows improvement at 10+ shots, often lagging significantly behind. Please note that mb-change\_cls\_ctx does not have few-shot data due to its smaller size originally.

% All evaluations are performed using the same models and hyperparameters to ensure a fair comparison.

\subsection{Segmentation Results}
\label{subsec:segmentation_results}

Figures \ref{fig:results_segmentation_actual_fe}, \ref{fig:results_segmentation_actual_tl}, and \ref{fig:results_segmentation_actual_st} present segmentation results (IoU scores) across all datasets under feature extraction, transfer learning, and training from scratch settings, respectively. Feature extraction consistently achieves the highest IoU scores across models, with UNet and SegFormer performing particularly well. Transfer learning performs moderately but remains behind feature extraction, while training from scratch shows the lowest performance and highest instability, especially on challenging datasets like \texttt{mb-conequest\_seg} and \texttt{mb-mmls}.

% Figure \ref{fig:results_segmentation_normalized_fe}, \ref{fig:results_segmentation_normalized_tl}, and \ref{fig:results_segmentation_normalized_st} show results of all segmentation datasets based on IoU for feature extraction, transfer learning, and trained from scratch training, respectively. Results show that feature extraction delivers the best IoU scores across models, especially for UNet and SegFormer. Transfer learning trails behind, while scratch remains unstable with low IoU, particularly on mb-conequest\_seg and mb-mmls.

\begin{figure}[htbp]
    \centering
    \includegraphics[width=0.99\textwidth]{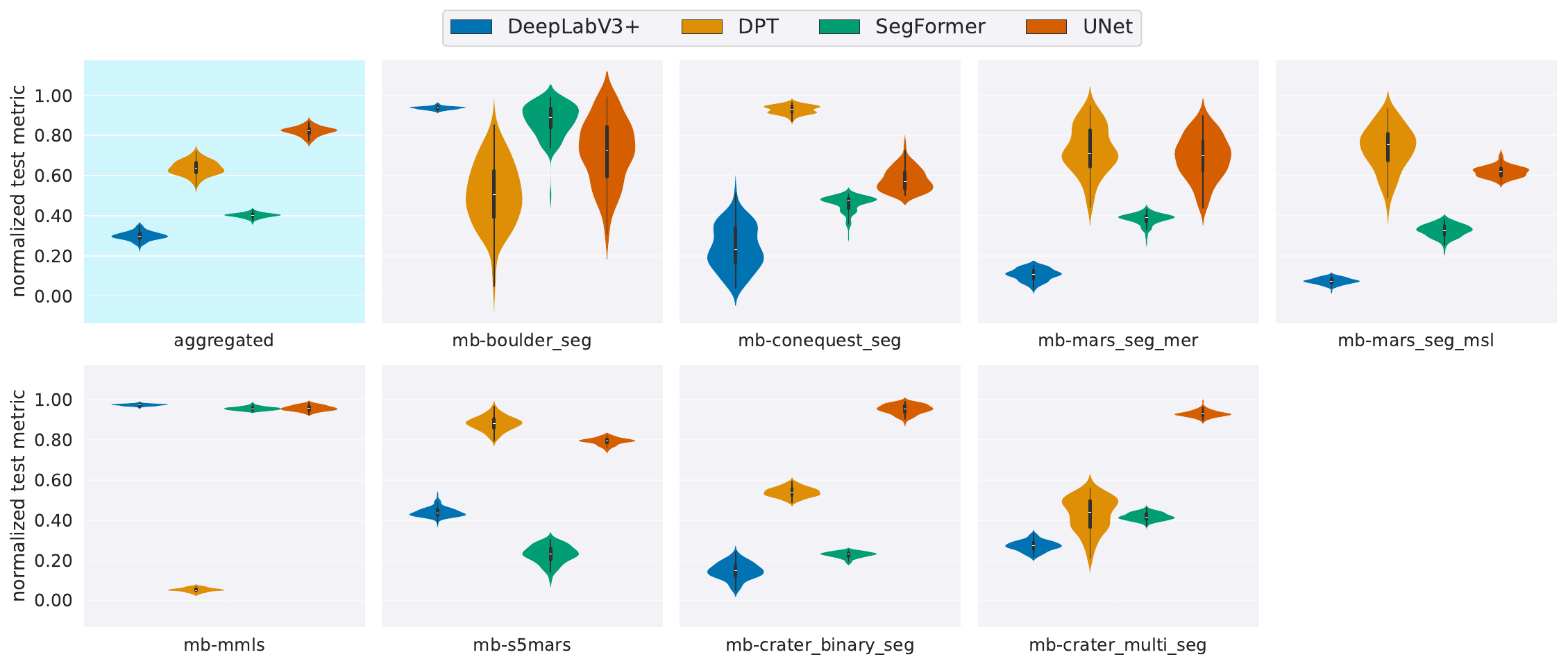}
    \caption{\textbf{Segmentation Benchmark under Feature Extraction setting:} Normalized IoU of various baselines (higher is better). Violin plots are obtained from bootstrap samples of normalized IQM (Section \ref{subsec:reporting_results_appendix}). The left plot reports the average across all tasks.}
    \label{fig:results_segmentation_normalized_fe}
\end{figure}

\begin{figure}
    \centering
    \includegraphics[width=0.99\textwidth]{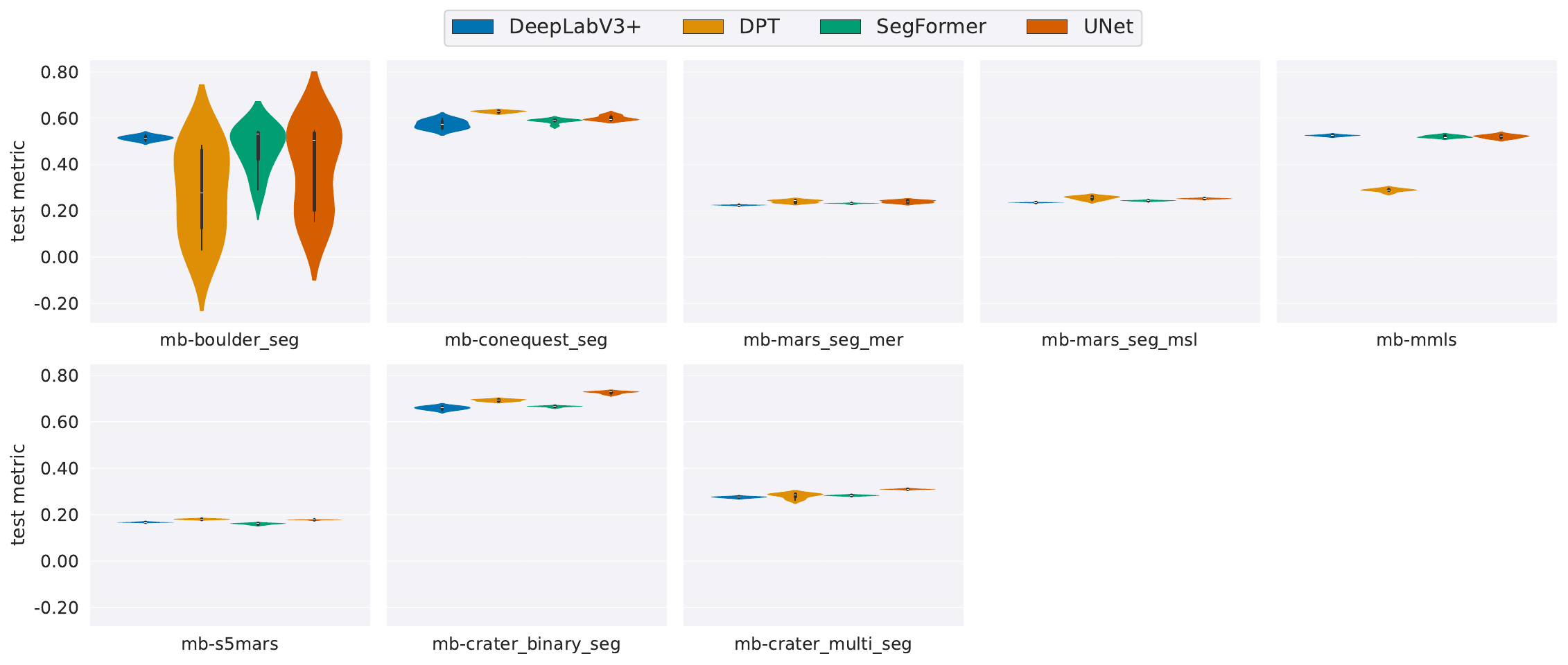}
    \caption{\textbf{Segmentation Benchmark under Feature Extraction setting:} Raw IoU of various baselines (higher is better). Violin plots represent the distribution of seeds.}
    \label{fig:results_segmentation_actual_fe}
\end{figure}

\begin{figure}
    \centering
    \includegraphics[width=0.99\textwidth]{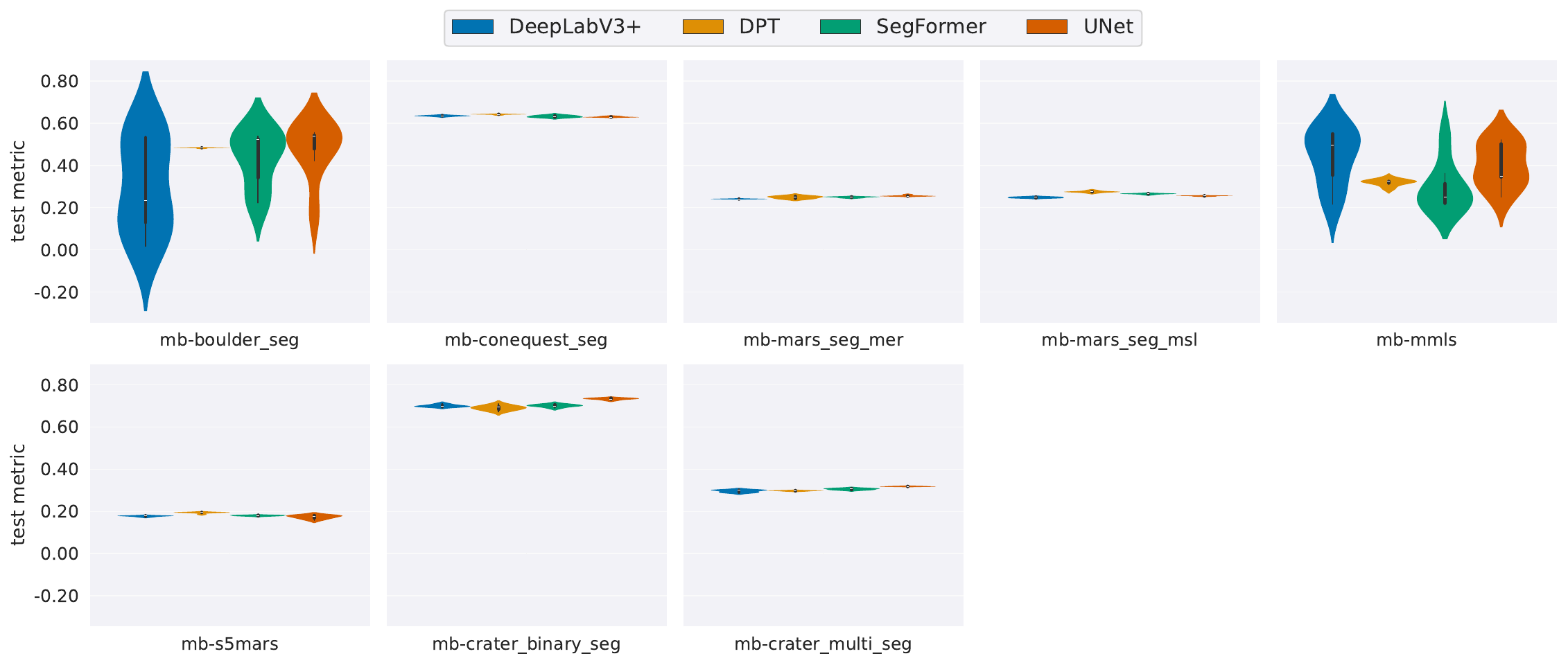}
    \caption{\textbf{Segmentation Benchmark under Transfer Learning setting:} Raw IoU of various baselines (higher is better). Violin plots represent the distribution of seeds.}
    \label{fig:results_segmentation_actual_tl}
\end{figure}

\begin{figure}
    \centering
    \includegraphics[width=0.99\textwidth]{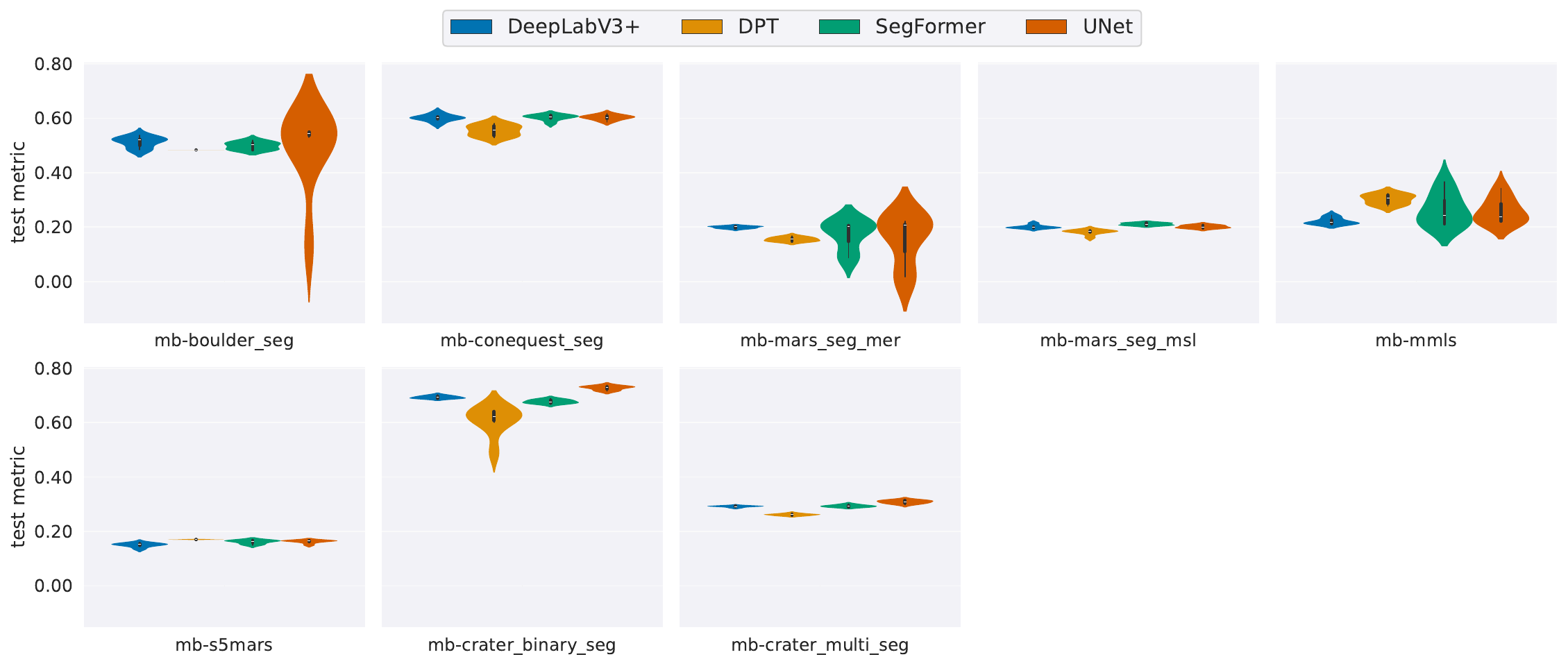}
    \caption{\textbf{Segmentation benchmark with models trained from scratch:} Raw IoU of various baselines (higher is better). Violin plots represent the distribution of seeds.}
    \label{fig:results_segmentation_actual_st}
\end{figure}

\begin{figure}
    \centering
    \includegraphics[width=0.99\textwidth]{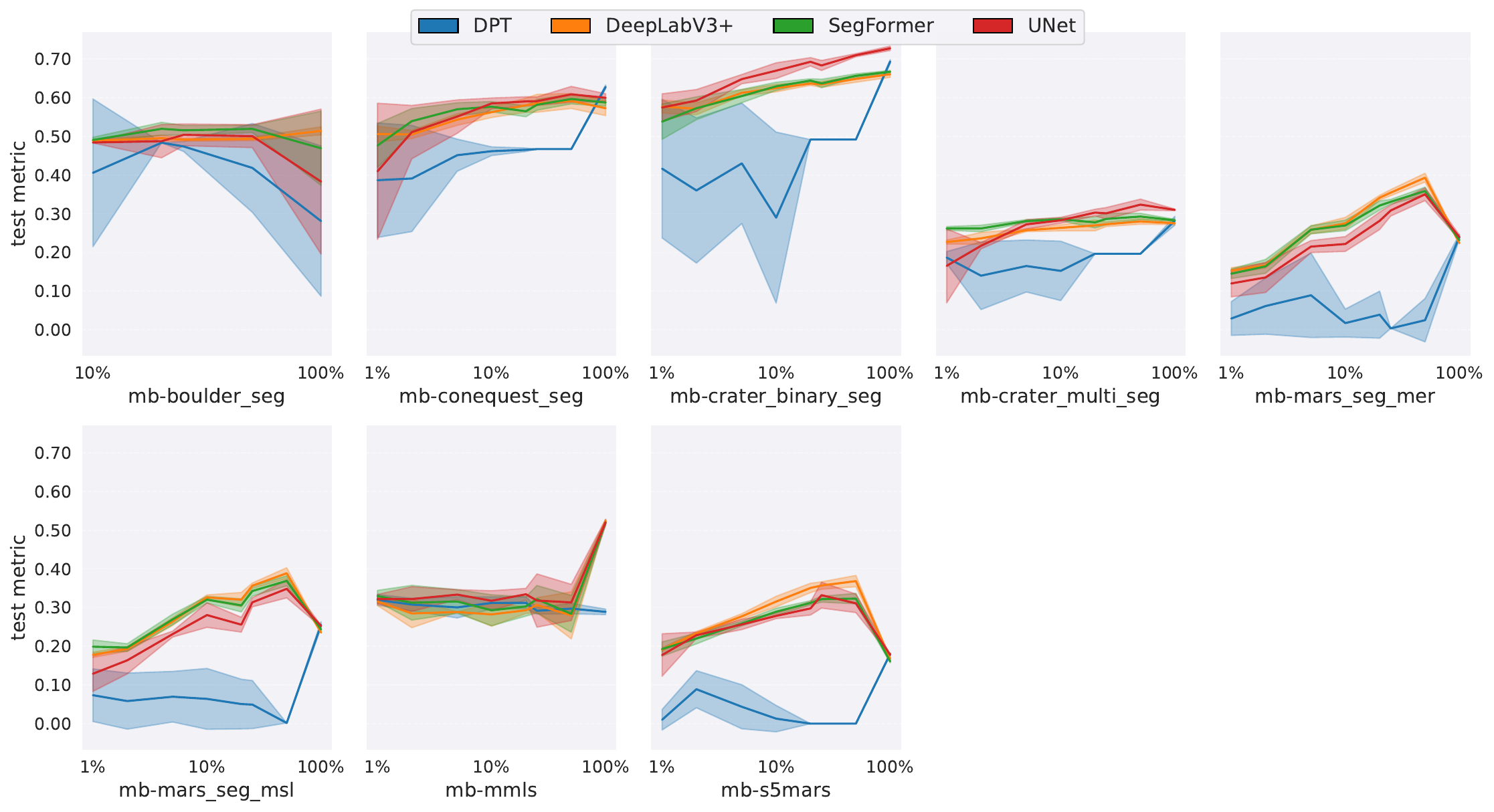}
    \caption{\textbf{Segmentation vs Train size under Feature Extraction setting:} Raw IoU of baselines with a growing size (from 1\% to 100\%) of the training set. Shaded regions indicate confidence intervals over multiple runs.}
    \label{fig:results_segmentation_partition_actual_fe}
\end{figure}

\begin{figure}
    \centering
    \includegraphics[width=0.99\textwidth]{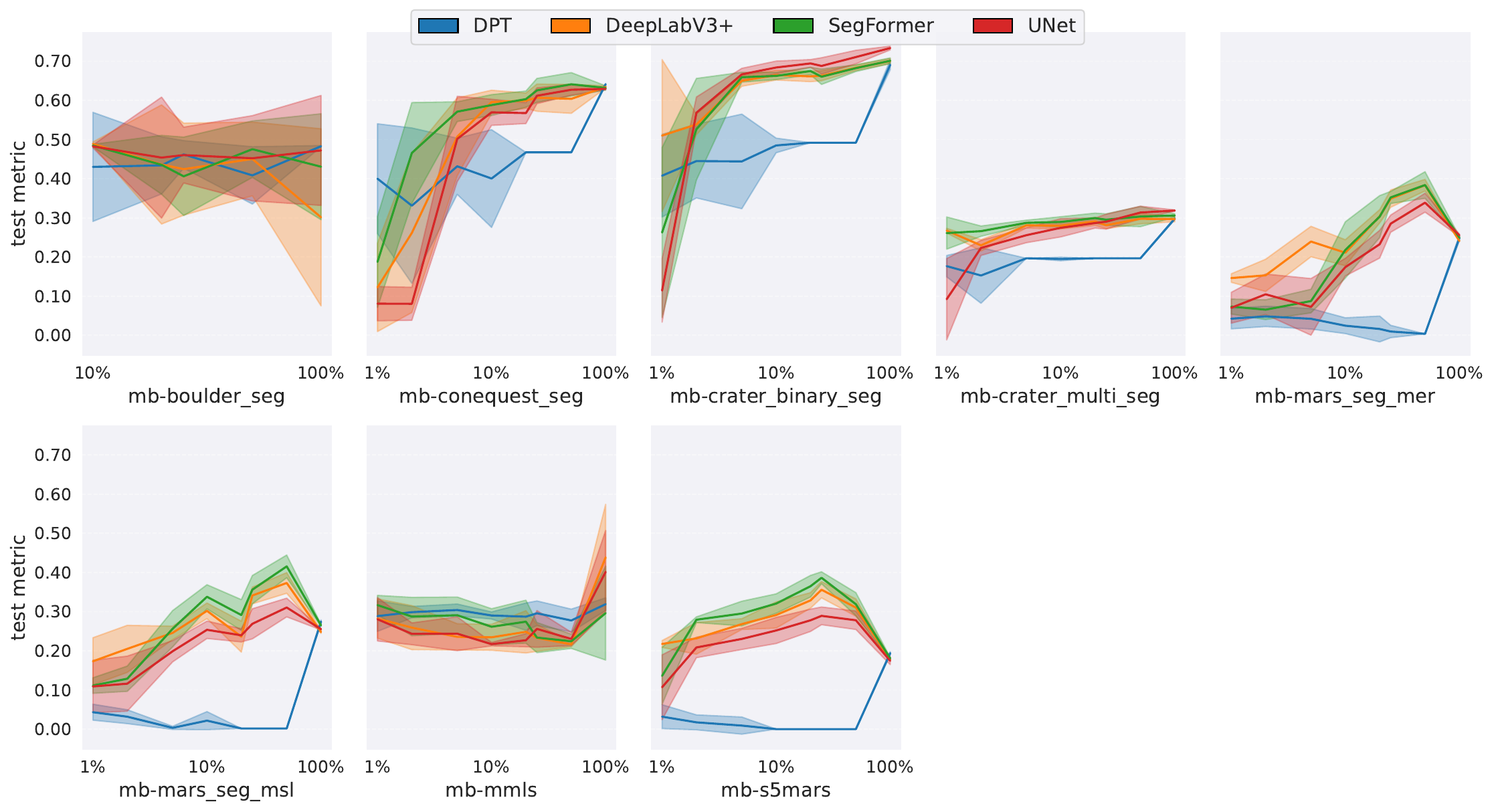}
    \caption{\textbf{Segmentation vs Train size under Transfer Learning setting:} Raw IoU of baselines with a growing size (from 1\% to 100\%) of the training set. Shaded regions indicate confidence intervals over multiple runs.}
    \label{fig:results_segmentation_partition_actual_tl}
\end{figure}

\begin{figure}
    \centering
    \includegraphics[width=0.99\textwidth]{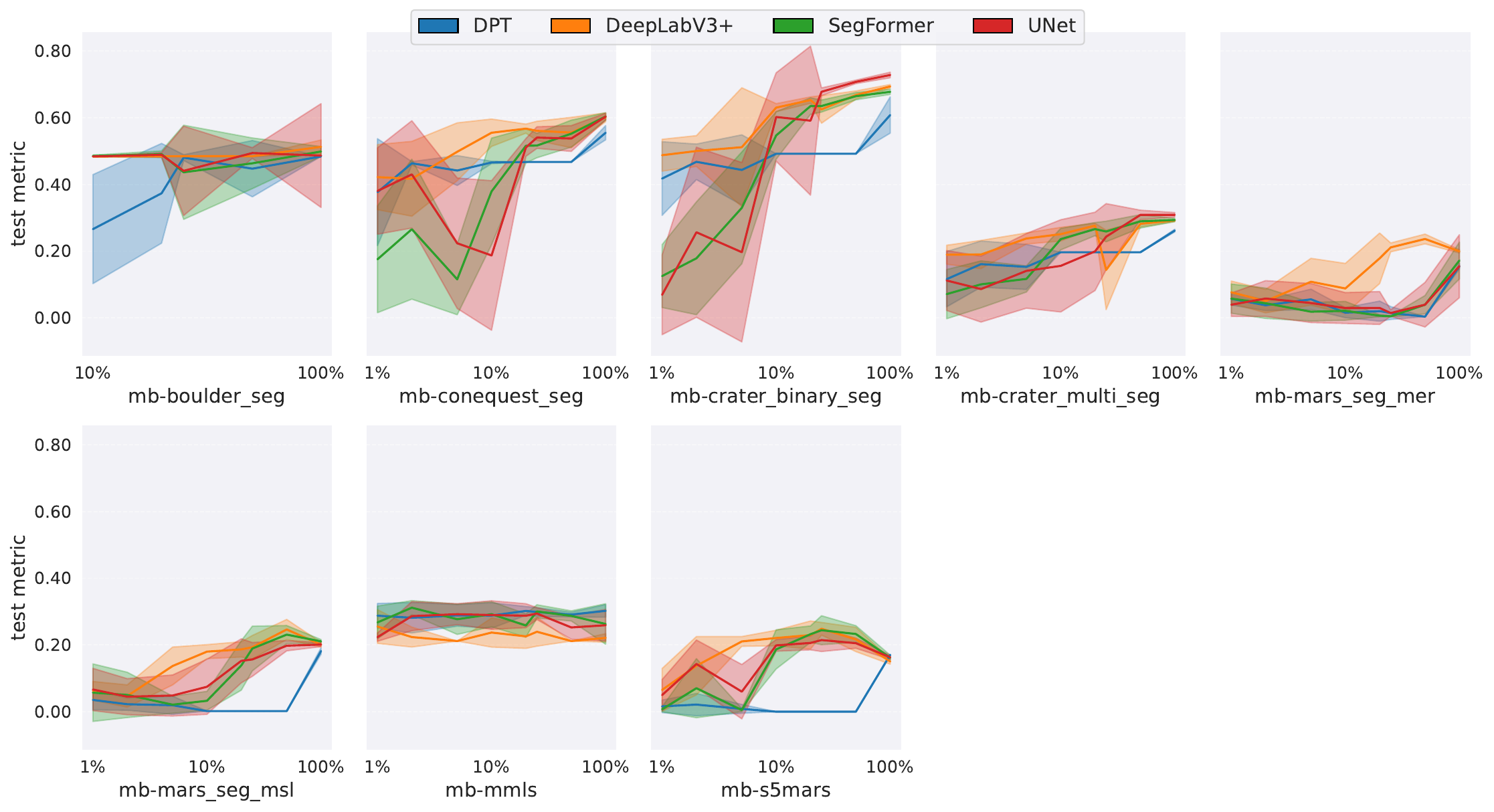}
    \caption{\textbf{Segmentation vs Train size with models trained from scratch:} Raw IoU of baselines with a growing size (from 1\% to 100\%) of the training set. Shaded regions indicate confidence intervals over multiple runs.}
    \label{fig:results_segmentation_partition_actual_st}
\end{figure}

Figures \ref{fig:results_segmentation_partition_actual_fe}, \ref{fig:results_segmentation_partition_actual_tl}, and \ref{fig:results_segmentation_partition_actual_st} show the impact of training set size on segmentation performance (IoU) for feature extraction, transfer learning, and training from scratch, respectively. As training size increases, feature extraction consistently yields higher and more stable performance across datasets. Transfer learning shows moderate gains but generally lags behind feature extraction. Training from scratch exhibits the lowest performance and highest variability, particularly on datasets with complex terrain or limited data availability, i.e., mb-mars\_seg\_mer, mb-mars\_seg\_msl, and mb-s5mars.

% Figure \ref{fig:results_segmentation_partition_actual_fe}, \ref{fig:results_segmentation_partition_actual_tl}, and \ref{fig:results_segmentation_partition_actual_st} show results of all segmentation datasets for training size vs performance based on IoU for feature extraction, transfer learning, and trained from scratch training, respectively. With increasing training size, Feature Extraction consistently performs better and stabilizes faster. Transfer Learning shows moderate improvement, while Scratch remains noisy and underperforms, especially on datasets with complex terrain or small training sets.

\subsection{Object Detection Results}
\label{subsec:object_detection_results}

\begin{figure}
    \centering
    \includegraphics[width=0.99\textwidth]{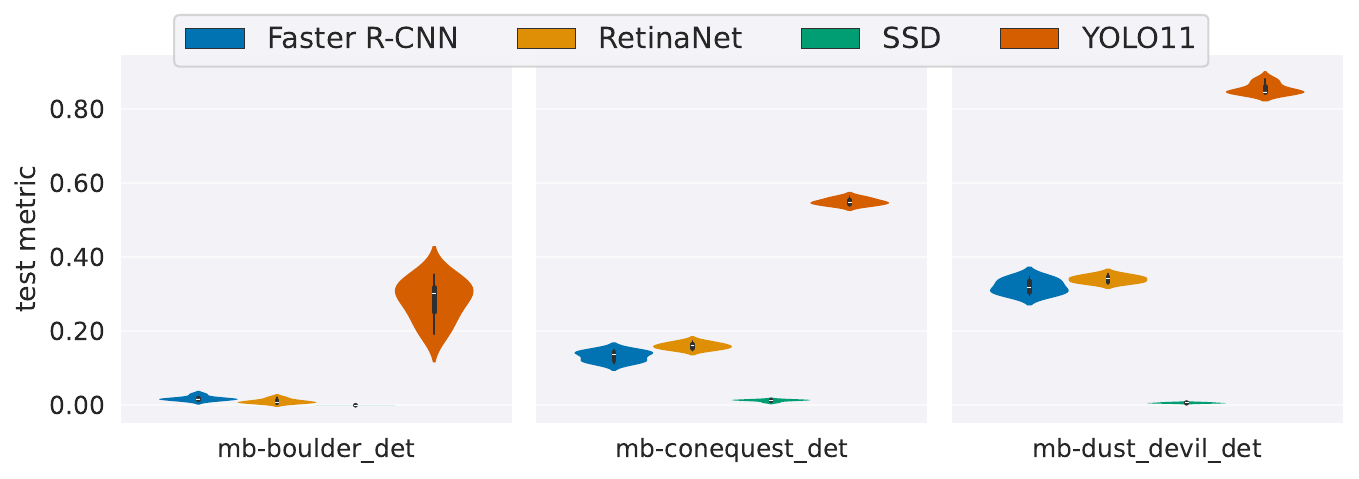}
    \caption{\textbf{Object Detection Benchmark under Feature Extraction setting:} Raw mAP of various baselines (higher is better). Violin plots represent the distribution of seeds.}
    \label{fig:results_object_detection_actual_fe}
\end{figure}

\begin{figure}
    \centering
    \includegraphics[width=0.99\textwidth]{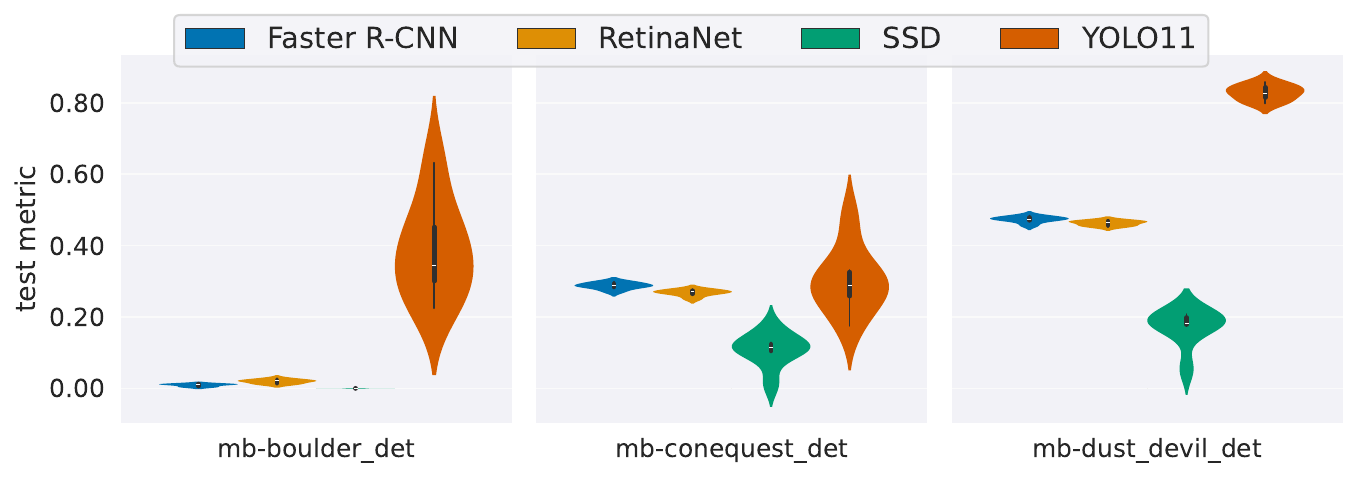}
    \caption{\textbf{Object Detection Benchmark under Transfer Learning setting:} Raw mAP of various baselines (higher is better). Violin plots represent the distribution of seeds.}
    \label{fig:results_object_detection_actual_tl}
\end{figure}

\begin{figure}
    \centering
    \includegraphics[width=0.99\textwidth]{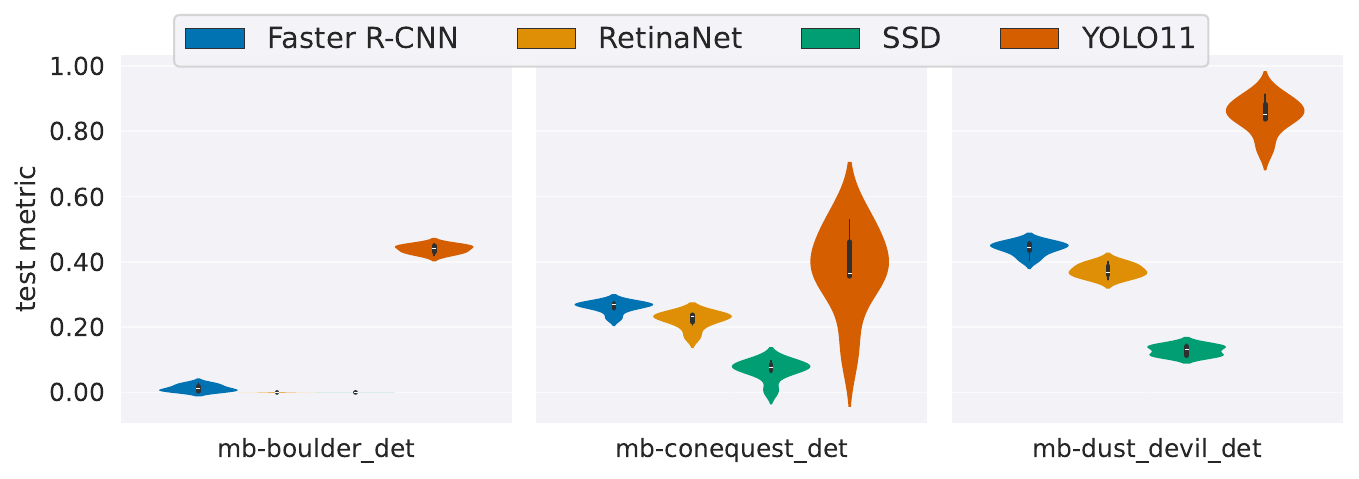}
    \caption{\textbf{Object Detection benchmark with models trained from scratch:} Raw mAP of various baselines (higher is better). Violin plots represent the distribution of seeds.}
    \label{fig:results_object_detection_actual_st}
\end{figure}

Figures \ref{fig:results_object_detection_actual_fe}, \ref{fig:results_object_detection_actual_tl}, and \ref{fig:results_object_detection_actual_st} present object detection results (mAP) across all datasets under feature extraction, transfer learning, and training from scratch settings, respectively. Feature Extraction achieves the best and most consistent mAP scores across all three detection datasets. Transfer learning performs slightly better than training from scratch, but both show high variability and generally low performance.

% Detection performance is particularly weak on mb-boulder\_det and mb-dust\_devil\_det.

% These challenges are primarily due to several factors:

% \begin{itemize}
%     \item The overall dataset size is significantly smaller compared to classification and segmentation datasets.
%     \item The number of objects per image is low, with many images containing only one or even zero target objects.
%     \item The grayscale nature of the imagery limits visual cues, and low object–background contrast (e.g., in dust devil detection) further complicates learning.
% \end{itemize}

As noted in Section \ref{subsec:analysis_1}, due to the consistently poor performance even on the full datasets, we did not perform partition-based experiments for object detection. We leave this open for the community to explore methods that improve detection under such constrained, low-data conditions.

% \begin{figure}
%     \centering
%     \includegraphics[width=0.99\textwidth]{figures/appendix_object_detection/detection_normalized_fe.pdf}
%     \caption{\textbf{Object Detection Benchmark under Feature Extraction setting:} Normalized mAP of various baselines (higher is better). Violin plots are obtained from bootstrap samples of normalized IQM (Section \ref{subsec:reporting_results_appendix}). The left plot reports the average across all tasks.}
%     \label{fig:results_object_detection_normalized_fe}
% \end{figure}

\newpage

\section{Prompts for VLM Evaluation}
\label{sec:prompts}

In this section, we provide the system instructions and prompts used for all six datasets evaluated on vision-language models (VLMs) in Section 5.4.

\subsection*{mb-domars16k}

\begin{tcolorbox}[
    colback=gray!5,
    colframe=gray!60!black,
    title=System Instructions,
    fonttitle=\bfseries,
    sharp corners=south,
    boxrule=1pt,
    breakable,
    enhanced
]
You are an expert Martian geologist AI. Your task is to classify Martian surface landform images. You will be provided with an image of a Martian surface landform.

You must respond with \textbf{ONLY the three-letter abbreviation} of the most prominent landform class present in the image.

Here are the possible landform classes, their abbreviations, and definitions:
\vspace{0.5em}

\textbf{Aeolian Bedforms:}
\begin{itemize}[left=2em]
    \item \textbf{(ael)} Aeolian Curved: Wind-formed bedforms with a curved, dune-like, or rippled appearance.
    \item \textbf{(aec)} Aeolian Straight: Wind-formed bedforms with a straight, linear, or elongated ridge-like appearance.
\end{itemize}

\textbf{Topographic Landforms:}
\begin{itemize}[left=2em]
    \item \textbf{(cli)} Cliff: A steep, near-vertical, or very abrupt rock exposure or slope.
    \item \textbf{(rid)} Ridge: An elongated, narrow elevation or crest of land.
    \item \textbf{(fsf)} Channel: A depression, groove, or trough, often suggesting past fluid flow (e.g., water or lava).
    \item \textbf{(sfe)} Mounds: Distinct, rounded, or irregularly shaped raised landforms or protuberances.
\end{itemize}

\textbf{Slope Feature Landforms:}
\begin{itemize}[left=2em]
    \item \textbf{(fsg)} Gullies: Small, incised channels or ravines, typically found on slopes, potentially formed by fluid or debris flows.
    \item \textbf{(fse)} Slope Streaks: Dark or light markings that appear on slopes, often attributed to dry granular flows or small avalanches.
    \item \textbf{(fss)} Mass Wasting: Features resulting from the downslope movement of rock, regolith, and soil under gravity (e.g., landslides, slumps).
\end{itemize}

\textbf{Impact Landforms:}
\begin{itemize}[left=2em]
    \item \textbf{(cra)} Crater: A bowl-shaped depression, typically circular or sub-circular, formed by an impact event.
    \item \textbf{(sfx)} Crater Field: An area characterized by a significant concentration or cluster of impact craters.
\end{itemize}

\textbf{Basic Terrain Landforms:}
\begin{itemize}[left=2em]
    \item \textbf{(mix)} Mixed Terrain: An area exhibiting a combination of characteristics from multiple distinct landform types, without one single dominant type.
    \item \textbf{(rou)} Rough Terrain: An area characterized by irregular, uneven, broken, or difficult-to-traverse surfaces.
    \item \textbf{(smo)} Smooth Terrain: An area characterized by relatively even, regular surfaces with little to no significant relief or texture.
    \item \textbf{(tex)} Textured Terrain: An area exhibiting a distinct or noticeable surface pattern, fabric, or texture that is not clearly one of the more specific landforms.
\end{itemize}

\vspace{0.5em}
Analyze the provided image and output \textbf{only the three-letter abbreviation} for the dominant landform.
\end{tcolorbox}
% Prompt (Bottom, visually "connected")
\begin{tcolorbox}[
    colback=gray!5,
    colframe=gray!60!black,
    title=Prompt,
    fonttitle=\bfseries,
    sharp corners=south,
    boxrule=1pt,
    breakable,
    enhanced
]
Classify the Martian surface landform in the following image.

Strictly use this format:

\textbf{Reasoning:} [step-by-step reasoning]

\textbf{Answer:} [Provide only the three-letter abbreviation for the dominant landform type]
\end{tcolorbox}

\subsection*{mb-surface\_cls}

% System Instructions (Top Box)
\begin{tcolorbox}[
    colback=gray!5,
    colframe=gray!60!black,
    title=System Instructions,
    fonttitle=\bfseries,
    sharp corners=south,
    boxrule=1pt,
    breakable,
    enhanced
]
You are an expert Martian surface classification AI. Your task is to classify Mars rover images into one of the scientific or engineering categories. You will be provided with an image captured by the Curiosity Rover's Mastcam or MAHLI instruments.

Your job is to visually analyze the image and identify the dominant object or surface class that best describes the main content shown.

You must respond with \textbf{ONLY the three-letter abbreviation} of the most appropriate class.

Here are the possible classes, their abbreviations, and their descriptions:
\begin{itemize}[left=2em]
    \item \textbf{(apx)} Alpha Particle X-Ray Spectrometer (APXS): Element analysis instrument mounted on the rover's robotic arm.
    \item \textbf{(act)} Alpha Particle X-Ray Spectrometer Calibration Target (APXS CT): Standard target for APXS instrument calibration.
    \item \textbf{(arm)} Arm Cover: Structural component covering parts of the robotic arm.
    \item \textbf{(art)} Artifact: Unusual or foreign features not naturally occurring on Mars.
    \item \textbf{(cct)} ChemCam Calibration Target: Laser calibration target used by ChemCam.
    \item \textbf{(cio)} CheMin Inlet Open: The inlet area of the CheMin instrument in open position.
    \item \textbf{(clr)} Close-Up Rock: Rock surfaces captured at close proximity to reveal texture.
    \item \textbf{(dls)} Distant Landscape: Martian terrain visible far from the rover's immediate location.
    \item \textbf{(dri)} Drill: The rover's drill tool, used to bore into Martian rock.
    \item \textbf{(drh)} Drill Holes: Resulting holes left after drilling into the Martian surface.
    \item \textbf{(drp)} Dust Removal Tool Spot: Brushed area exposed by the DRT cleaning tool.
    \item \textbf{(drt)} Dust Removal Tool: The brushing tool mounted on the arm to remove surface dust.
    \item \textbf{(flr)} Float Rock: Detached rocks lying loosely on the surface.
    \item \textbf{(gro)} Ground: Flat, featureless terrain directly surrounding the rover.
    \item \textbf{(hor)} Horizon: Distant skyline visible in landscape images.
    \item \textbf{(inl)} Inlet: Sample intake ports for rover's internal instruments.
    \item \textbf{(lar)} Layered Rock: Rock formations showing visible sedimentary layers.
    \item \textbf{(ltv)} Light-Toned Veins: Bright mineral veins possibly formed by fluid activity.
    \item \textbf{(mah)} MAHLI: The Mars Hand Lens Imager camera itself.
    \item \textbf{(mct)} MAHLI Calibration Target: Calibration board for the MAHLI camera.
    \item \textbf{(mas)} Mastcam: The main mast-mounted camera used for panoramic imaging.
    \item \textbf{(mca)} Mastcam Calibration Target: Target board for Mastcam image calibration.
    \item \textbf{(nsk)} Night Sky: The Martian sky captured during night or low light.
    \item \textbf{(obt)} Observation Tray: Platform used for holding or inspecting sampled material.
    \item \textbf{(pbo)} Portion Box: Compartment for storing soil or rock samples.
    \item \textbf{(ptu)} Portion Tube: Tube system used in handling and measuring material portions.
    \item \textbf{(pto)} Portion Tube Opening: The visible end of a portioning tube.
    \item \textbf{(rem)} REMS-UV Sensor (REMS-UV): The UV radiation sensor from the environmental monitoring suite.
    \item \textbf{(rrd)} Rover Rear Deck: The back platform of the rover, often showing structural parts.
    \item \textbf{(san)} Sand: Fine-grained Martian soil, often seen in dunes or ripples.
    \item \textbf{(sco)} Scoop: Tool used to collect loose surface material.
    \item \textbf{(sun)} Sun: The solar disk, typically visible in calibration or sky images.
    \item \textbf{(tur)} Turret: The rotating tool assembly at the end of the robotic arm.
    \item \textbf{(whe)} Wheel: One of the rover's mobility wheels.
    \item \textbf{(whj)} Wheel Joint: The mechanical joint connecting the wheel to the suspension.
    \item \textbf{(wht)} Wheel Tracks: Imprints left by the wheels in the Martian soil.
\end{itemize}

\vspace{0.5em}
Analyze the provided image and respond with only the three-letter abbreviation of the dominant class.
\end{tcolorbox}

% Prompt (Bottom Box, visually "connected")
\begin{tcolorbox}[
    colback=gray!5,
    colframe=gray!60!black,
    title=Prompt,
    fonttitle=\bfseries,
    sharp corners=south,
    boxrule=1pt,
    breakable,
    enhanced
]
Classify the primary subject in the following image from the Curiosity Rover.

Strictly use this format:

\textbf{Reasoning:} [step-by-step reasoning]

\textbf{Answer:} [Provide only the three-letter abbreviation of the class]
\end{tcolorbox}

\subsection*{mb-mars\_seg\_msl}

% System Instructions
\begin{tcolorbox}[
    colback=gray!5,
    colframe=gray!60!black,
    title=System Instructions,
    fonttitle=\bfseries,
    sharp corners=south,
    boxrule=1pt,
    breakable,
    enhanced
]
You are an expert Martian geologist AI. Your task is to identify all relevant terrain classes present in Martian surface images. You will be provided with an image of the Martian surface.

You must respond with the corresponding integers for all applicable classes. \textbf{Note:} A single image may contain multiple classes.

Below are the possible classes, their corresponding integer labels, and definitions:
\vspace{0.5em}
\begin{itemize}[left=2em]
    \item \textbf{0: Background} \\
    Areas that do not contain relevant terrain features or objects of interest; typically undefined or used as a default label.
    \item \textbf{1: Bedrock} \\
    Exposed, solid rock surfaces that are generally flat or massive, forming the foundational layer of the terrain with minimal loose material.
    \item \textbf{2: Gravel / Sand / Soil} \\
    Loose surface materials such as gravel (small rocks), sand (fine particles), and soil (organic or inorganic matter), typically covering natural ground surfaces.
    \item \textbf{3: Rock} \\
    Isolated or clustered rock fragments distinguishable from continuous bedrock; often angular and scattered across the surface.
    \item \textbf{4: Shadow} \\
    Darkened regions caused by obstructions blocking direct light. These are not terrain features themselves but affect the visual appearance of the surface.
    \item \textbf{5: Sky / Distant Mountains} \\
    The upper portion of the scene representing the sky or far-off mountainous terrain; often hazy or blue in appearance.
    \item \textbf{6: Track} \\
    Visible marks or paths created by vehicle wheels or movement, usually appearing as grooves or parallel lines in soil, sand, or gravel.
\end{itemize}

\vspace{0.5em}
Analyze the provided image and return a list of integers representing all terrain classes visible in the image.
\end{tcolorbox}

% Prompt
\begin{tcolorbox}[
    colback=gray!5,
    colframe=gray!60!black,
    title=Prompt,
    fonttitle=\bfseries,
    sharp corners=south,
    boxrule=1pt,
    breakable,
    enhanced
]
Classify the Martian surface features in the following image.

Strictly use the format below:

\textbf{Reasoning:} [Step-by-step explanation of how you identified the classes]

\textbf{Answer:} [List of integers corresponding to the identified classes]
\end{tcolorbox}

\subsection*{mb-atmospheric\_dust\_cls\_edr }

% System Instructions
\begin{tcolorbox}[
    colback=gray!5,
    colframe=gray!60!black,
    title=System Instructions,
    fonttitle=\bfseries,
    sharp corners=south,
    boxrule=1pt,
    breakable,
    enhanced
]
You are an expert Martian atmospheric science AI. Your task is to analyze image patches captured by the HiRISE instrument on the Mars Reconnaissance Orbiter (MRO) and determine whether the surface view is obscured by atmospheric dust.

You will be provided with a HiRISE image patch of the Martian surface. Your job is to visually analyze the image and identify whether it appears dusty or not.

Respond with only the class name that best describes the image content: \texttt{dusty} or \texttt{not\_dusty}.

Here are the possible classes and their descriptions:
\begin{itemize}[left=2em]
    \item \textbf{dusty}: The image is heavily obscured by atmospheric dust, making surface details difficult or impossible to see.
    \item \textbf{not\_dusty}: The image is clear, and surface features are unobstructed by dust in the atmosphere.
\end{itemize}

\vspace{0.5em}
Analyze the provided image and respond with only one of the two class labels.\\
You must respond with exactly one of the following two lowercase class names: \texttt{dusty} or \texttt{not\_dusty}.
\end{tcolorbox}

% Prompt
\begin{tcolorbox}[
    colback=gray!5,
    colframe=gray!60!black,
    title=Prompt,
    fonttitle=\bfseries,
    sharp corners=south,
    boxrule=1pt,
    breakable,
    enhanced
]
Classify the following high-resolution image of the Martian surface.

Strictly use this format:

\textbf{Reasoning:} [step-by-step reasoning]

\textbf{Answer:} [Provide only one of the two class names: \texttt{dusty} or \texttt{not\_dusty}]
\end{tcolorbox}

\subsection*{mb-crater\_multi\_seg}

% System Instructions
\begin{tcolorbox}[
    colback=gray!5,
    colframe=gray!60!black,
    title=System Instructions,
    fonttitle=\bfseries,
    sharp corners=south,
    boxrule=1pt,
    breakable,
    enhanced
]
You are an expert Martian geologist AI. Your task is to identify all relevant terrain classes present in Martian surface images. You will be provided with an image of the Martian surface.

Your job is to visually analyze the image and determine which morphological classes are present.

You must respond with the corresponding integers for all applicable classes. \textbf{Note:} A single image may contain multiple classes.

Below are the possible classes, their corresponding integer labels, and definitions:

\begin{itemize}[left=2em]
    \item \textbf{0: Background} \\
    Generic regions that do not contain any crater or relevant morphological features.
    \item \textbf{1: Other} \\
    Craters or terrain that do not fall under the predefined morphological categories; may include ambiguous or undefined features.
    \item \textbf{2: Layered} \\
    Crater ejecta with clearly layered or rampart-like deposits, such as LERS (Layered Ejecta Rampart Sinuous) or LARLE (Low-Aspect-Ratio Layered Ejecta).
    \item \textbf{3: Buried} \\
    Craters that are partially or mostly covered by overlying material or erosion, making their full structure less visible.
    \item \textbf{4: Secondary} \\
    Smaller craters formed by debris ejected from a larger primary impact crater; usually appear in clusters or chains.
\end{itemize}

\vspace{0.5em}
Analyze the provided image and return a list of integers representing all terrain classes visible in the image.
\end{tcolorbox}

% Prompt
\begin{tcolorbox}[
    colback=gray!5,
    colframe=gray!60!black,
    title=Prompt,
    fonttitle=\bfseries,
    sharp corners=south,
    boxrule=1pt,
    breakable,
    enhanced
]
Classify the morphological crater types present in the following image of the Martian surface.

Strictly use the format below:

\textbf{Reasoning:} [Step-by-step explanation of how you identified the classes]

\textbf{Answer:} [List of integers corresponding to the identified classes]
\end{tcolorbox}

\subsection*{mb-frost\_cls}

% System Instructions
\begin{tcolorbox}[
    colback=gray!5,
    colframe=gray!60!black,
    title=System Instructions,
    fonttitle=\bfseries,
    sharp corners=south,
    boxrule=1pt,
    breakable,
    enhanced
]
You are an expert Martian climate science AI. Your task is to analyze high-resolution visible imagery from the HiRISE instrument on the Mars Reconnaissance Orbiter (MRO) and determine the presence or absence of seasonal frost.

You will be provided with an image of the Martian surface. Your job is to visually analyze the image and identify whether frost is present or not.

Respond with only the class name that best describes the image content: \texttt{``frost''} or \texttt{``non\_frost''}.

Here are the possible classes and their descriptions:
\begin{itemize}[left=2em]
    \item \textbf{frost}: The image contains visible signs of seasonal surface frost, such as bright or whitish patches consistent with CO$_2$ or H$_2$O frost.
    \item \textbf{non\_frost}: The image does not contain any visible signs of surface frost; typical terrain or landform features are exposed without seasonal coverage.
\end{itemize}

\vspace{0.5em}
Analyze the provided image and respond with only one of the two class labels.\\
You must respond with exactly one of the following two lowercase class names: \texttt{frost} or \texttt{non\_frost}.
\end{tcolorbox}

% Prompt
\begin{tcolorbox}[
    colback=gray!5,
    colframe=gray!60!black,
    title=Prompt,
    fonttitle=\bfseries,
    sharp corners=south,
    boxrule=1pt,
    breakable,
    enhanced
]
Classify the following high-resolution image of the Martian surface.

Strictly use this format:

\textbf{Reasoning:} [step-by-step reasoning]

\textbf{Answer:} [Provide only one of the two class names: \texttt{frost} or \texttt{non\_frost}]
\end{tcolorbox}

\section{Evaluation on Vision-Language Models}
\label{sec:vlm_evaluation}

\begin{table*}[htbp]
    \centering
    
    \resizebox{0.99\linewidth}{!}{
      \begin{tabular}{lcccc}
        \toprule
         & \textbf{mb-domars16k}
         & \textbf{mb-surface\_cls}
         & \textbf{mb-frost\_cls} 
         & \textbf{mb-atmospheric\_dust\_cls\_edr} \\
        \midrule
        \textbf{clip-vit-base-patch16}    & 0.53 & 0.39 & 0.96 & 0.96 \\
        \textbf{siglip-base-patch16-224}  & 0.49 & 0.30 & 0.95 & 0.96 \\
        \textbf{SmolVLM-256M}             & 0.45 & 0.27 & 0.99 & 0.99 \\
        \bottomrule
      \end{tabular}
    }
  \caption{Performance of Vision-Language Models (VLMS) on selected classification datasets.}
  \label{tab:small_vlm_results}
\end{table*}

Recent advances in multimodal foundation models have demonstrated strong generalization capabilities across diverse visual and textual domains. To assess how such models perform within the \dataset{} benchmark, we extend our evaluation to three representative vision-language models (VLMs): CLIP \cite{radford2021learning}, SigLIP \cite{tschannen2025siglip}, and SmolVLM \cite{marafioti2025smolvlm}. Due to time constraints, we currently report fine-tuning results for classification tasks, and plan to include evaluations on other task types in future revisions.

We fine-tune the models on four representative classification datasets: (1) mb-domars16k, (2) mb-surface\_cls, (3) mb-frost\_cls, and (4) mb-atmospheric\_dust\_cls\_edr. These datasets encompass a diverse set of geologic and atmospheric phenomena: ranging from landmark recognition (15 classes) and surface type classification (36 classes) to binary detection of frost and atmospheric dust, enabling a comprehensive evaluation of the models’ generalization capabilities.

Table~\ref{tab:small_vlm_results} presents the fine-tuned results in terms of weighted F1-score. All three VLMs exhibit strong performance on the binary classification tasks mb-frost\_cls and mb-atmospheric\_dust\_cls\_edr, achieving F1-scores of approximately 0.96 for CLIP and SigLIP, and up to 0.99 for SmolVLM. These results indicate that the models effectively differentiate between visually distinct binary categories (e.g., frost vs. non-frost and dusty vs. non-dusty).

Performance declines on multi-class datasets, with mb-domars16k yielding moderate performance (average F1-score of 0.49) and mb-surface\_cls performing the worst (average F1-score of 0.32). The reduced performance in mb-surface\_cls is largely due to the high intra-class visual similarity among its 36 surface categories, making the task substantially more challenging for VLMs.

As discussed in Section 5.4 (main paper), we also evaluated GPT and Gemini models in zero-shot settings. Both models follow similar trends: high performance on binary tasks and lower scores on multi-class datasets, further validating the difficulty distribution and generalization spectrum of MarsBench.

\section{Societal Impact}
\label{sec:societal_impact}

This work introduces \dataset, a standardized benchmark aimed at advancing the development and evaluation of foundation models for Martian orbital and surface imagery. As a contribution to fundamental research in planetary science, it does not present any direct or immediate societal risks. The primary beneficiaries are planetary scientists and computer vision researchers focused on accelerating geological discovery on Mars and exploring domain adaptation in machine learning across specialized, low-resource domains.

% Our work introduces Mars-Bench, a standardized benchmark for developing and evaluating foundation models on Martian orbital and surface imagery. As foundational research in planetary science, there is no direct application that poses immediate societal harm; the primary beneficiaries are planetary scientists and computer scientists seeking to accelerate geological discovery of Mars and domain adaptation studies.

Moreover, \dataset{} draws on expert-annotated, small-scale datasets which may reflect biases in geographical sampling (e.g., over-representation of certain landing sites or terrains). While these biases do not impact human groups directly, they could influence model performance unevenly across different Martian regions. We therefore urge future work to expand dataset diversity, report performance across partitions, and explore techniques for addressing data imbalance (e.g., re-sampling, domain adaptation).

% In summary, while Mars-Bench itself presents minimal direct societal risk.

% responsibly handling geospatial AI advances, through transparent licensing, bias assessment, and community standards, will be essential as these methods migrate to broader Earth- and space-based applications.

% \newpage
% \bibliographystyle{plain}
% \bibliography{references}

% \end{document}

\end{document}